\documentclass[a4paper,fleqn]{cas-dc}

\usepackage[numbers]{natbib}
\usepackage{amsmath,amssymb,amsfonts}
\usepackage{bbm}
\usepackage{caption}
\usepackage{algorithmic}
\usepackage{graphicx,multicol,multirow}
\usepackage{textcomp}
\usepackage{subfigure}
\usepackage{float}
\usepackage{makecell}
\usepackage{diagbox}

\usepackage[lined,boxed,ruled,commentsnumbered]{algorithm2e}
\def\tsc#1{\csdef{#1}{\textsc{\lowercase{#1}}\xspace}}
\tsc{WGM}
\tsc{QE}

\begin{document}
\let\WriteBookmarks\relax
\def\floatpagepagefraction{1}
\def\textpagefraction{.001}

\shorttitle{3D instance segmentaion}    

\shortauthors{Y. XU and S. Arai}  

\title [mode = title]{FPCC: Fast Point Cloud Clustering based Instance Segmentation for Industrial Bin-picking}  

\tnotemark[1] 

\tnotetext[1]{This research did not receive any specific grant from funding agencies in the public, commercial, or
not-for-profit sectors.} 

%

\author[1]{Yajun Xu}[]


\ead{y_xu@sdm.ssi.ist.hokudai.ac.jp}


\credit{designed the algorithm, carried out the experiment, and wrote this manuscript}

\affiliation[1]{organization={Department of Systems Science and Informatics, Graduate School of Information Science and Techn, Hokkaido University},
            addressline={Kita 14, Nishi 9, Kita-ku}, 
            city={Sapporo},
            postcode={060-0814}, 
            state={Hokkaido},
            country={Japan}}

\author[2]{Shogo Arai}[]
\cormark[1]

\ead{arai@tohoku.ac.jp}


\credit{designed the algorithm, carried out the experiment, and wrote this manuscript}

\affiliation[2]{organization={Department of Robotics, Graduate School of Engineering, Tohoku University},
            addressline={Aoba-6-6 Aramaki, Aoba Ward}, 
            city={Sendai},
            postcode={980-8579},  
            state={Miyagi},
            country={Japan}}

\cortext[1]{Corresponding author}


\author[2]{Diyi Liu}[]
\ead{diyi.liu.890101@outlook.com}
\credit{designed a part of the algorithm}

\author[3]{Fangzhou Lin}[]
\ead{lin.fangzhou.p2@dc.tohoku.ac.jp}
\credit{designed a part of the algorithm, carried out a part of experiment, and worte this manuscript}
\affiliation[3]{organization={Department of Applied Information Sciences, Graduate School of Information Sciences, Tohoku University},
            addressline={Aoba-6-309 Aramaki, Aoba Ward}, 
            city={Sendai},
            postcode={980-8579},  
            state={Miyagi},
            country={Japan}}

\author[2]{Kazuhiro Kosuge}[]

\ead{kosuge@irs.mech.tohoku.ac.jp}



\begin{abstract}
Instance segmentation is an important pre-processing task in numerous real-world applications, such as
robotics, autonomous vehicles, and human-computer interaction. 
Compared with the rapid development of deep learning for two-dimensional (2D) image tasks, 
deep learning-based instance segmentation of 3D point cloud still has a lot of room for development. 
In particular, distinguishing a large number of occluded objects of the same class is a highly challenging problem, which is seen in a robotic bin-picking. 
In a usual bin-picking scene, many identical objects are stacked together 
and the model of the objects is known. 
Thus, the semantic information can be ignored; instead, the focus in the bin-picking is put on the segmentation of instances. 
Based on this task requirement, we propose a Fast Point Cloud Clustering (FPCC) for instance segmentation of bin-picking scene. FPCC includes a network named FPCC-Net and a fast clustering algorithm. 
FPCC-net has two subnets, one for inferring the geometric centers for clustering 
and the other for describing features of each point. 
FPCC-Net extracts features of each point and infers geometric center points of each instance simultaneously.  
After that, the proposed clustering algorithm clusters the remaining points 
to the closest geometric center in feature embedding space. 
Experiments show that FPCC also surpasses the existing works in bin-picking scenes and is more computationally efficient. Our code and data are available at \url{https://github.com/xyjbaal/FPCC}.
\end{abstract}



\maketitle
\section{Introduction}
%
%
%
%
Acquisition of three-dimensional (3D) point cloud is no longer difficult due to advances in 3D measurement technology, such as passive stereo vision\cite{stereo,stereo_camera_sensor_scheduling_communication,fss_sds,stereo_vision_1}, phase shifting method\cite{phase-shifting}, gray code\cite{gray_code}, and other methods\cite{arai_3d, gray_code_phase_shift}.  
As a consequence, efficient and effective processing of 3D point cloud has become a new challenging problem. Segmentation of 3D point cloud is usually required as a pre-processing step in real-world applications, such as autonomous vehicles\cite{L3-Net}, human-robot interaction\cite{F-PREMO,kanazawa,AR}, robotic bin-picking\cite{bin-picking_1,PPF-MEAM,PPF_MEAM_robio2018,bin_picking_part_seg}, pose estimation\cite{PPF_MEAM_robio2019,ppr-net,PPF_MEAM_2,Pose_Estimation,Pose_Estimation_rgb}, visual servoing\cite{PC-serve,Visual_Servoing_2}, and various types of 3D point cloud processing\cite{path_planning,3d_keypoint_detection,3D-upsampling,FADA-3K,Tooth_ins_seg_3d}. In the field of robotics, bin-picking scenes have received a wide range of attention in the past decade. In this scene, many objects of the same category are stacked together. The difficulty of bin-picking scenes in logistics warehouses is that there are too many categories and unknown objects\cite{ins_seg_Amazon_Picking,grasp_Amazon_Picking,6D_pose_estimation_Amazon_Picking}, while the problem of industrial bin-picking scenes is that it is difficult to distinguish the same objects and make datasets. At present, an application of convolutional neural networks (CNNs) to instance segmentation of 3D point cloud is still far behind its practical use. The technical key points can be summarized as follows: 1) convolution kernels are more suitable for handling structured information, while raw 3D point cloud is unstructured and unordered; 2) the availability of high-quality, large-scale image datasets\cite{coco,imagenet,places} has driven the application of deep learning to 2D images, but there are fewer 3D point cloud datasets; and 3) instance segmentation on 3D point cloud based on CNNs is time-consuming.

For key point 1), PointNet\cite{PointNet} has been proposed as the first framework which is suitable for processing unstructured and unordered 3D point clouds. 
PointNet does not transform 3D point cloud data to 3D voxel grids such as\cite{voxelized-point,3d-shapenets}, 
but uses multi-layer perceptions (MLPs) to learn the features of each point and has adopted max-pooling to obtain global information. 
The pioneering work of PointNet has prompted further research, and several researchers have introduced the structure of PointNet as the backbone of their network\cite{PointNet++,SGPN,jsis3d}. 
It is known that PointNet processes each point independently and it results in learning less local information\cite{PointNet++,DGCNN}. 
To enable learning of the 3D point cloud's local information, the methods proposed in\cite{GAC,point_web,DGCNN,PointCNN,recurrent,DGCNN_2,Geom_gcn,PointVGG} have increased the network's ability to perceive local information by exploring adjacent points. 
Following our previous work\cite{xu}, we employ DGCNN\cite{DGCNN} as our feature extractor because DGCNN is flexible and robust to process point clouds with only coordinates.

For key point 2), some well-known 3D point cloud datasets include indoor scene datasets such as S3DIS\cite{S3DIS} and SceneNN\cite{scenenn}, driving scenario datasets such as KITTI dataset\cite{KITTI} and Apollo-SouthBay dataset\cite{Apollo}, and single object recognition dataset likes ShapeNet dataset\cite{3d-shapenets}. 
For robotic bin-picking, it is a huge and hard work to provide a general training dataset of various industrial objects and there is no such dataset currently. 
Synthesizing training data through simulation provides a feasible way to alleviate the lack of training dataset\cite{large-binpicking,Training_syn,xu,SSD6D,AAE,SD-Mask-R-CNN}. At this stage, we argue that training the network with synthetic data is an economical and feasible strategy. Our network is trained by synthetic dataset and shows acceptable results on real data.

For key point 3), the reasons why instance segmentation on 3D point cloud by CNNs is time-consuming are described as follows. 
Instance segmentation locates different instances, even if they are of the same class. 
As instances in the scene are disordered and their number is unpredictable, it is impossible to represent instance labels with a fixed tensor. 
Therefore, the study of instance segmentation includes two methods: the proposal-based method requiring an object detection module and the proposal-free method without an object detection module. 
Proposal-based methods require complex post-processing steps to deal with many proposal regions and have poor performance in the presence of strong occlusion. 
For the instance segmentation of 3D point cloud, most researchers adopt the proposal-free method\cite{SGPN,jsis3d,asis,PointGroup,HAIS,OccuSeg,3D_MPA}. 
The proposal-free method usually performs semantic segmentation at first and then distinguishing different instances via clustering or metric learning\cite{SGPN,jsis3d,asis,xu}. The current clustering methods first generates multiple candidate groups and then merge them, which is a very time-consuming process. In contrast, our clustering algorithm does not generate candidate groups, but directly generates instances based on the feature distance between the object’s center point and the rest of the points. This way dramatically improves the speed of instance generation and avoids the case that a point belongs to multiple instances at the same time.
\begin{figure}[!t]
\centering
\subfigure{
\includegraphics[width=1.\columnwidth]{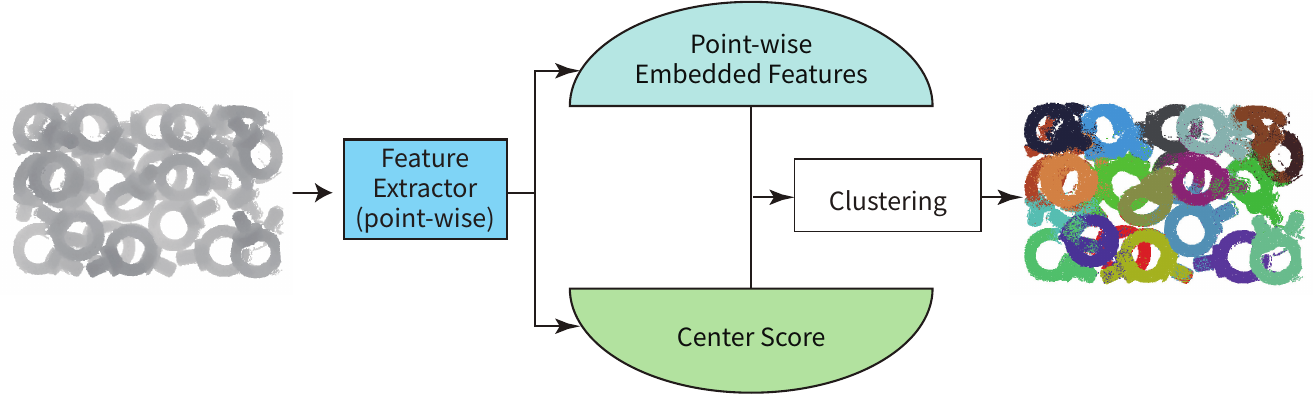}}
\caption{Instance segmentation results using FPCC-Net. FPCC-Net has two branches: the embedded feature branch and the center score branch.}
\label{fig: illustrator}
\end{figure}

This paper aims to design and propose a fast point cloud clustering for instance segmentation method named FPCC consisting of FPCC-Net and a fast clustering algorithm based on the output of FPCC-Net.  
FPCC-Net is a graph convolutional neural network that can effectively segment the 3D point cloud at instance-level without training by any manually annotated data. 
FPCC-Net involves mapping all points to a discriminative feature embedding space, which satisfies the following two conditions: 1) points of the same instance have similar features, 2) points of different instances are widely separated in the feature embedding space. 
Simultaneously, FPCC-Net finds center points for each instance, 
and the center points are used as the reference point of the clustering process. 
After that, the fast clustering is performed based on the center points as shown in Fig. \ref{fig: illustrator}.

The main contributions of this work are as follows:
\begin{itemize}
\item A high-speed instance segmentation scheme for 3D point cloud is proposed. 

\item The proposed scheme consists of a novel network of 3D point cloud for instance segmentation named FPCC-Net and a novel clustering algorithm using the found center points. 

\item A hand-crafted attention mechanism is introduced into the loss function to improve the performance of FPCC-Net, and its effectiveness is verified in an ablation study. 

\item Experiments show that FPCC-Net trained by synthetic data demonstrates excellent performance on real-world data compared with existing methods. 

\item We annotate instance information for parts that have not been labeled in XA Bin-Picking dataset \cite{xu}. The completed dataset is available at \url{https://github.com/xyjbaal/FPCC}.

\end{itemize}
The remainder of this paper is organized as follows. Section \ref{Related Work} discusses the progress of instance segmentation on images and 3D point cloud. 
Section \ref{method} shows the structure and principle of FPCC-Net. 
Experimental analyses are provided in Section \ref{experiment}. 
Finally, Section \ref{conclusions} concludes the paper. 

We use the following notations in this paper. 
A real number set is represented by $\mathbb{R}$. 
A coordinate of point $i$ is denoted by $p_i = (x_i, y_i, z_i) \in \mathbb{R}^{3}$. 
Point cloud containing $N$ points is denoted by $\mathbb{P} = \{p_1,p_2,...,p_N\}$. 
Distance function is denoted by
\begin{align}
d(a, b) = \Vert a - b \Vert_2,
\end{align}
where $d(a, b)$ denotes Euclidean distance between $a \in \mathbb{R}^{n}$ and $b \in \mathbb{R}^{n}$. 
For a matrix $A \in \mathbb{R}^{n \times m}$, $(i, j)$-th element of A is denoted by $a_{(i,j)}$.

\begin{figure*}[!t]
\centering
\includegraphics[width=2\columnwidth]{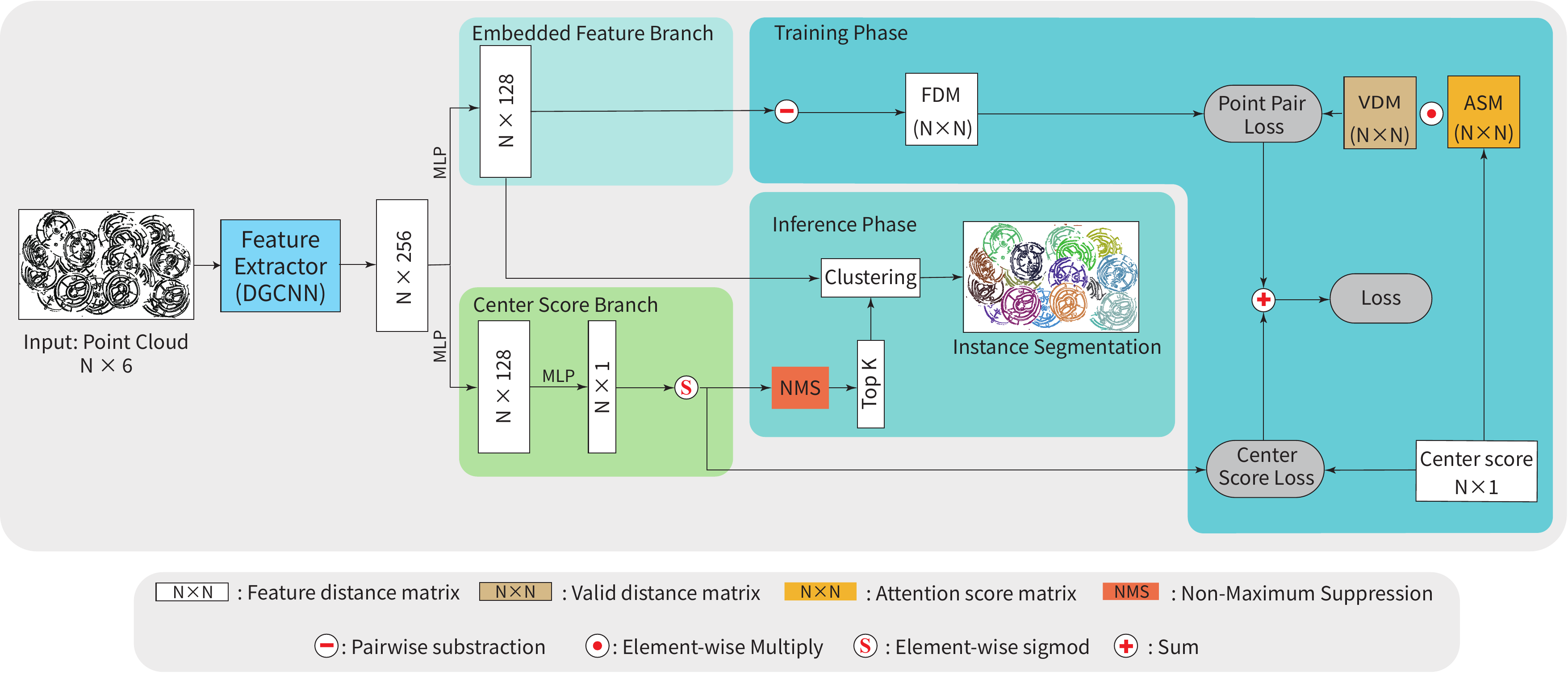}
\caption{Network architecture of FPCC-Net. 
The represented 3D point cloud $[\overline{x}_i, \overline{y}_i, \overline{z}_i, n_{i,x}, n_{i,y}, n_{i,z}]^\top$ $(i=1,2,\cdots,N)$ is fed into the network and the instance label is outputted for each point. 
The features of each point are extracted using a feature extractor and then sent to two respective branches. The embedded feature branch extracts 128-dimensional features of each point and the center score branch predicts the center score of each point. 
For supervising FPCC-Net, the following matrices are introduced. 
Valid distances matrix defined by (\ref{eq:vdm}) is a binary matrix used to ignore some point pairs whose distance is greater than a certain threshold. 
Attention score matrix defined by (\ref{eq: asm}) is used to increase the weight of point pairs closer to the center position.}
\label{Fig:FPCC}
\end{figure*}

\section{Related Works}
\label{Related Work}
With the emergence of CNNs, the methods of feature extraction from images and 3D point cloud have been changing from manual design to automatic learning\cite{hand-crafted,hand-crafted1,hand-crafted2}. 
Instance segmentation is one of the most basic tasks in the field of computer vision and receives much attention. 
Segmentation on two-dimensional (2D) images has been almost fully developed\cite{PanopticSeg,Panoptic_fpn}, but 3D point cloud segmentation has remained underdeveloped. 

\subsection{Instance segmentation on 2D}
Current 2D instance segmentation methods can be roughly divided into two categories, two-stage method, and one-stage method. Two-stage means they perform object detection first by utilizing proposal generation, then followed by mask prediction. Mask R-CNN \cite{Mask-R-CNN} is one of the representative two-stage instance segmentation methods. Mask R-CNN decomposes the problem of instance segmentation into object detection and pixel-level segmentation of a single object. Mask R-CNN has added a branch to predict object masks based on Faster R-CNN \cite{Faster-Rcnn}. 
Based on the Mask R-CNN\cite{Mask-R-CNN} and FPN\cite{FPN}, PANet\cite{PANet} enhanced mask prediction by improving the information propagation between lower and higher levels. MaskLab\cite{MaskLab} predicted the direction of each pixel to its corresponding instance center for instance segmentation of the same class. TensorMask\cite{TensorMask} regards dense instance segmentation as a prediction task over 4D tensor and proposes a general framework for operations on 4D tensors. 
The one-stage methods are usually faster than the two-stage methods because the one-stage methods do not contain the proposal generation and pooling step. YOCLAT\cite{YOLACT} generates instance masks by linearly combining prototype and mask coefficients, sacrificing a little performance for computational speed. BlendMask\cite{BlendMask} combined instance-level information with semantic information to enhance mask prediction, and CenterMask\cite{CenterMask} added a spatial attention-guided mask branch to the object detector to predict masks. PolarMask\cite{PolarMask} describes the mask of each instance by its center and rays emitted from the center to the contour. 
Other approaches which are not categorized into one-stage nor two-stage methods, such as clustering or metric learning, do not perform well on 2D instance segmentation\cite{discri_loss, deep_watershed_transform}. 
Although the deep learning-based methods on 2D instance segmentation utilizing a large number of high-quality and manually labeled datasets\cite{coco, cityscapes} have great achievements, they still have drawbacks in industrial bin-picking scenes, where labeled datasets are often difficult to obtain. 
To overcome the difficulties of making datasets, SD Mask R-CNN\cite{SD-Mask-R-CNN} has been proposed as an extension of Mask R-CNN. The paper\cite{SD-Mask-R-CNN} has presented a method to generate synthetic datasets rapidly, and their network was trained only with the generated synthetic depth images instead of RGB images. 
However, SD Mask R-CNN does not show enough performance, especially for multiple object instances with occlusion in the scene.

\subsection{Instance segmentation on 3D point cloud}
Some researchers have adopted proposal-based methods that detect objects and predict the instance masks. 3D-SIS\cite{3D-SIS} have combined 2D images and 3D geometric information to infer the bounding boxes of objects in 3D space and the corresponding instance mask. The demand for 2D images limits the application of this method because it is expensive to train the network with 2D images.  
GSPN\cite{GSPN} continued the idea of Mask R-CNN. They first predicted the candidate regions, and then refined the candidate regions with R-PointNet to obtain the result of instance segmentation.  
However, the region proposal network of 3D-SIS and GSPN consumes a long computing time. 3D-MPA\cite{3D_MPA} has used an object-centric voting scheme to generate instance proposals, which are robust to potential outliers in the instance proposal stage. 
3D-BoNet\cite{3dbonet} provided by B. Yang et al. directly regresses the bounding box of each instance in the 3D point cloud and simultaneously predicts point-level masks for each instance. 3D-BoNet performs well on the completed point cloud because there is little overlap of objects’ bounding box in these datasets e.g., S3DIS \cite{S3DIS}, ScanNet\cite{Scannet}. In other words, the overlap and incompleteness of objects in the bin-picking scenes make 3D-BoNet difficult to regress reasonable bounding boxes.  

Proposal-free methods extract the features of the points at first and then group points into instances.  
SGPN\cite{SGPN} is the first direct 3D instance segmentation method that assumed the points of the same instance should have similar features. 
Three sub-networks predict semantic labels, confidence score, and point-wise features, respectively.  
The confidence score of each point indicates the confidence of the reference point of clustering. 
SGPN\cite{SGPN} have highlighted an interesting phenomenon that the clustering confidence scores of the points located in the boundary area are lower than others. 
Inspired by this, FPCC takes only one point that is most likely to be the geometric center of an object as the reference point of clustering for the object. Some methods combining semantic with instance information have been proposed to improve performance\cite{asis,jsis3d,3d_mt}.
Q. Phm et al.\cite{jsis3d} have used a Multi-Value Conditional Random Field to learn semantic and instance labels simultaneously. 
J. Lahoud et al.\cite{3d_mt} have performed instance segmentation by clustering 3D points and mapping the features of points to the feature embedding space according to relationships of point pairs. 
However, the objects in the bin-picking scene are all of the same types. Thus the semantic information of each point is the same. The way of combining semantic and instance information with each other loses its effectiveness in this case. PointGroup\cite{PointGroup} and HAIS\cite{HAIS} employ a similar strategy. They predict the offset vector for moving each point towards its instance center and then cluster the moved points in Euclidean space.
OccuSeg\cite{OccuSeg} has constrained the clustering based on predicted occupancy size and the clustered occupancy size, which help to correctly cluster hard samples and avoid over-segmentation. 
B, Zhang, et al.\cite{3d_Probabilistic_Embeddings} have presented a probabilistic embedding framework to encode the features of each point and a novel clustering step.

Although previous proposal-free methods have used various ways to extract features of points, they all need to find points far greater than the number of instances as reference points for clustering, and each reference point corresponds to a potential group. Each group is merged by intersection over union (IoU). The process of merging takes a lot of time. The reason for time-consuming is described in more detail in Section \ref{Time}. In contrast, FPCC does not need the process of merging. 
In addition, all these methods are based on public datasets such as S3DIS\cite{S3DIS} and SceneNN\cite{scenenn} with rich annotations.
Making such a dataset for bin-picking scenes is a time-consuming and laborious task\cite{xu}. FPCC shows an acceptable performance on real-world data even trained by synthetic data.  
\subsection{Instance segmentation for bin-picking scenes}
Overall, the existing instance segmentation methods for bin-picking scenes can be divided into two groups: logistics-oriented methods and industrial-oriented methods. The former is multi-class, multi-instance learning. The latter tends to be one-class, multi-instance learning from cluttered scenes without predicting semantic labels. Researchers\cite{2d_bin_picking,Joint_ins_seg_binpicking,seg_binpicking,SD-Mask-R-CNN,mask-bin-picking, 6D_pose_estimation_Amazon_Picking} have adopted the mainstream 2D detection or 2D instance segmentation network for logistics scenes to locate objects. However, logistics scenes are usually less cluttered than industrial scenes.

Currently, few works have focused on instance segmentation for industrial bin-picking scenes. Because industries primarily use point cloud, and current 3D instance segmentation is far slower than 2D instance segmentation and challenging to annotate. D. Liu et al.\cite{PPF_MEAM_2} have located the bounding boxes of unoccluded objects from RGB image by RetinaNet and then projected the bounding boxes into 3D space. 
Further, M. Grard et al.\cite{unoccluded_binpicking} proposed a residual encoder-decoder design to predict the masks of unoccluded objects. 
They did not segment industrial objects in 3D space and required 2D image annotation. 
W. Abbeloos et al.\cite{ppf_detection} used point pair features to match model points and scene points. They have reduced the complexity of the problem by reducing the search space using simple heuristics. However, the number of objects found in this way is affected by the degree of clutter, and the computation time is sensitive to the number of points. 
PPR-net\cite{ppr-net} has regressed the 6D transform of the instances corresponding to each point, and then a density-based clustering method clustered these transformed points in the pose space. The primary goal of PPR-net is to regress the 6D transform of the objects, and the results of instance segmentation are generated based on pose regression.

\section{Method}
\label{method}
This paper proposes a novel clustering method for instance segmentation on the 3D point cloud. 
The training data are 3D point cloud without color and can be automatically generated in simulation by using a 3D shape model of the target object. 
The main idea of fast clustering is to find geometric centers of each object, and then use these points as reference points for clustering. 
Compared with the existing cluster-based methods, we have two advantages. The first is that we can achieve an acceptable result by synthetic point cloud without color information. The second is that, theoretically, we can find center points equivalent to the number of instances in the scene as reference points for clustering, so there is no need for redundant merged algorithms \cite{asis,SGPN}, which consumes a lot of computing time and easily introduce errors in severely overlapping scenes.
\subsection{Backbone of FPCC-Net}
In the first step, coordinates of original 3D point cloud $p_i = (x_i, y_i, z_i)$, $i=1,2,...,N$ is converted to new coordinate system by 
\begin{equation}
\begin{split}
\overline{x}_i &= x_i - \min{\{x_1, x_2, ..., x_N\}} \\
\overline{y}_i &= y_i - \min{\{y_1, y_2, ..., y_N\}} \\
\overline{z}_i &= z_i - \min{\{z_1, z_2, ..., z_N\}},
\end{split}
\end{equation}
Then converted each point is represented by a six-dimensional (6D) vector of $\overline{x}$, $\overline{y}$, and $\overline{z}$ and a normalized location ($n_x, n_y, n_z$) as to the whole scene (from 0 to 1). The represented 3D point cloud is fed into the network and outputs are 128-dimensional features and the center score of each point.

As shown in Fig. \ref{Fig:FPCC}, FPCC-Net has two branches that encode the feature of each point in the feature embedding space and infer each point's center score. 
First, the point-wise features of $N$ points are extracted through a feature extractor. 
In FPCC-Net, DGCNN\cite{DGCNN} without the last two layers is adopted as the feature extractor. 
DGCNN has better performance than PointNet in extracting the features of point cloud without color\cite{xu}. 
The extracted point-wise features with size $N \times 256$ are fed into two branches, \textbf{embedded feature branch} and \textbf{center score branch}. 

In the embedded features branch, the extracted features pass through an MLP to generate an embedded feature with size $N \times 128$. 
The center score branch is parallel to the embedded feature branch and used to infer the center score of each point. 
In the center score branch, the point-wise features generated by the feature extractor are activated by a sigmoid function after passing through two MLP. Then, predicted center score $\widehat{s}_{\mathrm{center}}$ with size $N \times 1$ is obtained. 
After the prediction of the center scores, We use algorithm \ref{NMS} to find the points most likely to be the geometric centers of each object, and the found points are taken as reference points in the clustering process.

\subsection{Inference phase}
As described in the previous section, two branches of FPCC-Net output the embedded features and the center scores of each point. 
Non-maximum suppression is performed on all points with center scores to find the centers of each instance. 
The points with a center score higher than $0.6$ are considered as candidates of the center points. 
The point with the highest center score is selected as a first candidate of the center point, and all the other points located in the sphere with the center being the candidate point and radius $\gamma d_{\rm{max}}$ are removed, where $d_{\max}$ is the maximum distance from the geometric center to the farthest point of the object, and $\gamma$ is the screening factor.
This process is repeated until there are no more points left. 
The detailed processes of selecting the center points are presented in Algorithm \ref{NMS}.

\begin{algorithm}[t]
\caption{\label{NMS} Non-maximum suppression algorithm on points; $N$ is the number of points; $K$ is the number of center points.}   
\KwIn{
\, Threshold for center score $\theta_{\mathrm{th}}$;\\
\qquad  \qquad Screening radius $\gamma d_{\mathrm{max}}$;\\
\qquad  \qquad Set of points $\mathbb{P} = \{p_1,p_2,...,p_N\}$; \\
\qquad  \qquad Corresponding predicted center scores of \\
\qquad  \qquad points $\mathbb{S} = \{s_1,s_2,...,s_N\}$
  }
\KwOut{\,Center points $\mathbb{C}= \{c_1,c_2,...,c_K\} $}
\For{$i=1$ \rm{to} $N$}
{
  \If{$s_i \le \theta \mathrm{th}$}
  {
    $\mathbb{P} \gets \mathbb{P} \backslash \{p_i\} $\;
    $\mathbb{S} \gets \mathbb{S} \backslash \{s_i\} $\;
  }
}
$\mathbb{C} \gets \{\}$\;
\While{$\mathbb{P} \not= \emptyset$}
  { $m^* \gets \arg\max_{m} \{s_{m}~|~s_{m} \in \mathbb{S} \}$\;
    $\mathbb{C} \gets p_{m^*}$\;
    $\mathbb{P} \gets \mathbb{P} \backslash \{p_{m^*}\} $\;
    $\mathbb{S} \gets \mathbb{S} \backslash \{s_{m^*}\} $\;
    \For{$p_i$ in $\mathbb{P}$}
    {
      \If{ $d(p_{m^*},\ p_i) \le \ \gamma d_{\mathrm{max}}$}
      {
        $\mathbb{P} \gets \mathbb{P} \backslash \{p_i\} $\;
        $\mathbb{S} \gets \mathbb{S} \backslash \{s_i\} $\;
      }
    }
  }
  return $\mathbb{C}$\;
\end{algorithm}

After the above process, the feature distances between the center points and the other points are computed by 
\begin{equation}
\label{ins_id}
d( e_{F}^{(i)}, e_{F}^{(k)}) = \Vert e_{F}^{(i)} - e_{F}^{(k)} \Vert_2,
\end{equation}
where $e_{F}^{(k)}$ represents the feature of $k$-th center point selected by Algorithm \ref{NMS}, and $e_{F}^{(i)}$ represents the feature of $i$-th point in the remaining points. All points except the center points are clustered with the nearest center point in terms of the feature distance. Note that, we find the nearest center point $c_k$ of point $p_i$ in the feature embedding space, and then calculate Euclidean distance $d(p_i,c_k)$ between $p_i$ and $c_k$ in 3D space. If $d(p_i,c_k)$ exceeds $d_{\rm{max}}$, $p_i$ is regarded as noise, and an instance label will not be assigned to $p_i$. 

It should be emphasized that our clustering method is novel. The conventional clustering method \cite{SGPN,asis} is to downsample the points of the whole scene into multiple batches. The points of each batch are clustered, and then the points of all batches are integrated. In contrast, we perform clustering after all the points have been feed into the network in batches. After the feature and center score of each point are obtained, predicted instances are directly generated based on the feature distance from the center points without any merge or integration steps. In this way, we reduce the calculation time and reduce the accumulated errors due to the merged process. 

\subsection{Training phase}
The loss of the network is a combination of two branches: $L = L_{EF} + \alpha L_{CS}$, where $L_{EF}$ and $L_{CS}$ represent the losses of the embedded feature branch and the center score branch, respectively. 
The symbol $\alpha$ is a constant that makes $L_{EF}$ and $L_{CS}$ terms are roughly equally weighted. 
We introduce three matrices, feature distance matrix, valid distance matrix (VDM), and attention score matrix (ASM) for the learning of embedded features. 

We explain our design for the training phase in the following order: feature distance matrix, valid distance matrix, center score, attention score matrix, embedded feature loss, and center score loss. 
Attention score matrix is obtained from the center score of each point. 

\subsubsection{Feature distance matrix}
\label{FDM}
In the feature embedding space, the points belonging to the same instance should be close, while the points of different instances should be apart from each other. 
To make features of the points in the same instance similar, we introduce the following feature distance matrix $D_F \in \mathbb{R}^{N\times N}$. 
The $(i,j)$-th element of $D_F$ is represented by 
\begin{align}
d_{F(i,j)} = \Vert e_F^{(i)} - e_F^{(j)}\Vert_2. 
\end{align}

\subsubsection{Valid distance matrix}
\label{VDM}
The valid distance matrix $D_V \in \mathbb{R}^{N\times N}$ is a binary matrix in which each element is 0 or 1. The purpose of introducing $D_V$ is to make the network focus on distinguishing whether point pairs within a certain Euclidean distance belong to the same instance or not. 
In the inference phase, points are clustered based on the feature distance and the Euclidean distance of the point pair at the same time. 
If the Euclidean distance of two points exceeds twice the maximum distance $d_{\rm{max}}$, 
the two points cannot belong to the same instance.
Therefore, we ignore these point pairs with too far distance so that they do not contribute to the loss. 
The $(i,j)$-th element of $D_V$ is defined by
\begin{equation}
d_{V(i,j)} =
\left\{
             \begin{array}{lr}
                          1 & \mathrm{if}\,\, \Vert p_i - p_j\Vert_2 < 2 d_{\rm{max}}\\
                          0 & \mathrm{otherwise}
             \end{array}
\right.
\label{eq:vdm}
\end{equation}
(\ref{eq:vdm}) indicates whether the Euclidean distance between point $p_i$ and $p_j$ is within a reasonable range or not.

\subsubsection{Center score}
\label{CS}

\begin{figure}[!t]
\subfigure[$\beta = 1$]{
\includegraphics[width=0.48\columnwidth]{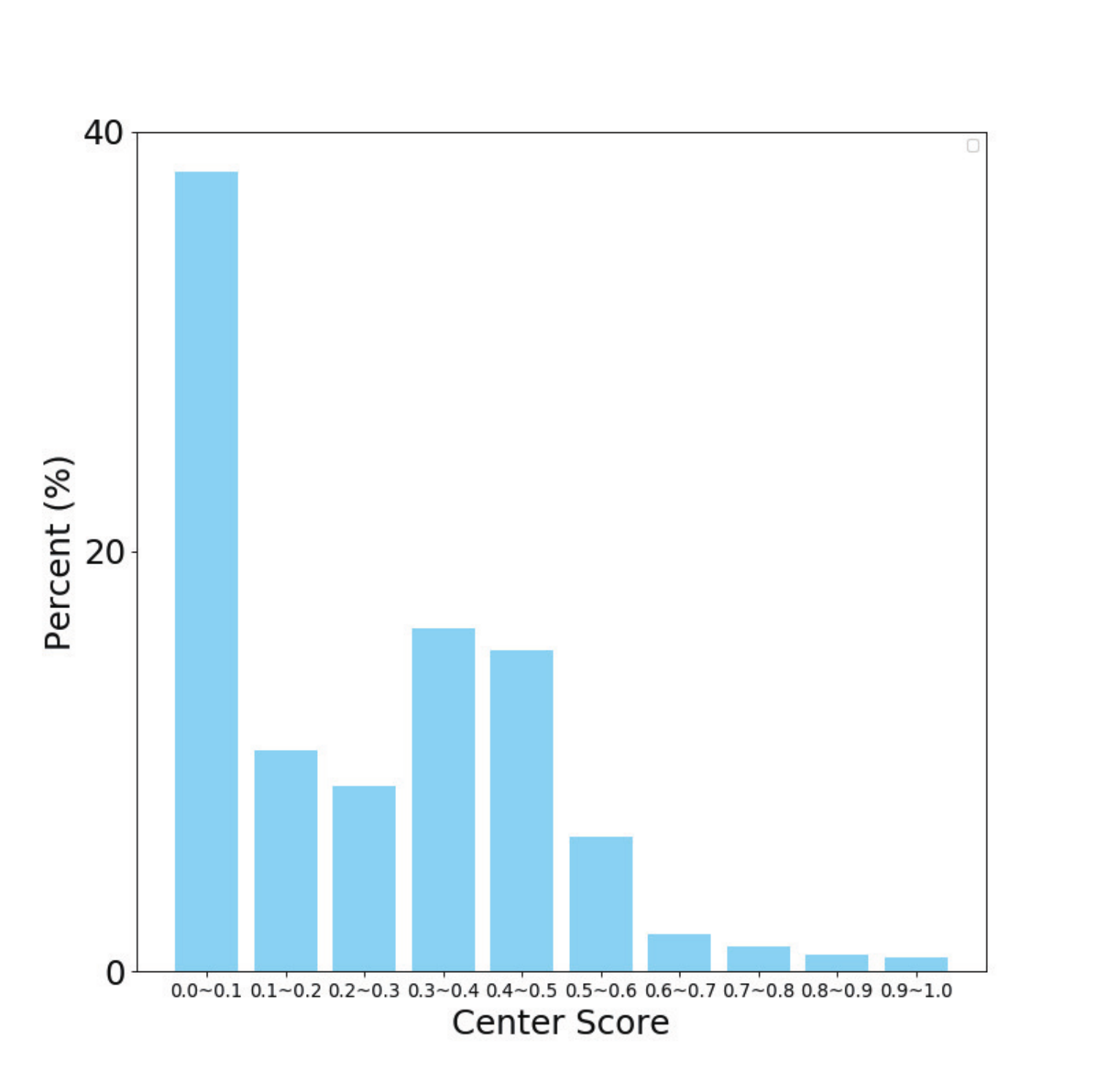}}
\subfigure[$\beta = 2$]{
\includegraphics[width=0.48\columnwidth]{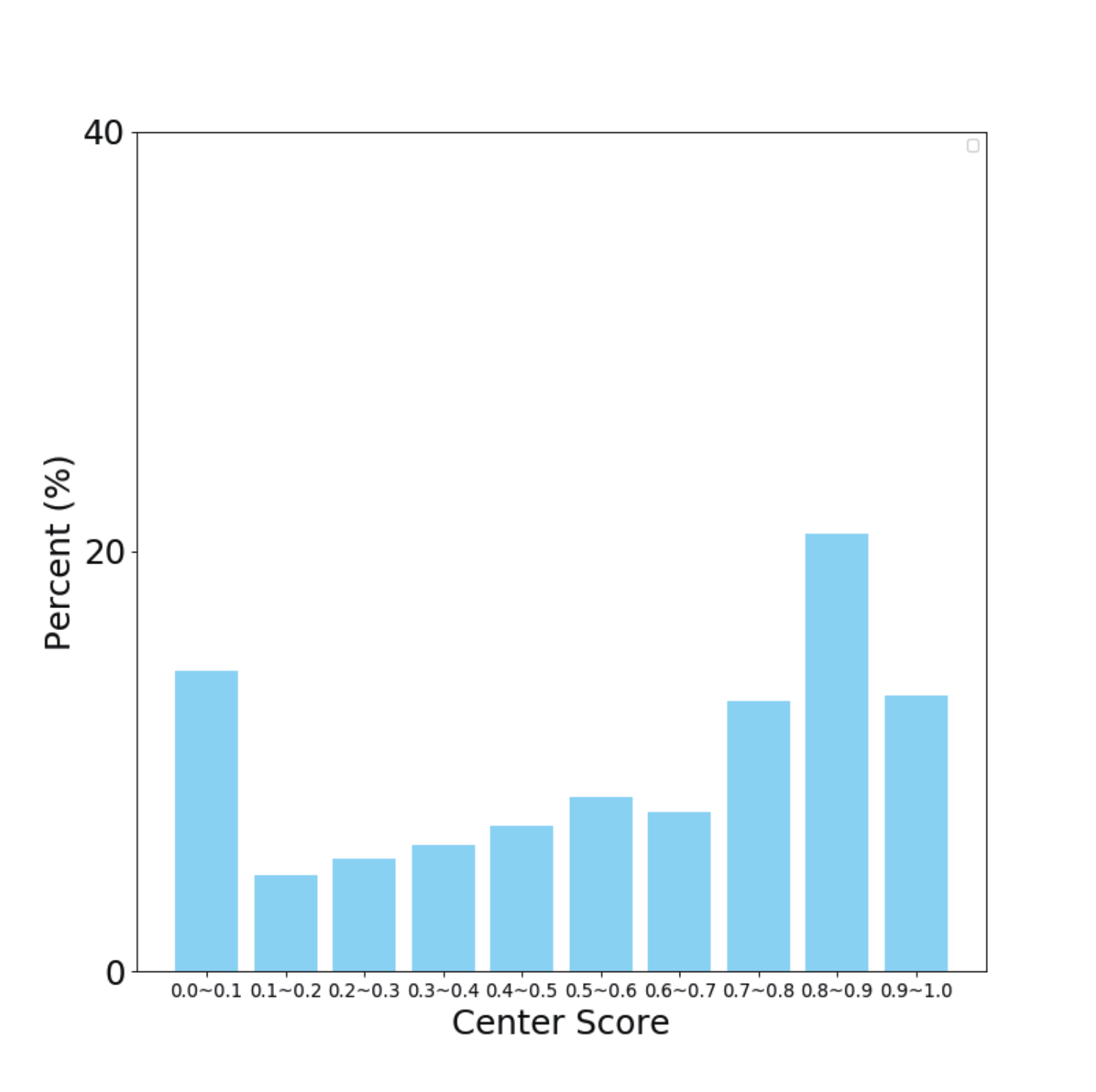}}
\caption{Distribution of the center score in the same scene. It is apparent that $\beta = 2$ makes the scores more uniform in the 0-1 interval.}
\label{Fig:distribution}
\end{figure}

\begin{figure}[!t]
\centering
\includegraphics[width=1\columnwidth]{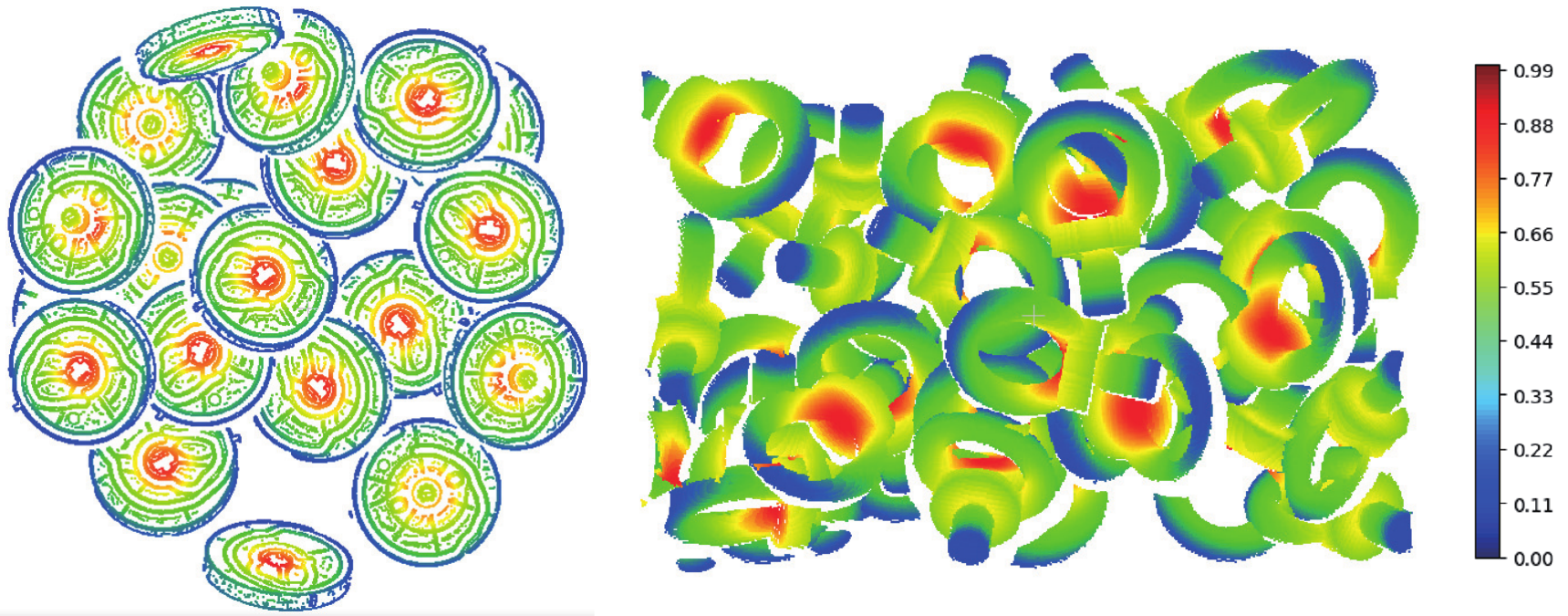}
\caption{Visualization of center score ($\beta=2$). Red indicates a higher score. 
The points at the boundary are mostly scored 0.}
\label{Fig:center_score}
\end{figure}

The center score is designed such that it should reflect the distance between a point and its corresponding center. The points near the center of object have higher scores than the points on the boundary. Based on this concept, the center score of $p_i$ is introduced by
\begin{equation}
s_{\mathrm{center}(i)} = 1 - \left(\frac{\Vert p_i-c_i \Vert_2}{d_{\rm{max}}}\right)^\beta ,
\label{eq:center}
\end{equation}
where $\beta$ is positive constant and $c_i$ is the coordinate of the geometric center of the instance to which point $p_i$ belongs. 
The value of $s_{\mathrm{center}(i)}$ is in the range $[0,1]$. 
If $\beta = 1$, the distribution of the center score will lead to imbalances, as shown in Fig. \ref{Fig:distribution}: 
only a very small number of points have higher scores, while most points have lower scores.  
This causes the center score branch to fail to effectively predict the center scores (all scores are biased towards zero). 
Fig. \ref{Fig:distribution} shows that $\beta=2$ leads more uniform balance than $\beta=1$.  
Thus, $\beta$ is set to 2 in our implementation. 
The center scores are visualized in Fig. \ref{Fig:center_score}. 
Fig. \ref{Fig:center_score} shows that the points on the boundary area are mostly scored 0, and those near the center are approximately scored 1.

\begin{figure}[!t]
\centering
\subfigure{
\begin{tabular}{ccccc}
\centering
\includegraphics[width=0.14\columnwidth]{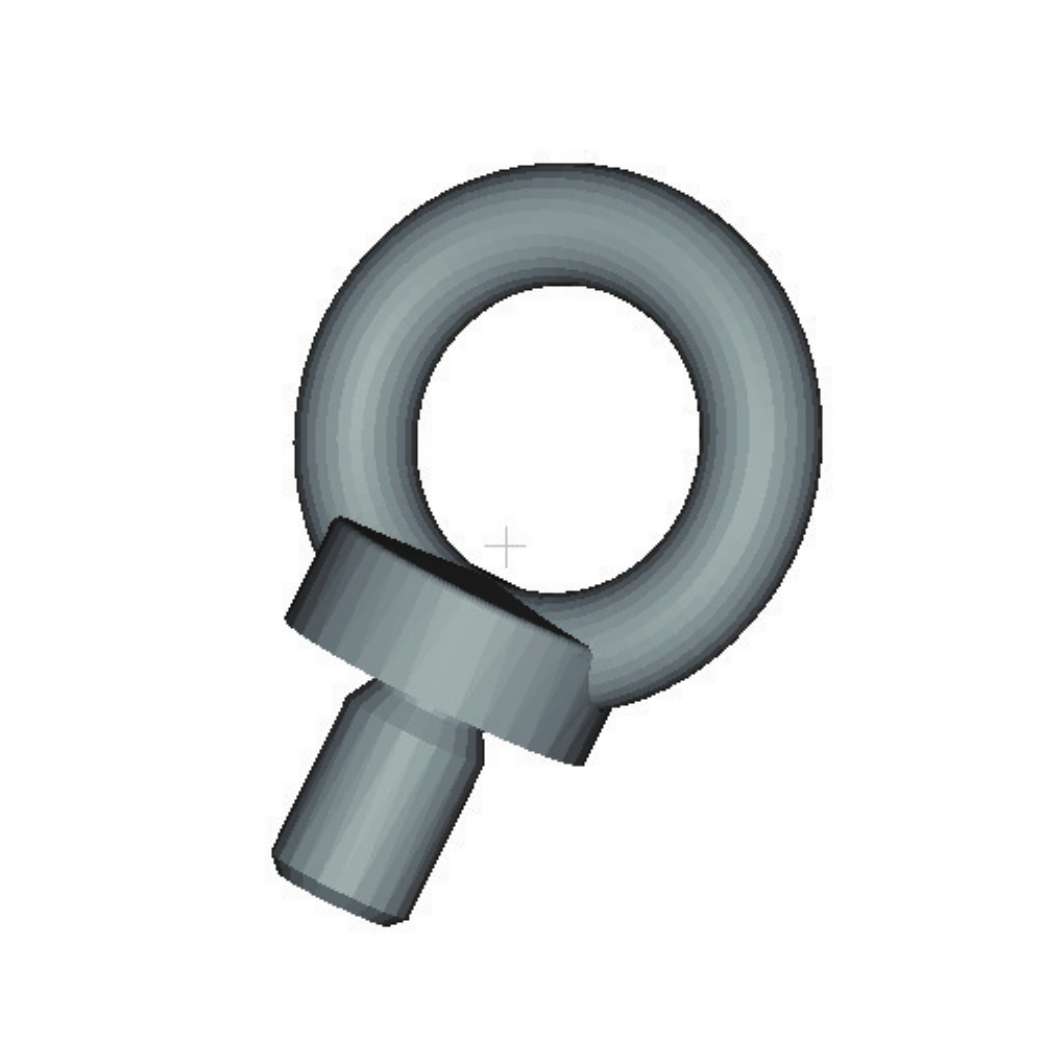}&
\includegraphics[width=0.14\columnwidth]{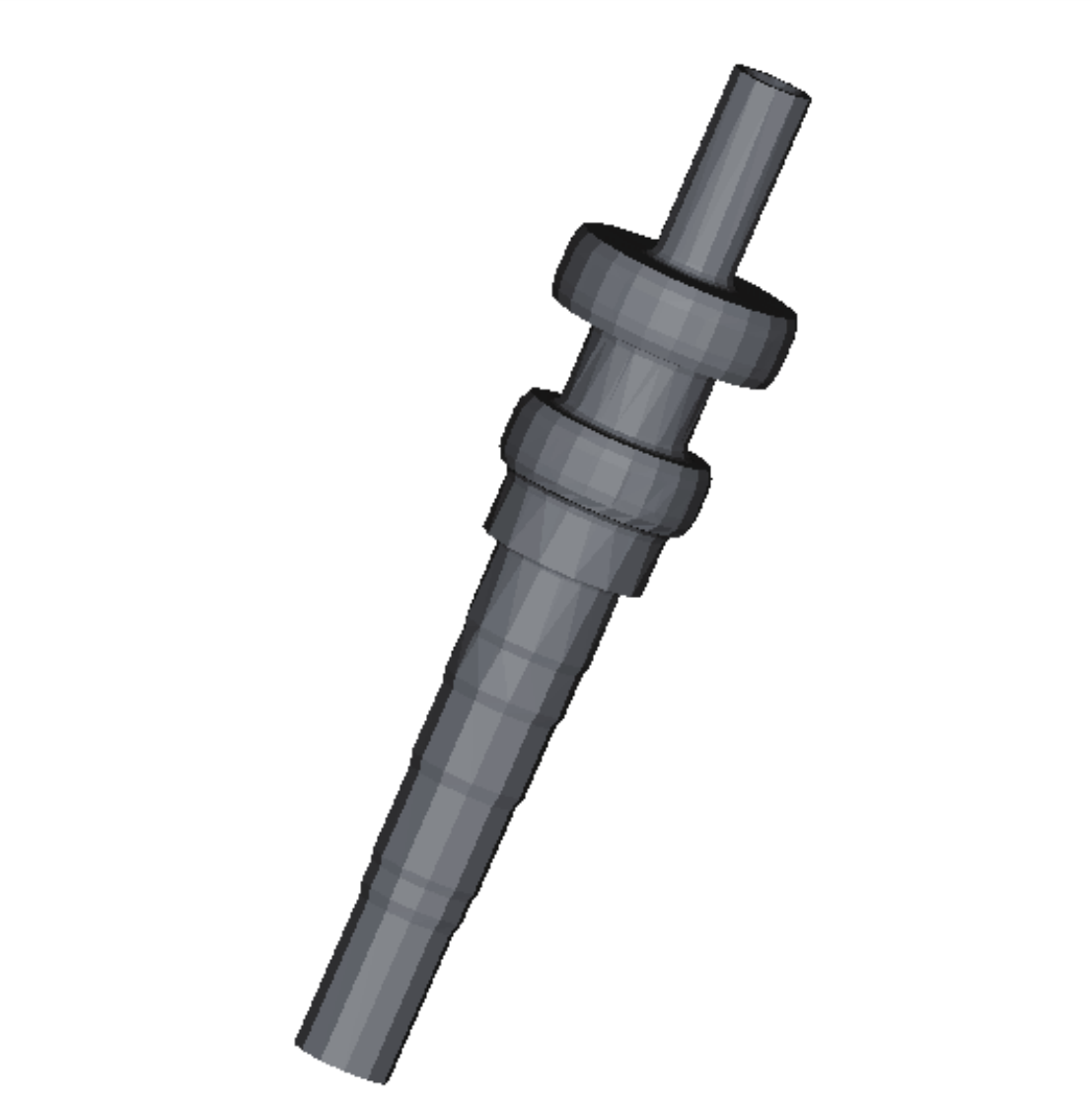}&
\includegraphics[width=0.14\columnwidth]{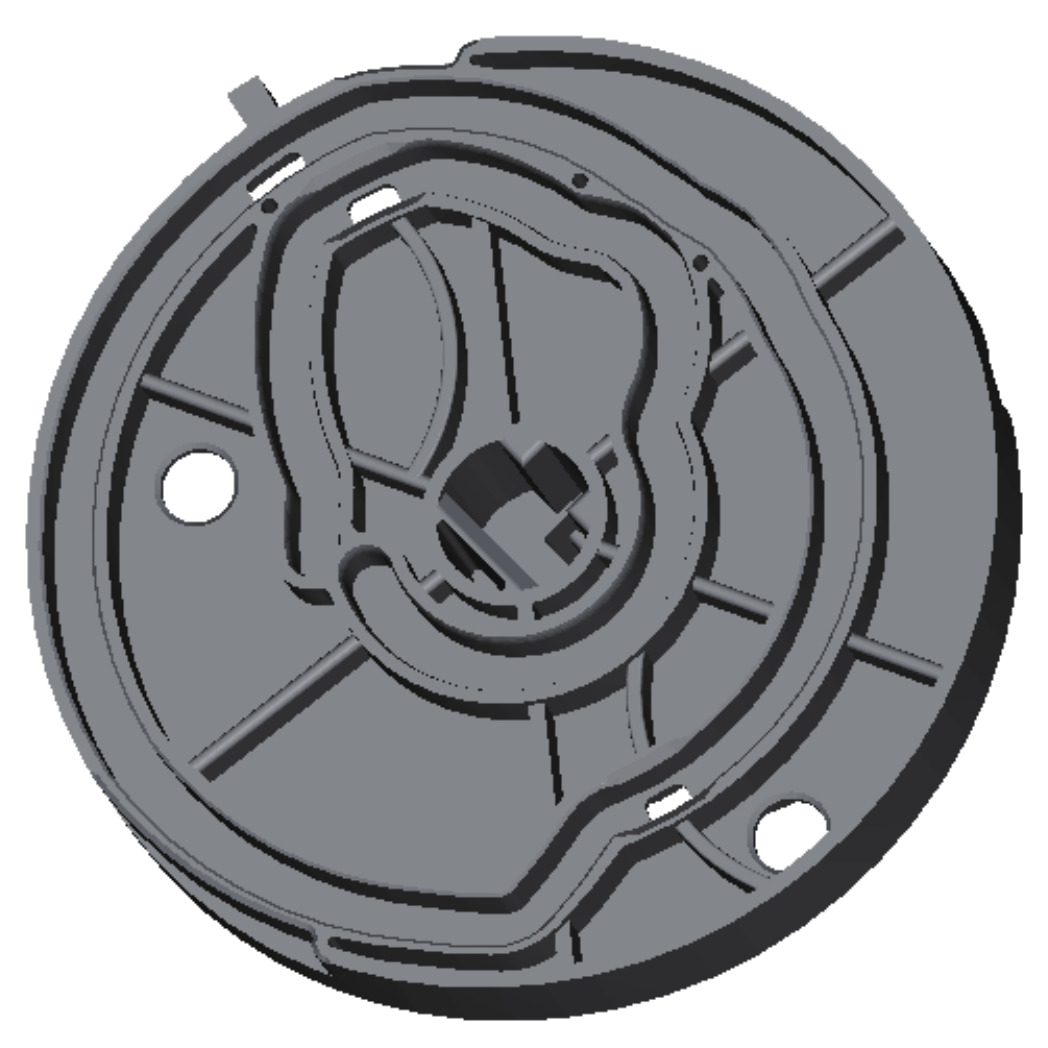}&
\includegraphics[width=0.14\columnwidth]{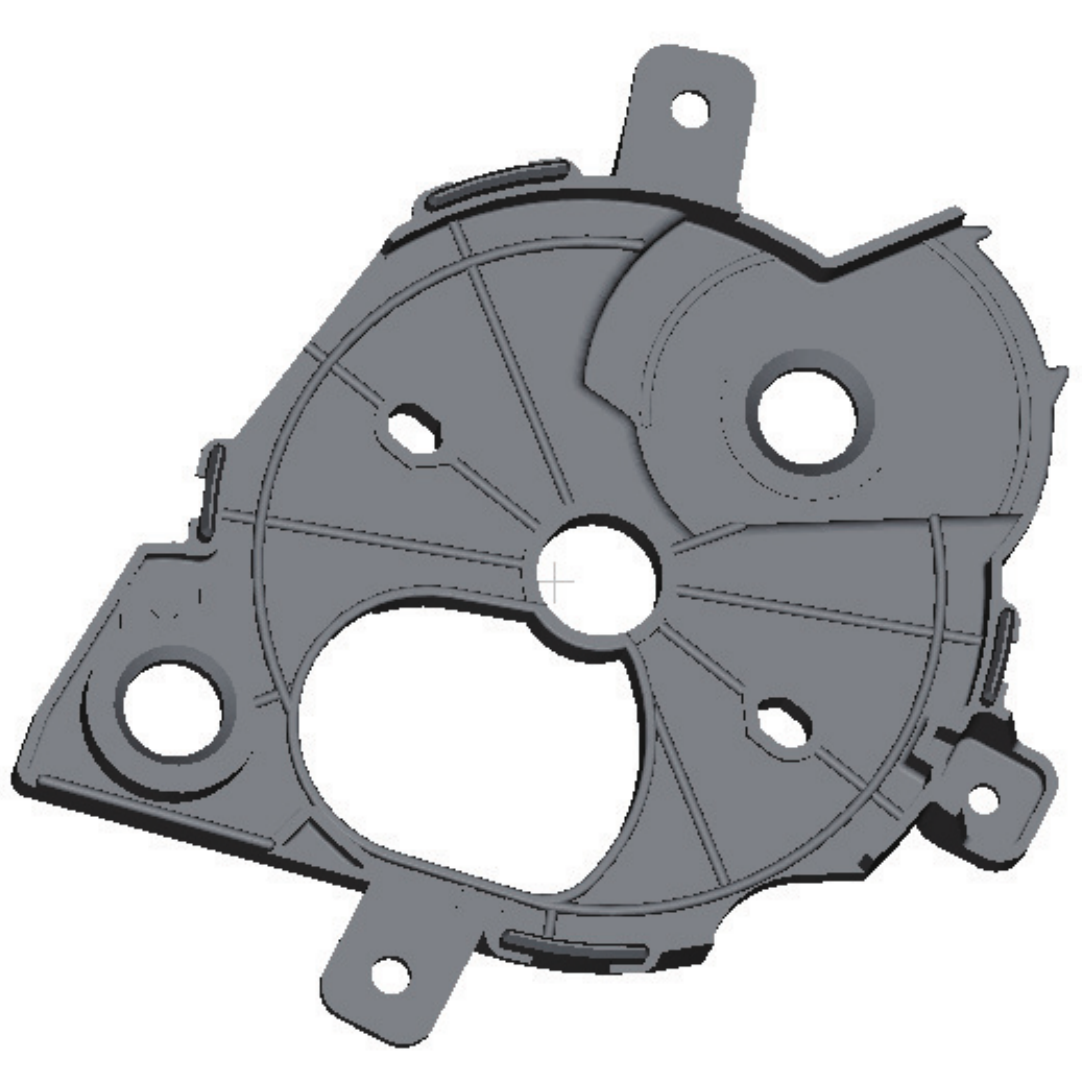}&
\includegraphics[width=0.14\columnwidth]{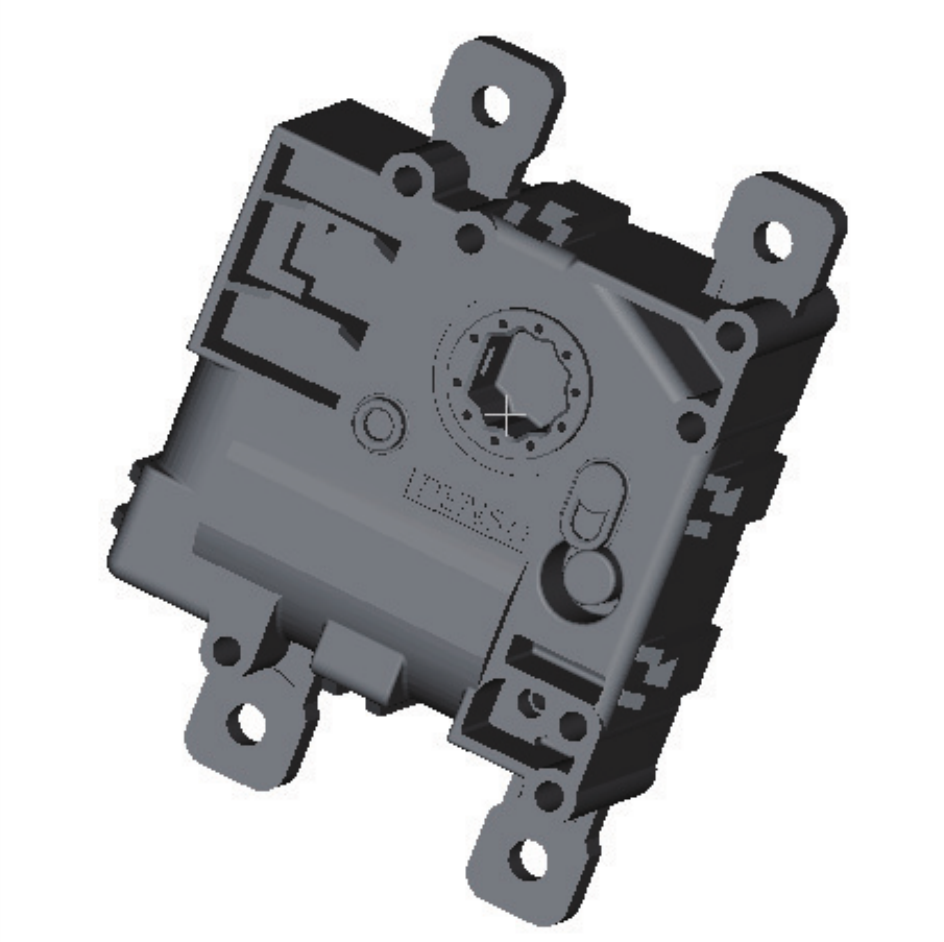}\\
Ring Screw&Gear Shaft&Object A&Object B&Object C\\
\end{tabular}
}
\caption{Models of objects used in the experiment. 
The gear shaft and ring screw are from the Fraunhofer IPA Bin-Picking dataset\cite{large-binpicking}, 
and Object A, B, C are from XA Bin-Picking dataset\cite{xu}.}
\label{fig: models}
\end{figure}
\begin{figure}[!t]
\centering
\subfigure[]{
\includegraphics[width=0.18\columnwidth]{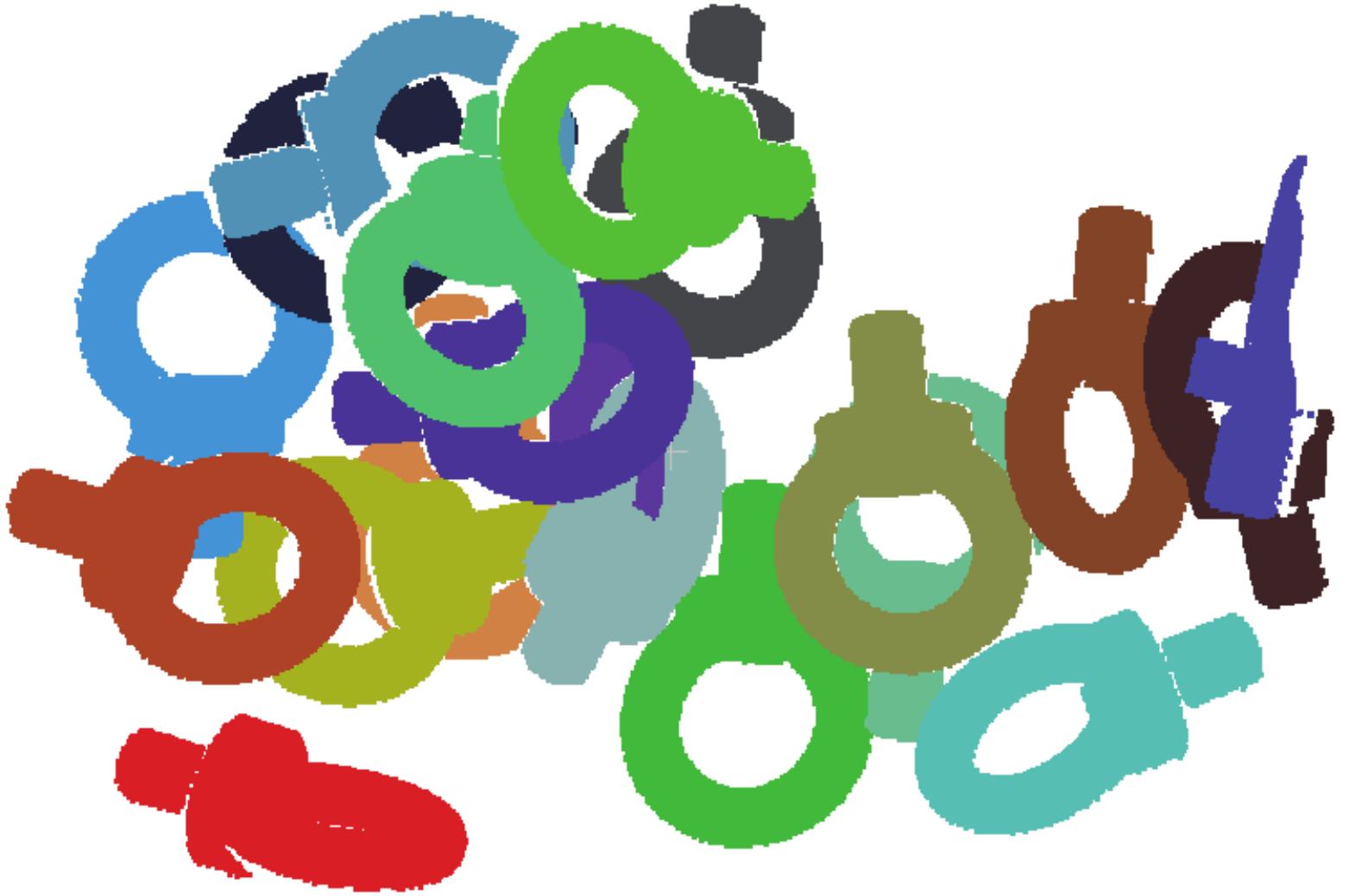}
\includegraphics[width=0.18\columnwidth]{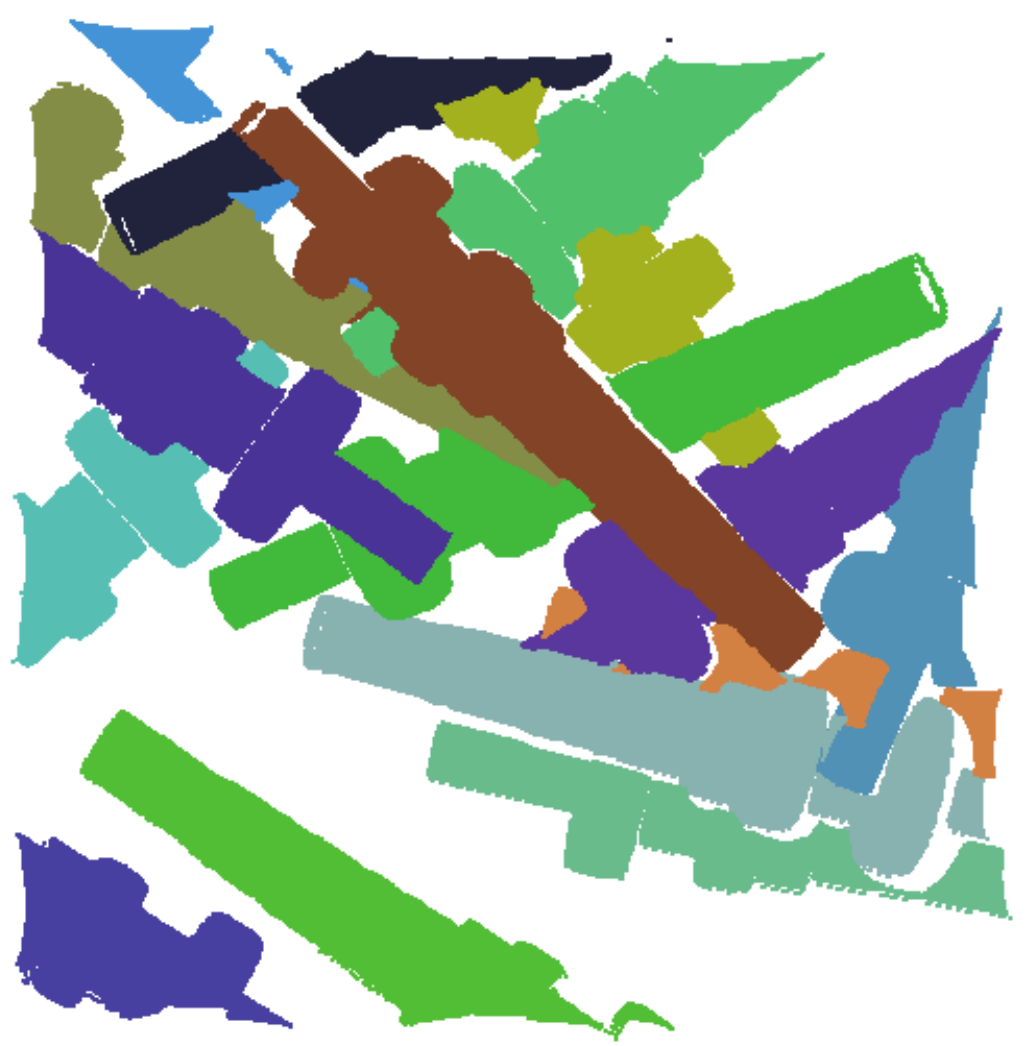}
\includegraphics[width=0.18\columnwidth]{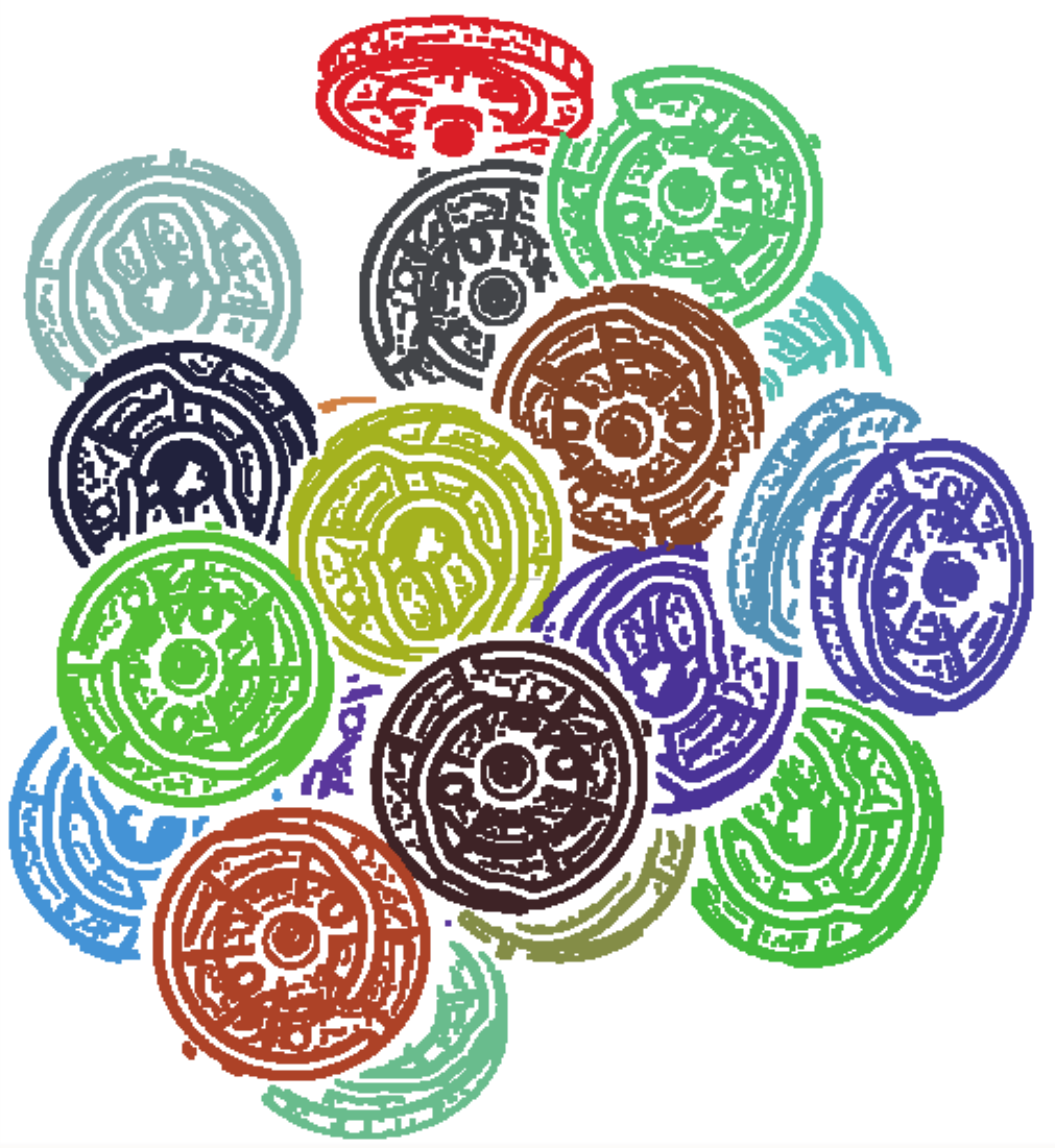}
\includegraphics[width=0.18\columnwidth]{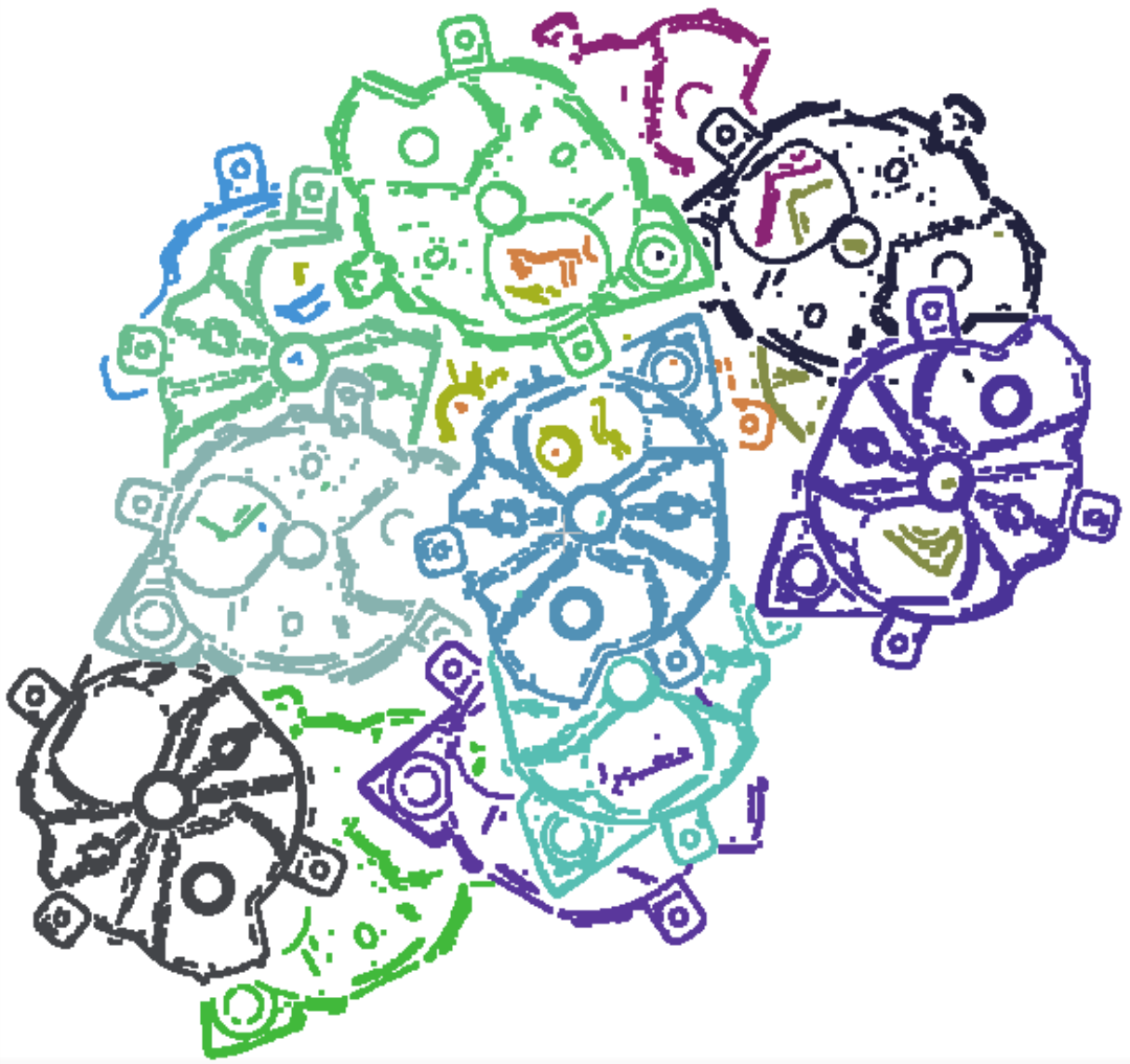}
\includegraphics[width=0.18\columnwidth]{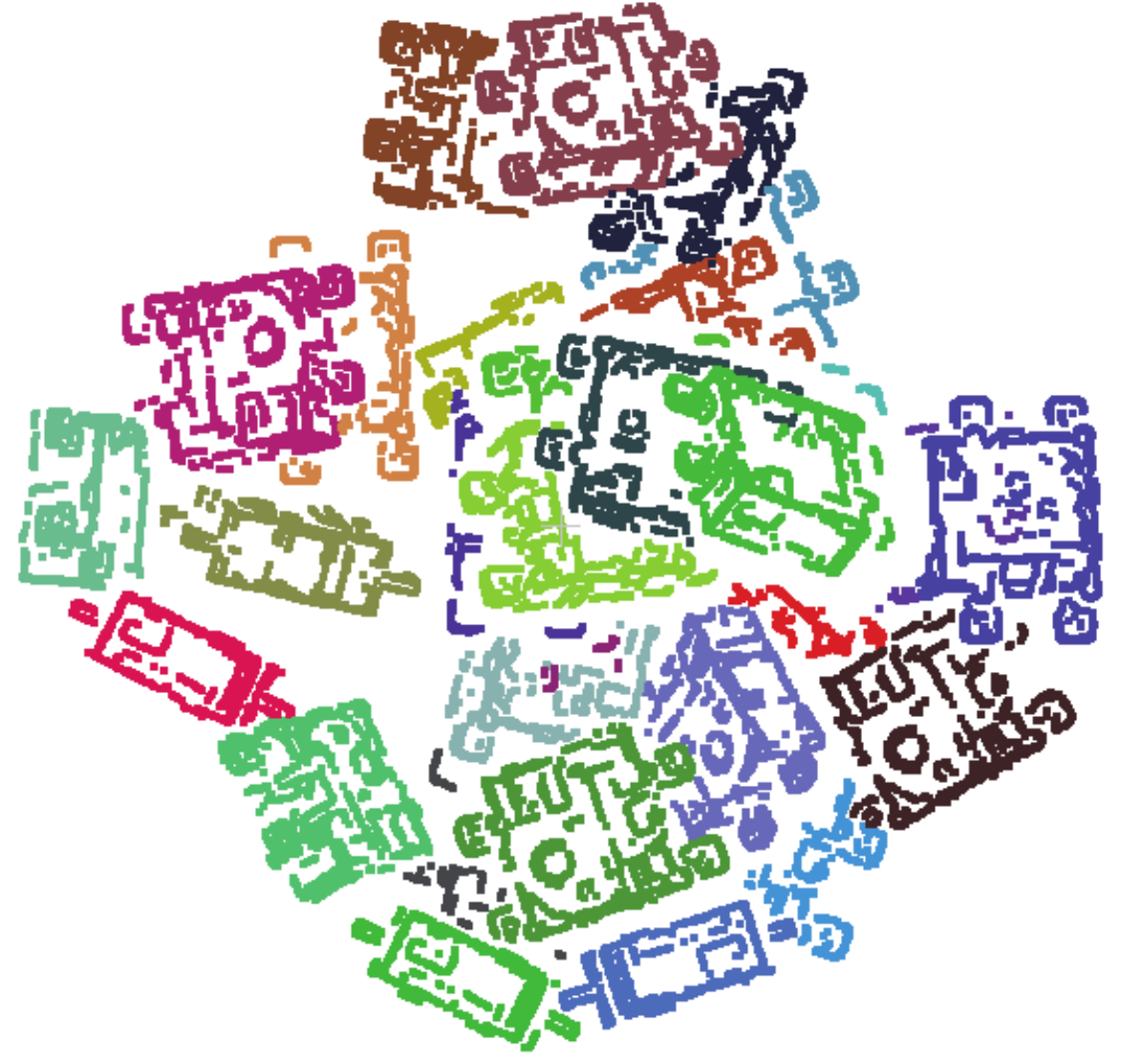}}

\subfigure[]{
\includegraphics[width=0.18\columnwidth]{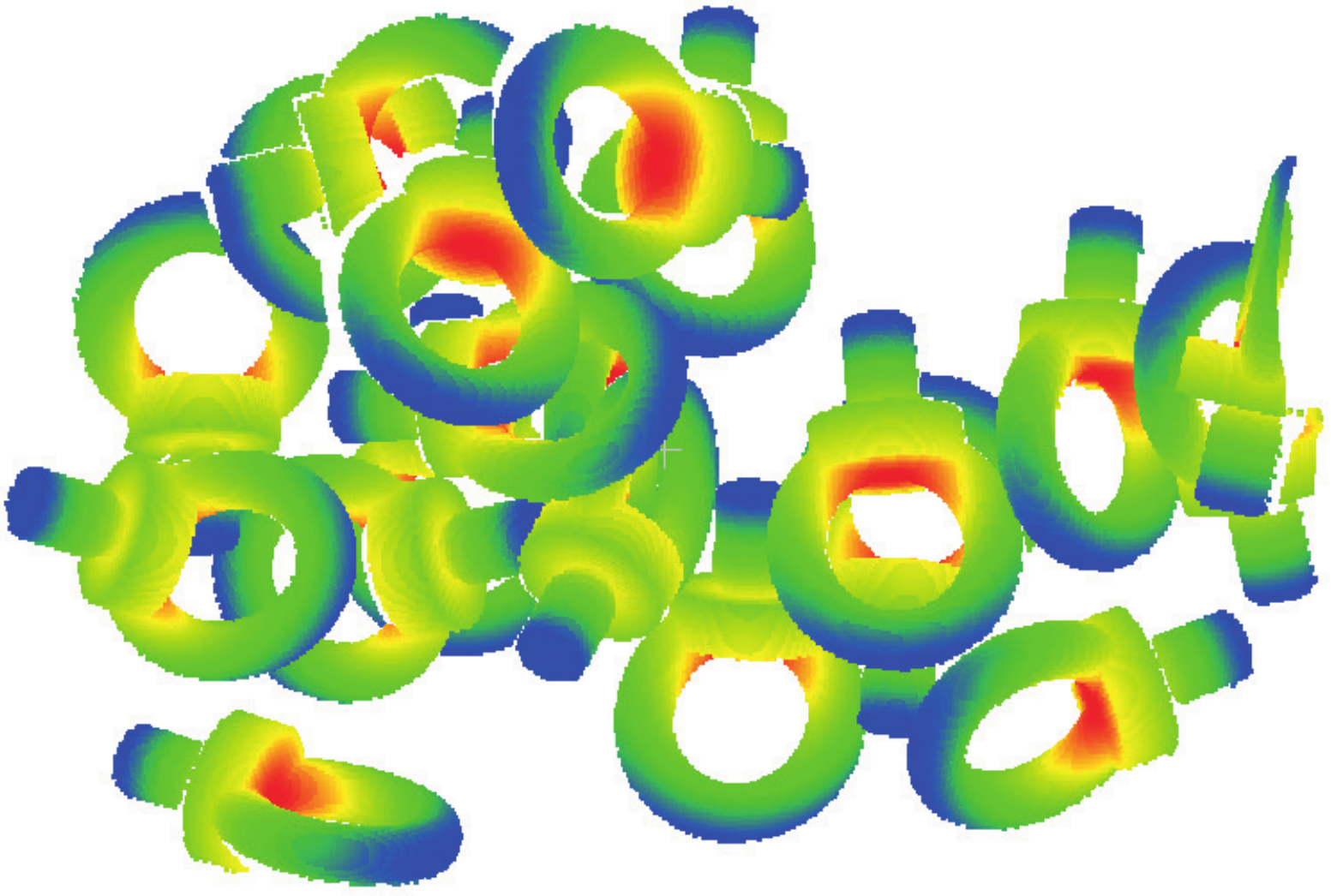}
\includegraphics[width=0.18\columnwidth]{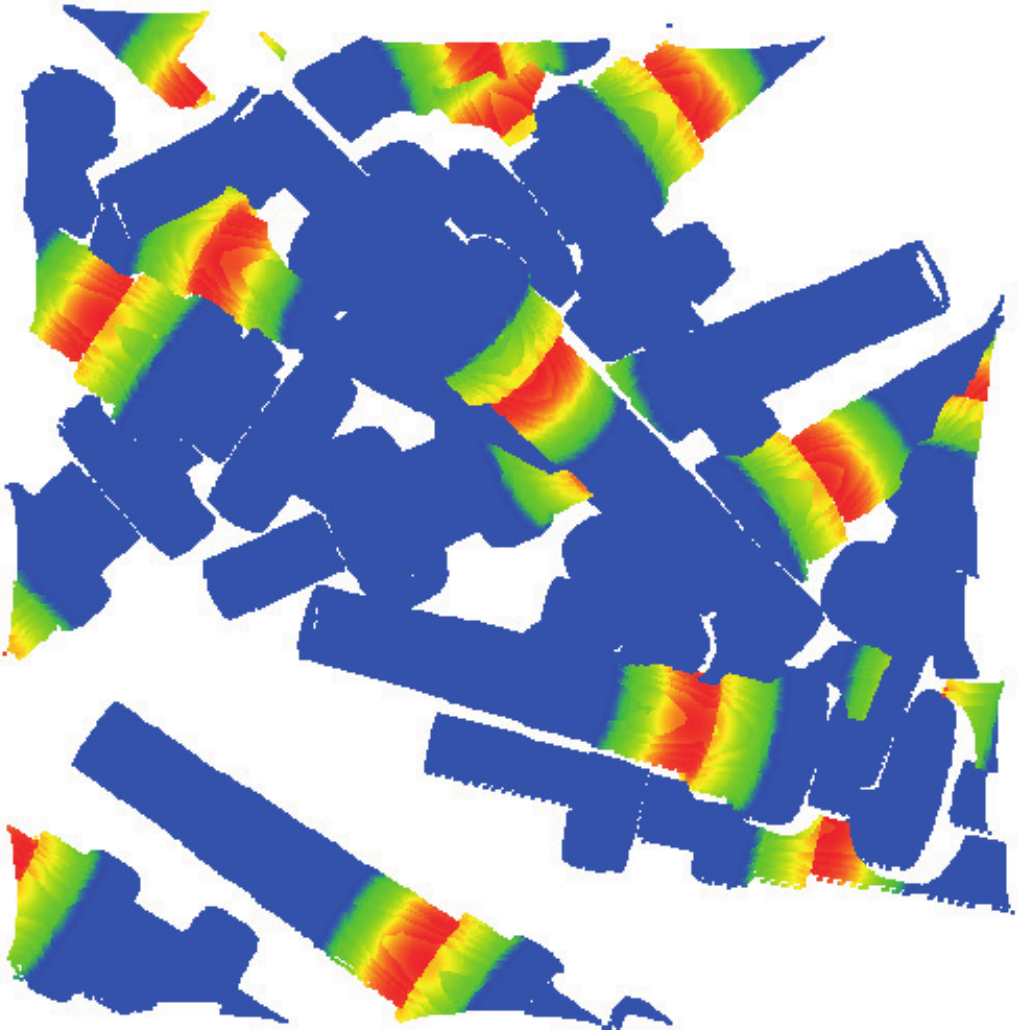}
\includegraphics[width=0.18\columnwidth]{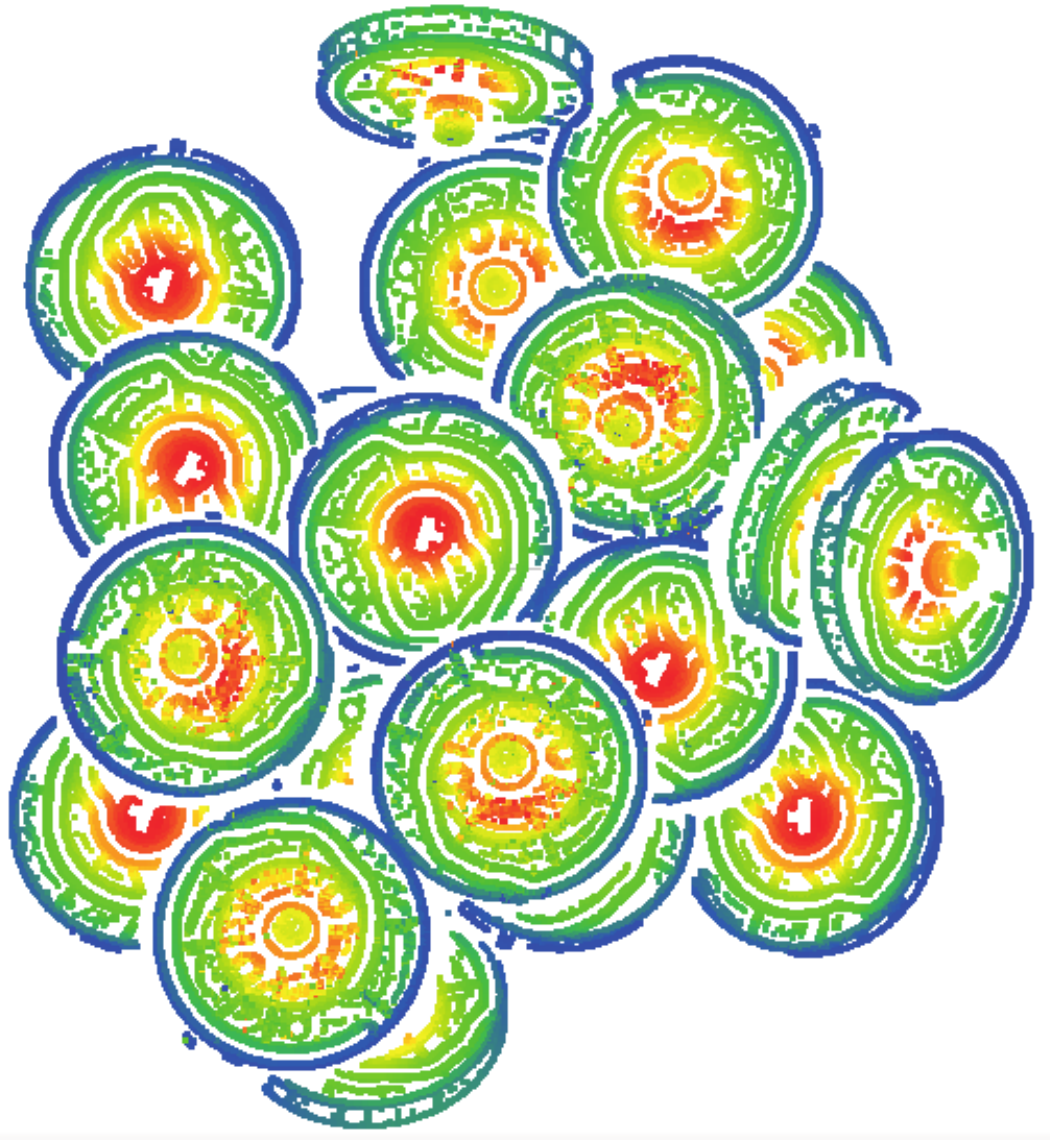}
\includegraphics[width=0.18\columnwidth]{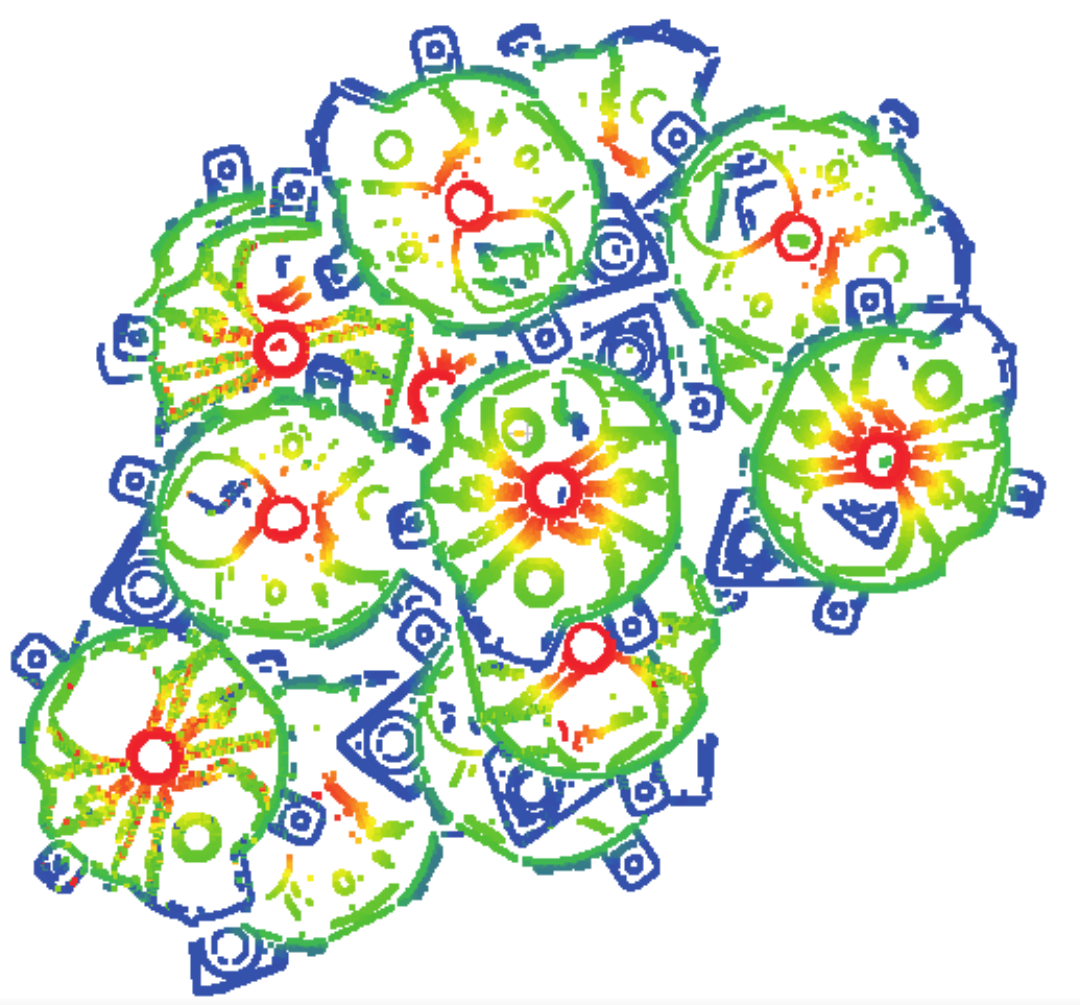}
\includegraphics[width=0.18\columnwidth]{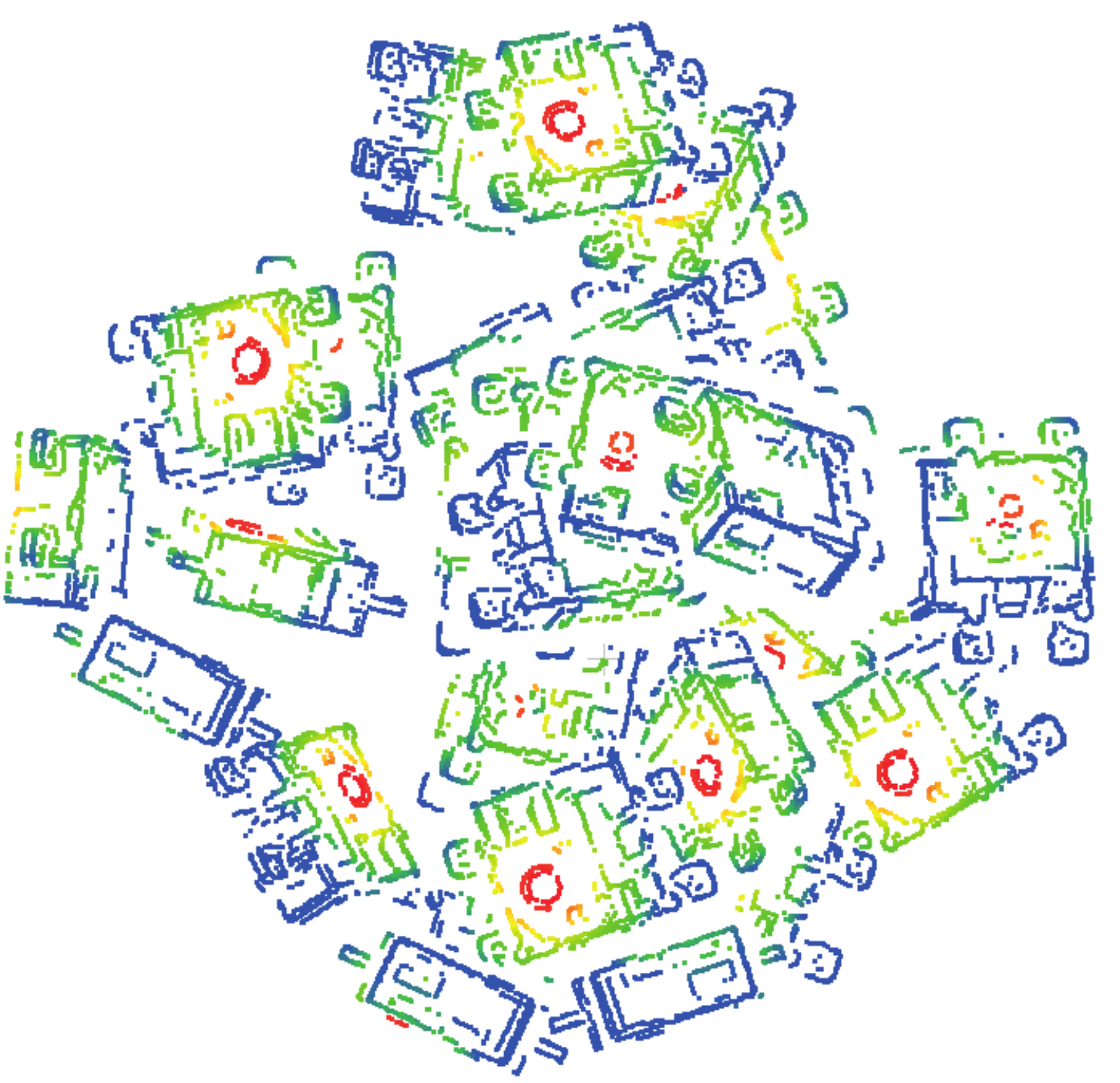}}
\caption{(a) Synthetic 3D point cloud scenes. 
(b) Center score computed by (\ref{eq:center}). 
}
\label{TrainSet}
\end{figure}

\subsubsection{Attention score matrix}
\label{ASM}
We introduce attention score matrix $S_A \in \mathbb{R}^{N\times N}$ to increase the weight of important point pairs. 
The $(i,j)$-th element of $S_A$ represents the weight of a point pair for point $i$ and $j$. 
Because the center point is used as the reference point for clustering in inference phase, the point pair closer to the center position should have a higher weight. 
In this paper, $s_{A(i,j)}$ is computed by
\begin{align}
\label{eq: asm}  
s_{A(i,j)} =  \mathrm{min}(1, s_{\mathrm{center}(i)} + s_{\mathrm{center}(j)}). 
\end{align}

\subsubsection{Embedded feature loss}
\label{Lef}

 
A point pair $(p_i, p_j)$ has two possible relationships as follows: 1) $p_i$ and $p_j$ belong to the same instance; and 2) $p_i$ and $p_j$ belong to different instances. 
By considering this, embedded feature loss $L_{\rm{EF}}$ is defined by
\begin{equation}
\label{ef_loss}
L_{\mathrm{EF}} = \sum_{i}^{N} \sum_{j}^{N} w_{(i,j)} \kappa_{(i,j)},
\end{equation}
where $w_{(i,j)}$ is a element of $W \in \mathbb{R}^{N\times N}$ which is a weight matrix obtained through element-wise multiplying $D_V$ by $D_A$, that is,
\begin{align}
w_{(i,j)} = d_{V(i,j)} s_{A(i,j)}.
\end{align}
In (\ref{ef_loss}), $\kappa_{(i,j)}$ is the loss based on the relationships of point pair and it is defined as:
\begin{scriptsize}
\begin{align}
\kappa_{(i,j)} =
\left\{
             \begin{array}{lr}
                          \mathrm{max}(0, d_{F(i,j)} - \epsilon_1) & \mathrm{if}\, p_i\, \mathrm{and\,} p_j\, \mathrm{in\, the\,same\,instance} \\
             \mathrm{max}(0, \epsilon_2 - d_{F(i,j)}) & \mathrm{otherwise}  
             \end{array}
\right. 
\label{eq:kappa}
\end{align}
\end{scriptsize}
where $\epsilon_1$, $\epsilon_2$ are constants 
and set to satisfy the condition $0 < \epsilon_1 < \epsilon_2$, because the feature distance of point pairs in different instances should be greater than those belonging to the same instance\cite{SGPN}. 
We do not need to make the feature distance for point pairs in the same instance close to zero but smaller than the threshold $\epsilon_1$, which is helpful for learning\cite{xu}.

\subsubsection{Center score loss}
\label{Lcs}
Smooth $L1$ loss is used as a loss function for the center score branch 
because of robustness of L1 loss function\cite{Fast-Rcnn}. 
The center score loss $L_{\rm{CS}}$ is defined by
\begin{equation}
L_{\mathrm{CS}} = \frac{1}{N} \sum_{i}^{N} \mathrm{smooth}_{L1}(s_{\mathrm{center}(i)}-\widehat{s}_{\mathrm{center}(i)}),
\end{equation}
where $\widehat{s}_{\mathrm{center}(i)}$ represents the predicted center score and  
\begin{align}
\mathrm{smooth}_{L1}(x) =
\left\{
             \begin{array}{lr}
                          0.5 \left|x\right|^2 & \mathrm{if}\, \left|x\right| < 1\\
             \left|x\right| -0.5 & \mathrm{otherwise}
             \end{array}
\right.
\end{align}

\begin{figure*}[!t]
\subfigure{
\begin{tabular}{ccccccc}
\includegraphics[width=0.24\columnwidth]{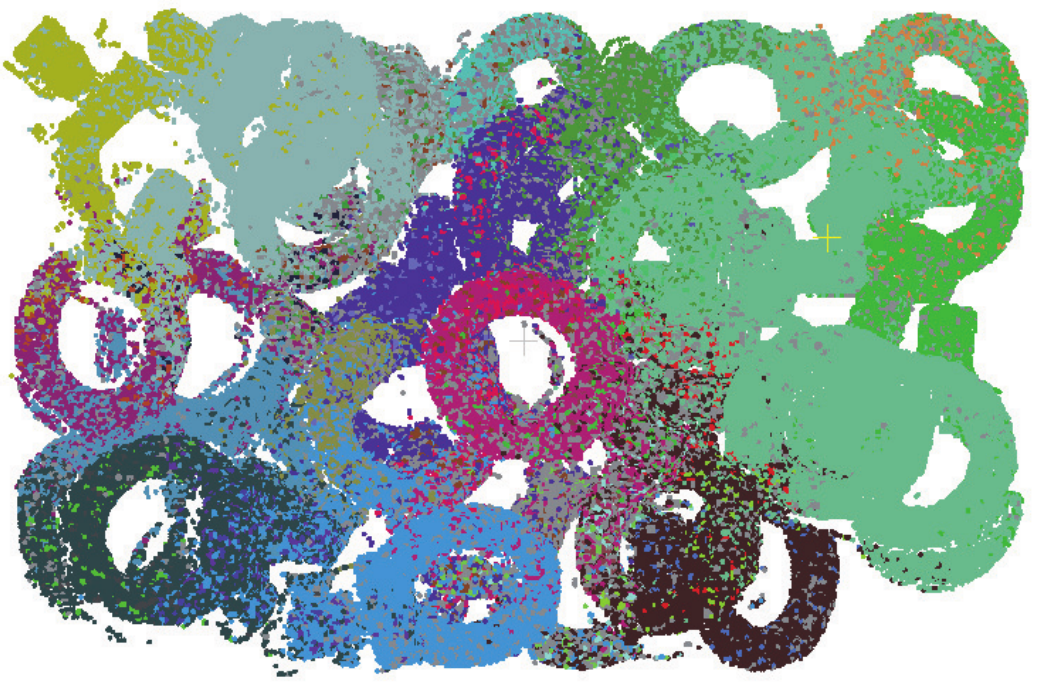}&
\includegraphics[width=0.24\columnwidth]{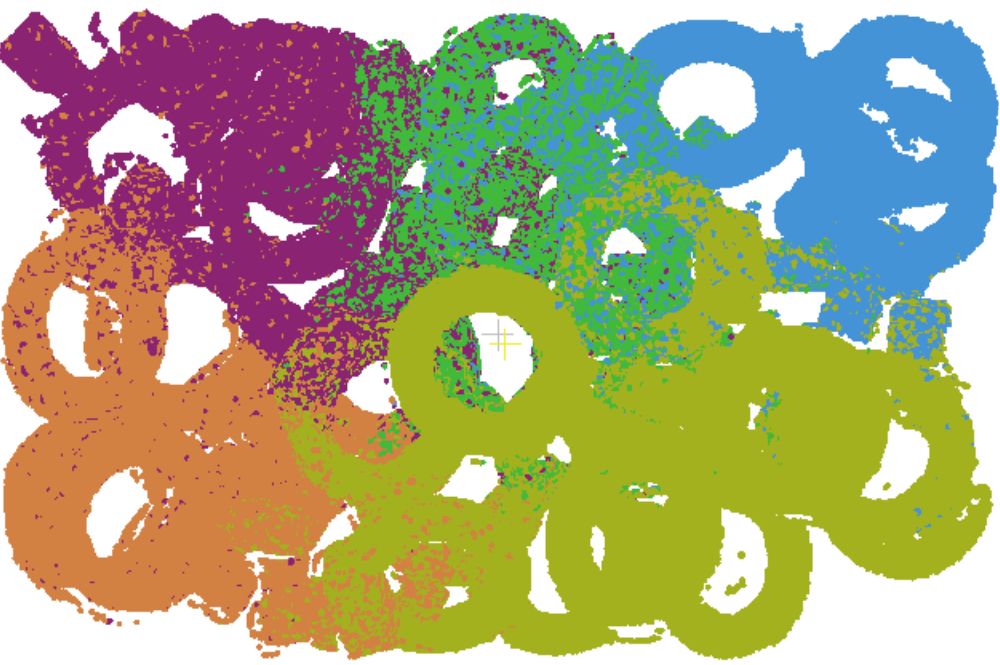}&
\includegraphics[width=0.24\columnwidth]{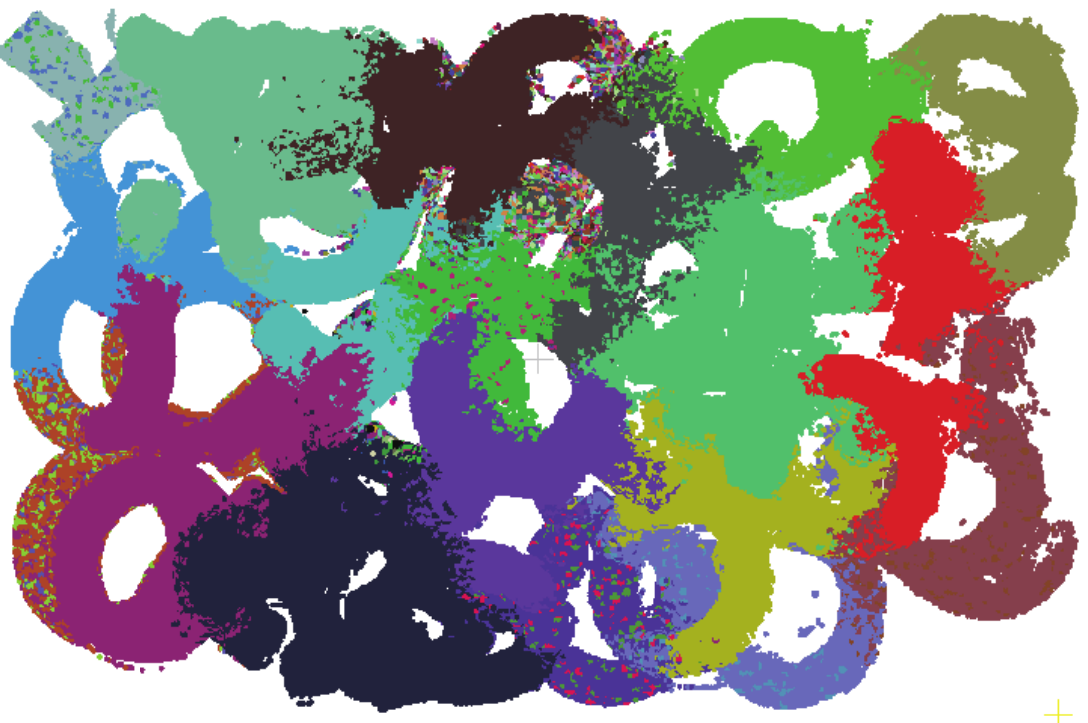}&
\includegraphics[width=0.24\columnwidth]{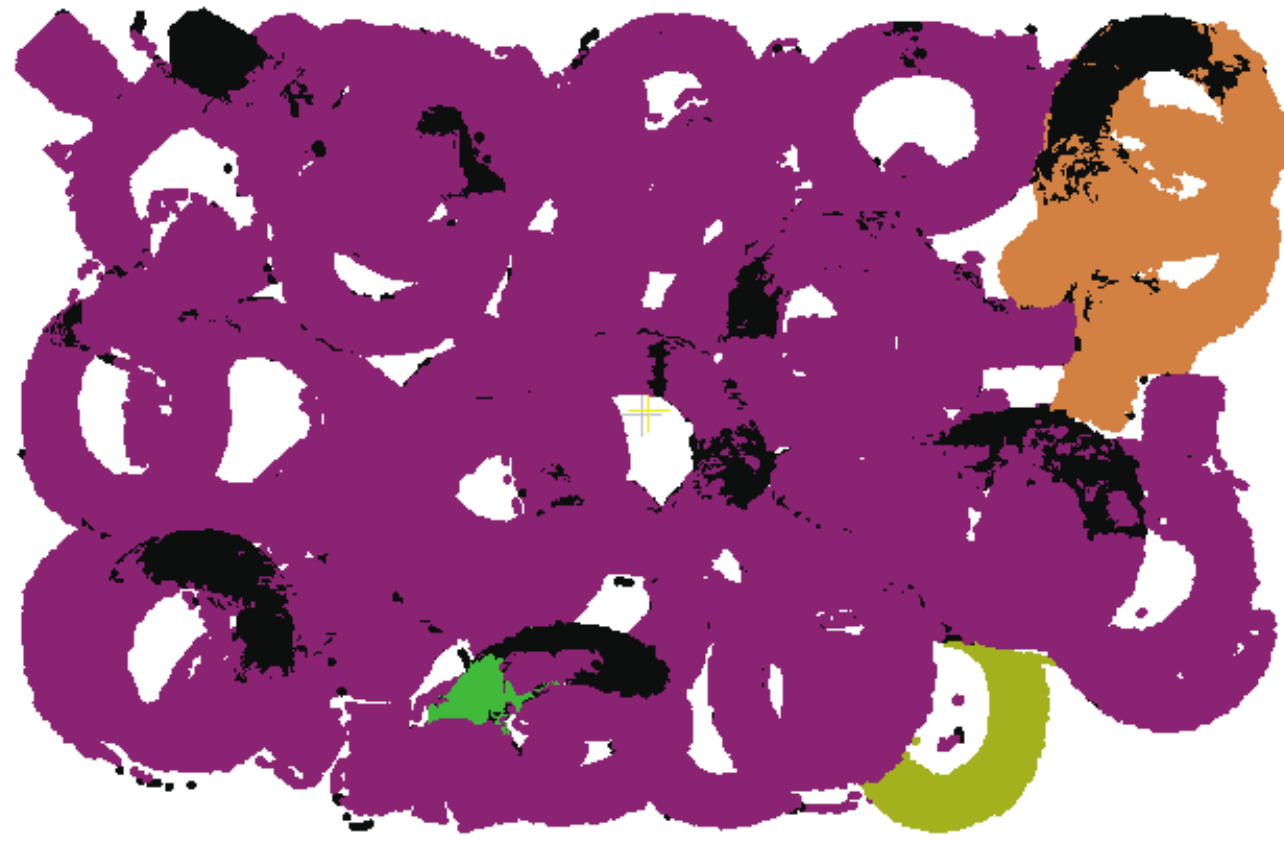}&
\includegraphics[width=0.24\columnwidth]{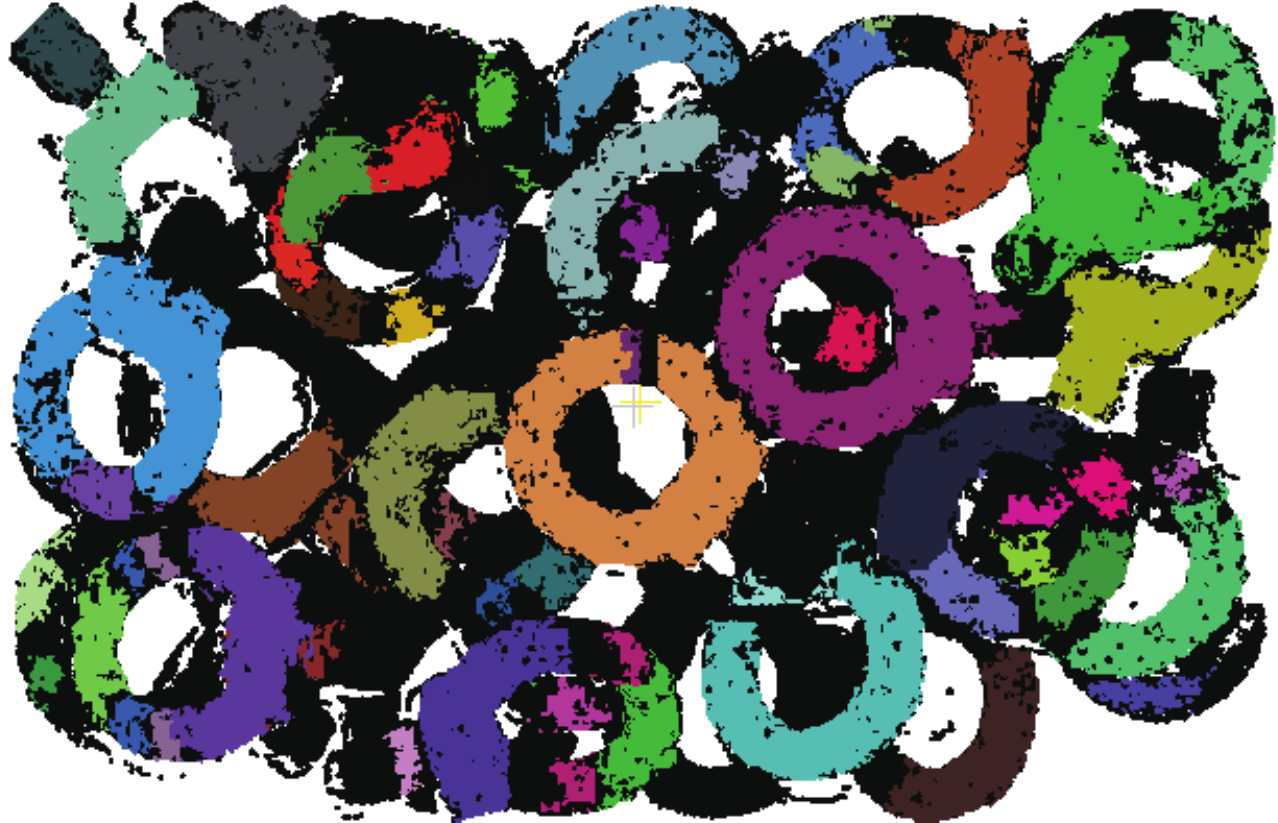}&
\includegraphics[width=0.24\columnwidth]{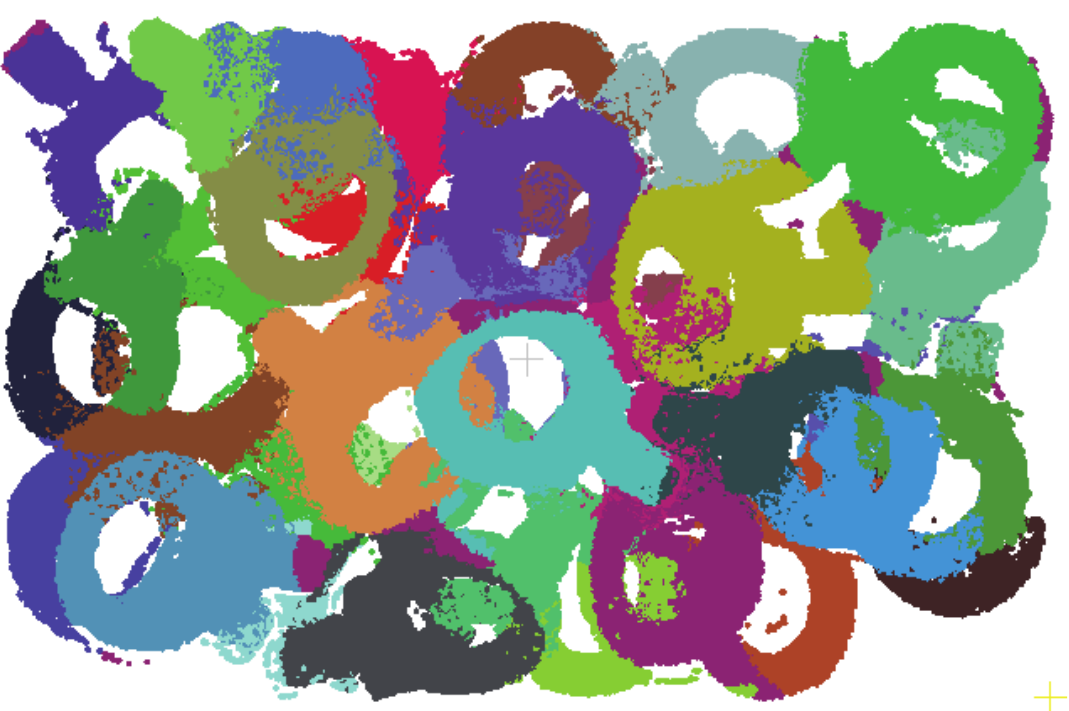}&
\includegraphics[width=0.24\columnwidth]{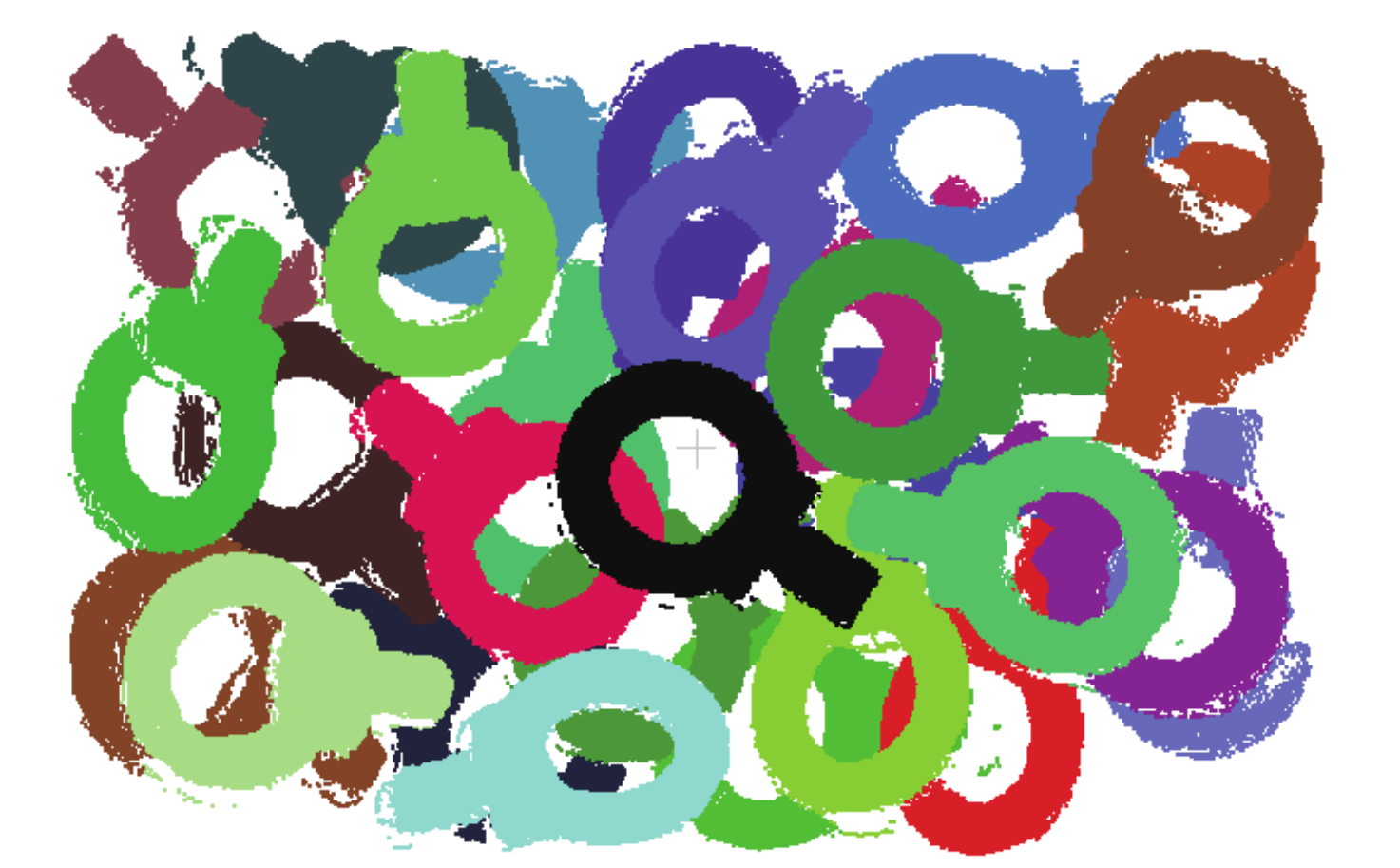}\\
\end{tabular}
}
\subfigure{
\begin{tabular}{ccccccc}
\includegraphics[width=0.24\columnwidth]{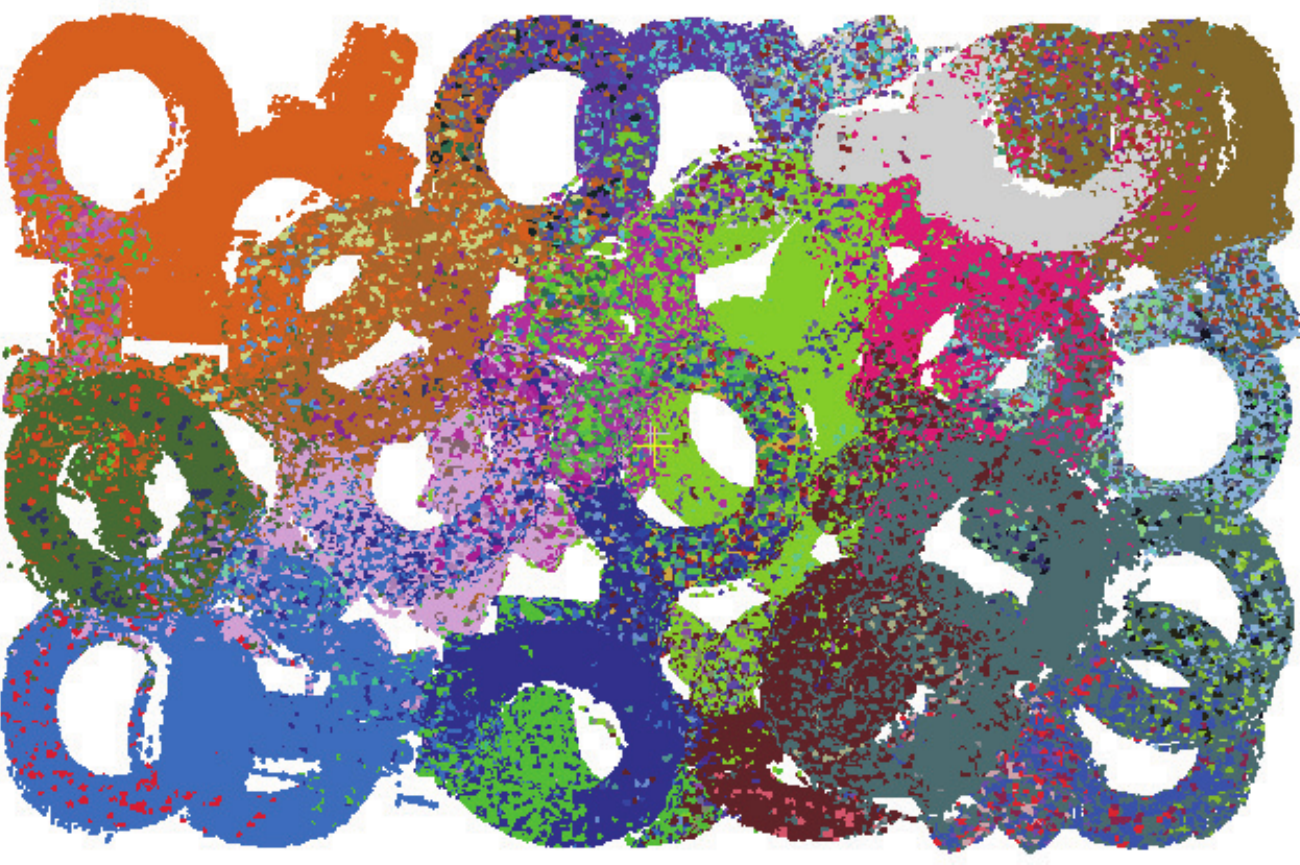}&
\includegraphics[width=0.24\columnwidth]{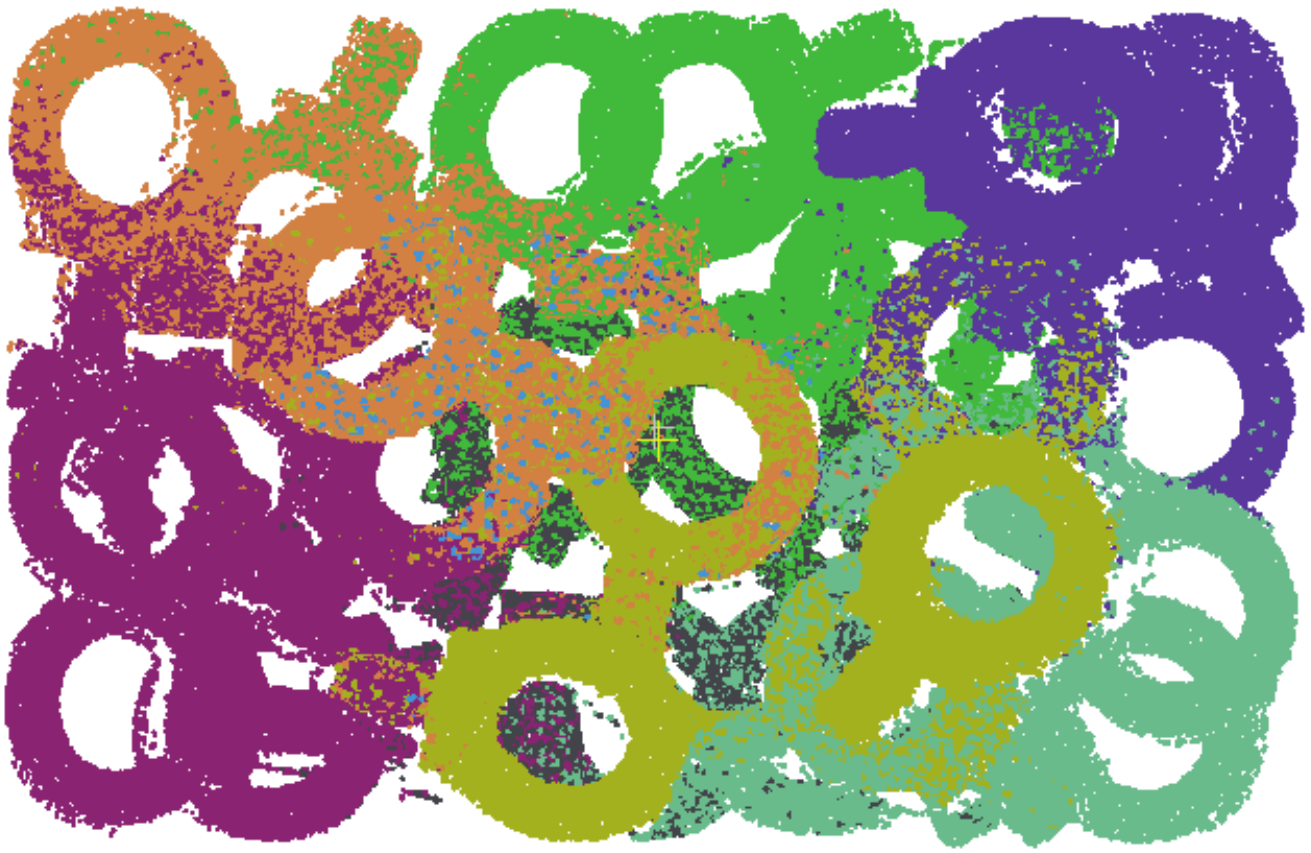}&
\includegraphics[width=0.24\columnwidth]{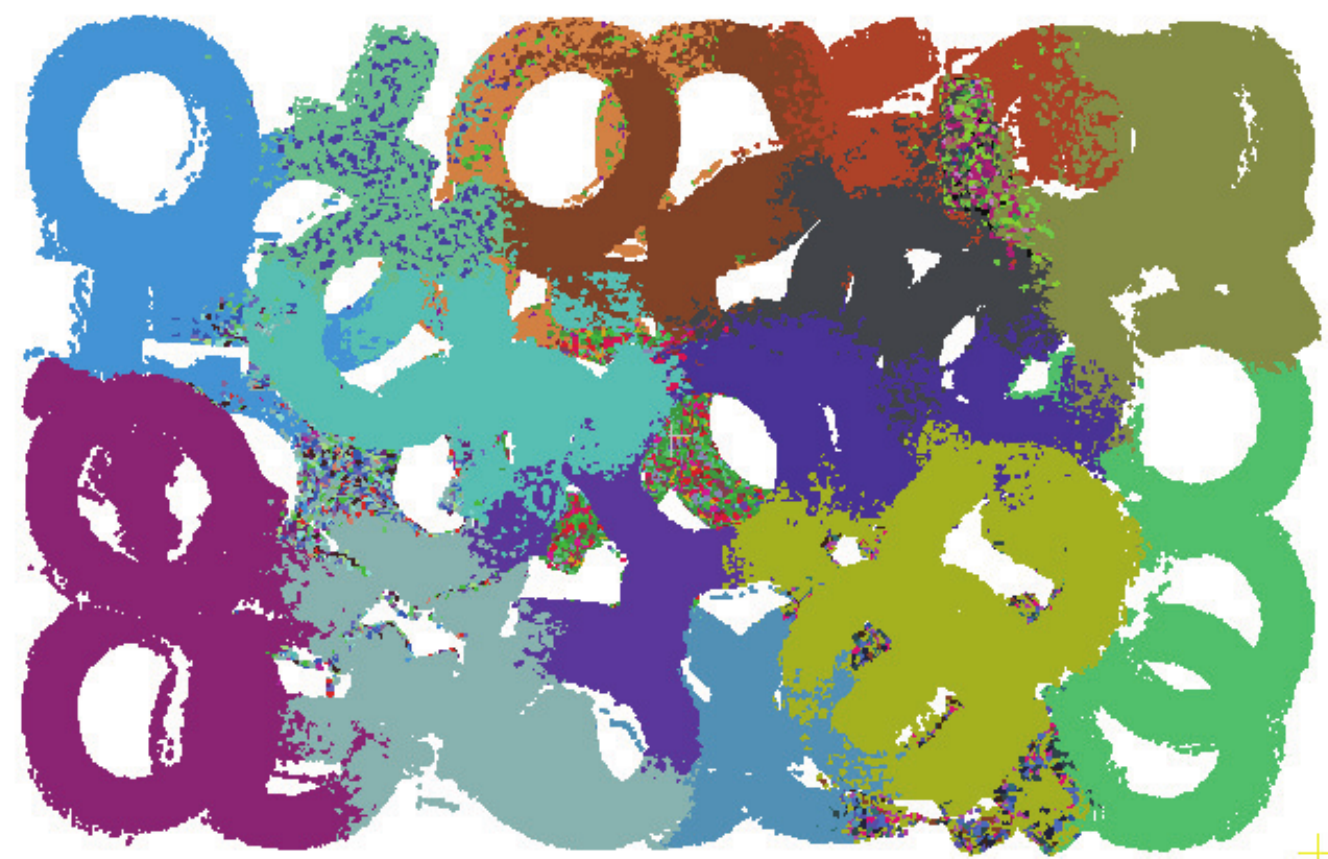}&
\includegraphics[width=0.24\columnwidth]{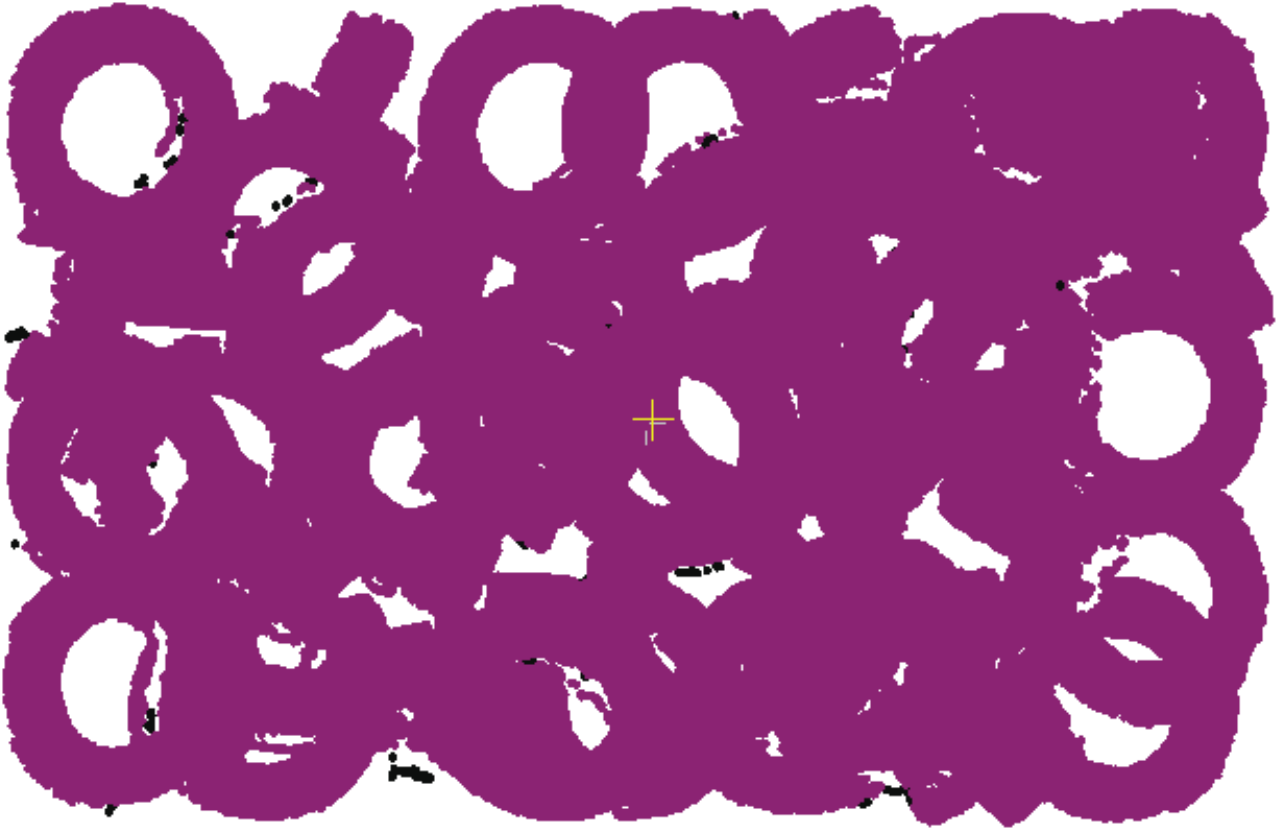}&
\includegraphics[width=0.24\columnwidth]{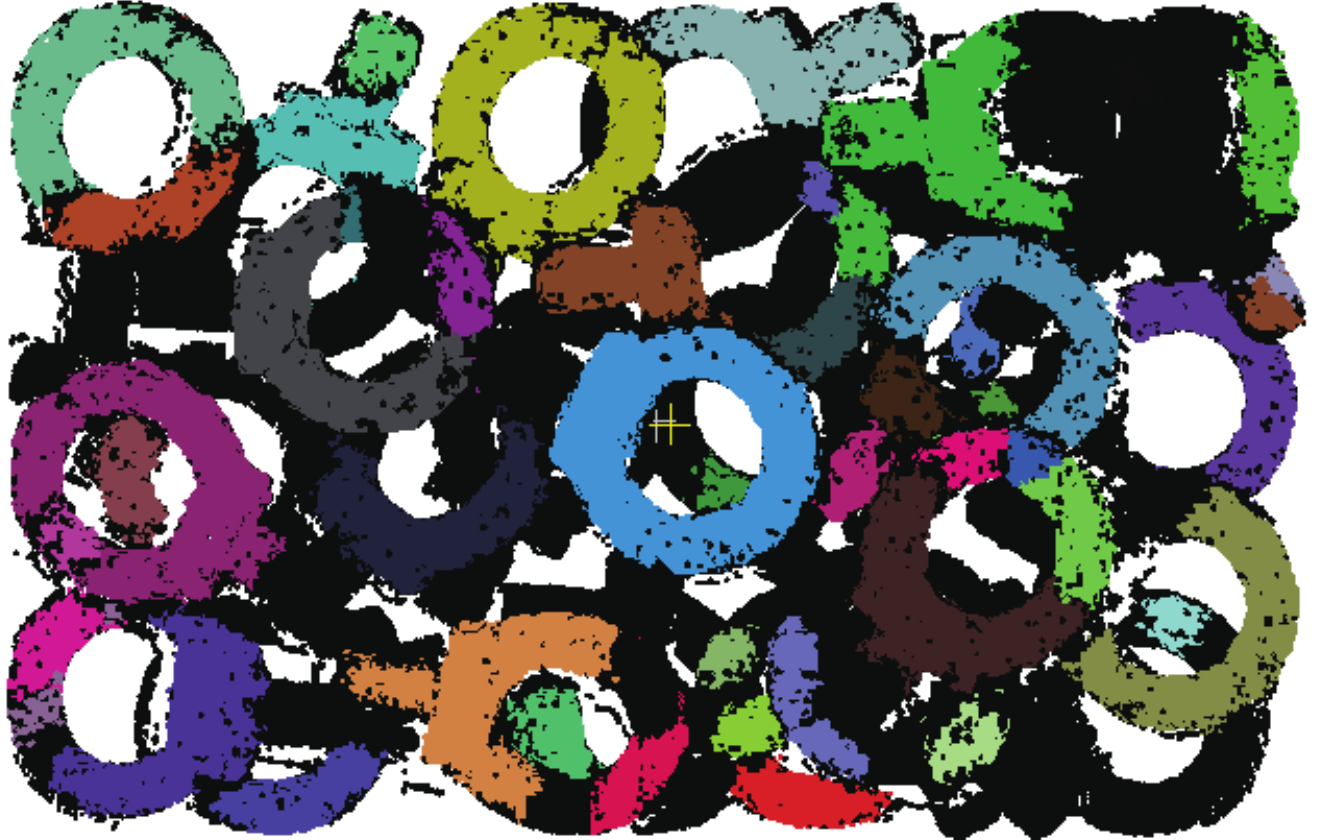}&
\includegraphics[width=0.24\columnwidth]{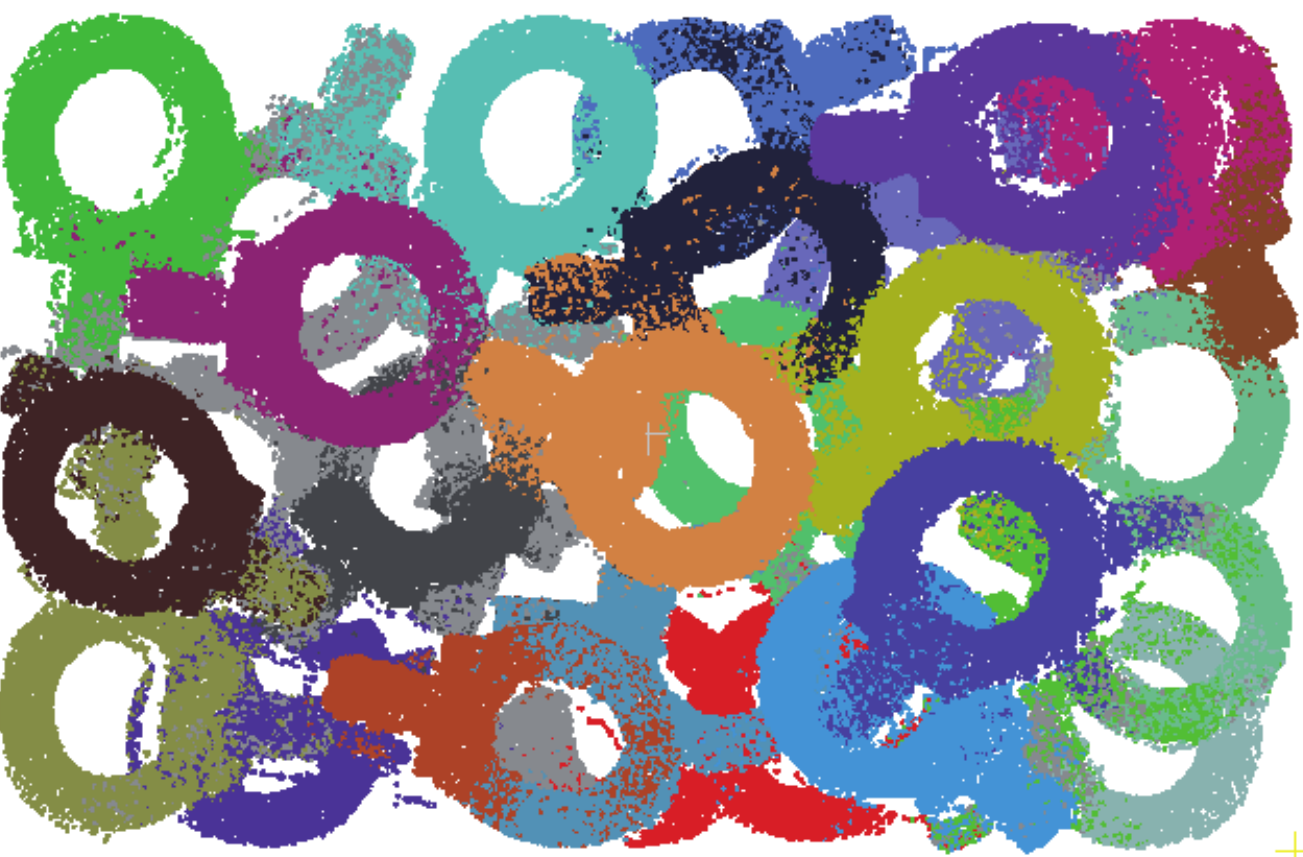}&
\includegraphics[width=0.24\columnwidth]{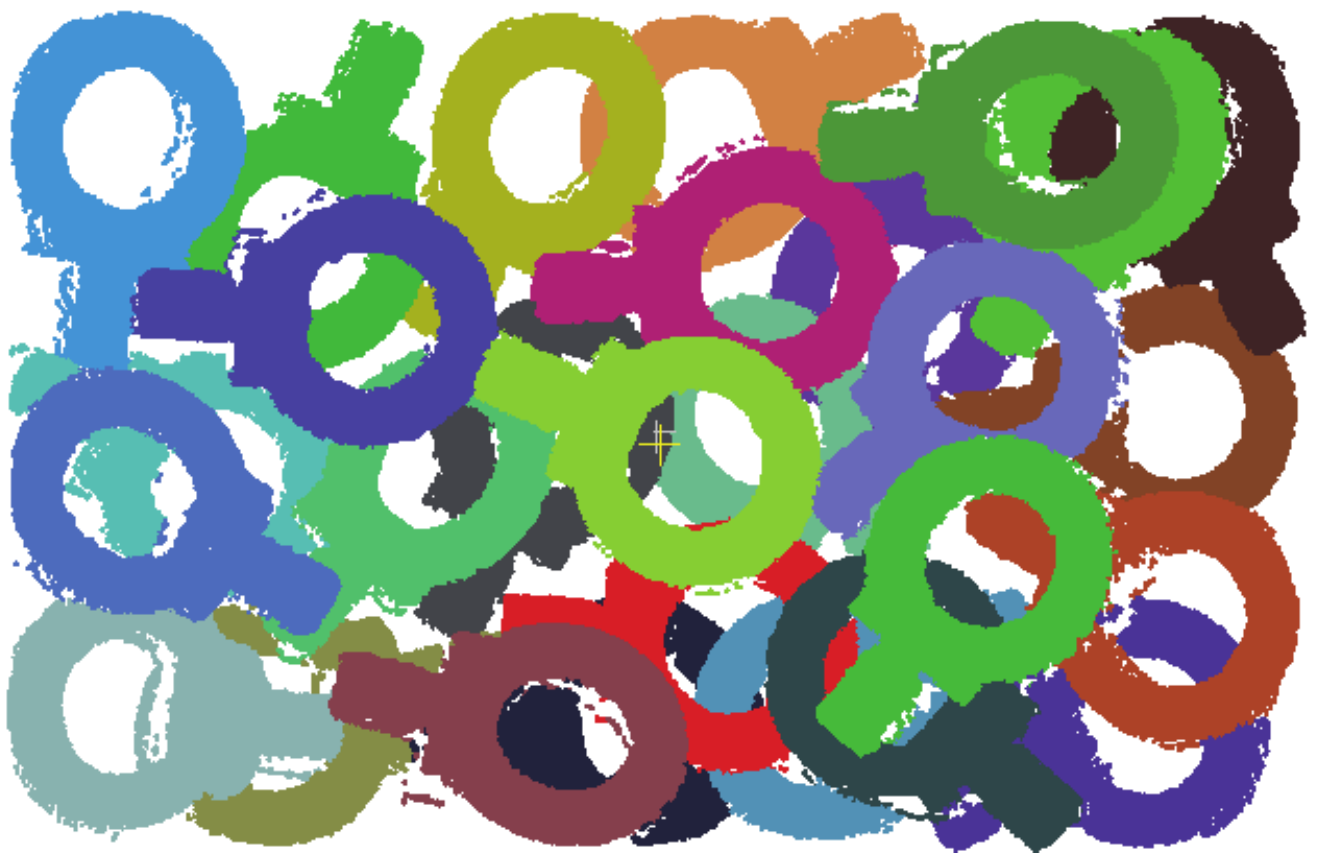}\\
\end{tabular}
}
\subfigure{
\begin{tabular}{ccccccc}
\includegraphics[width=0.24\columnwidth]{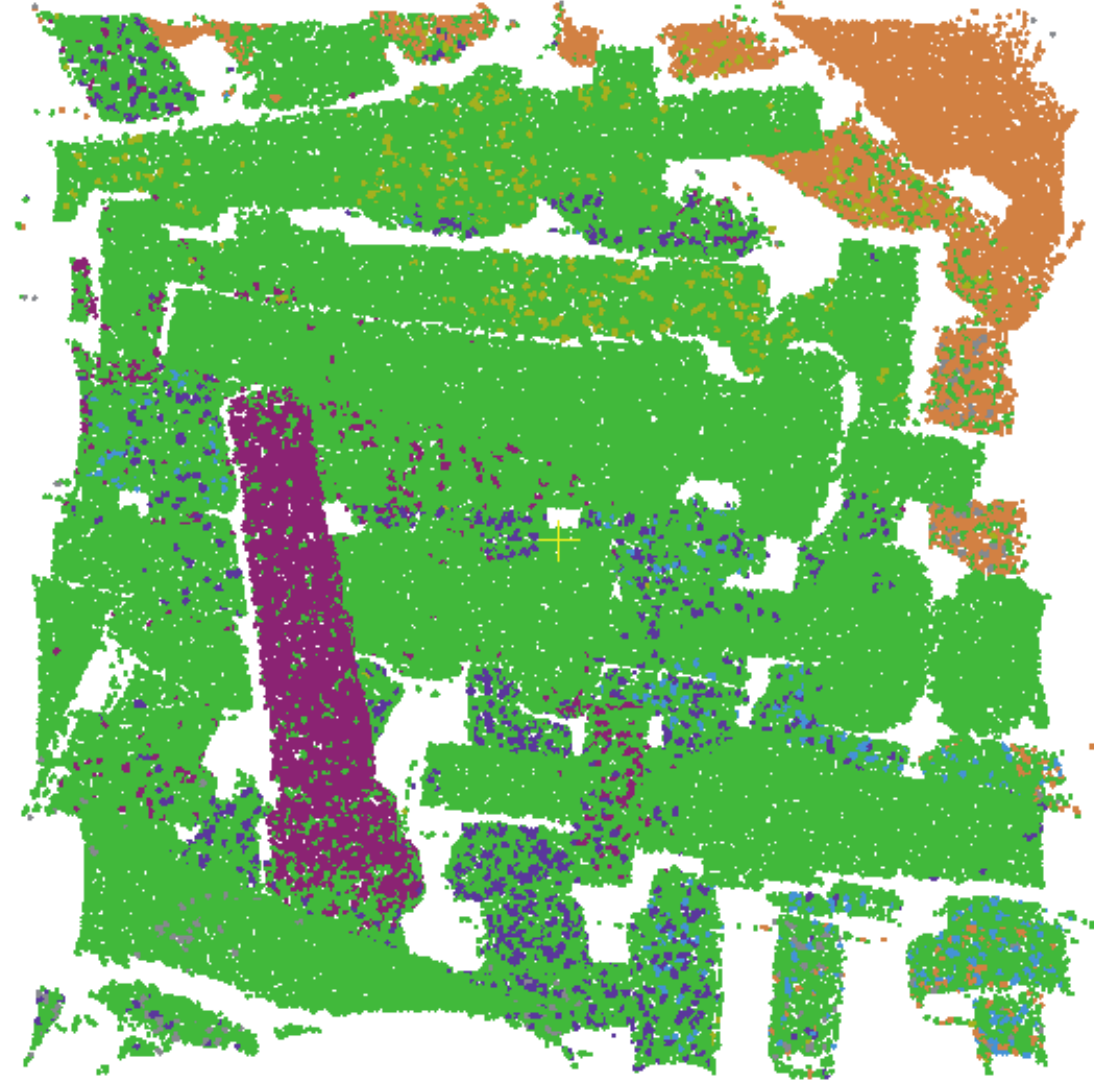}&
\includegraphics[width=0.24\columnwidth]{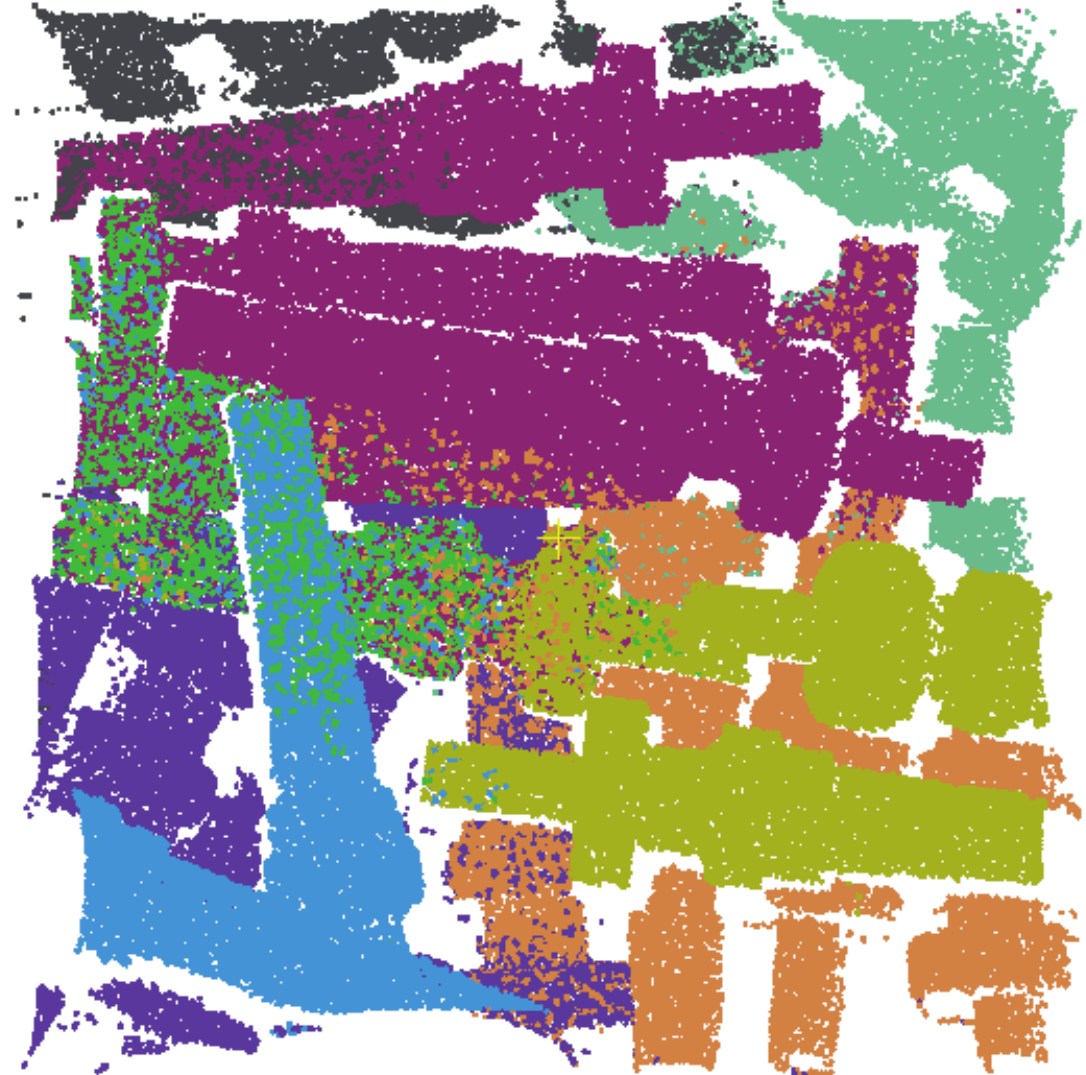}&
\includegraphics[width=0.24\columnwidth]{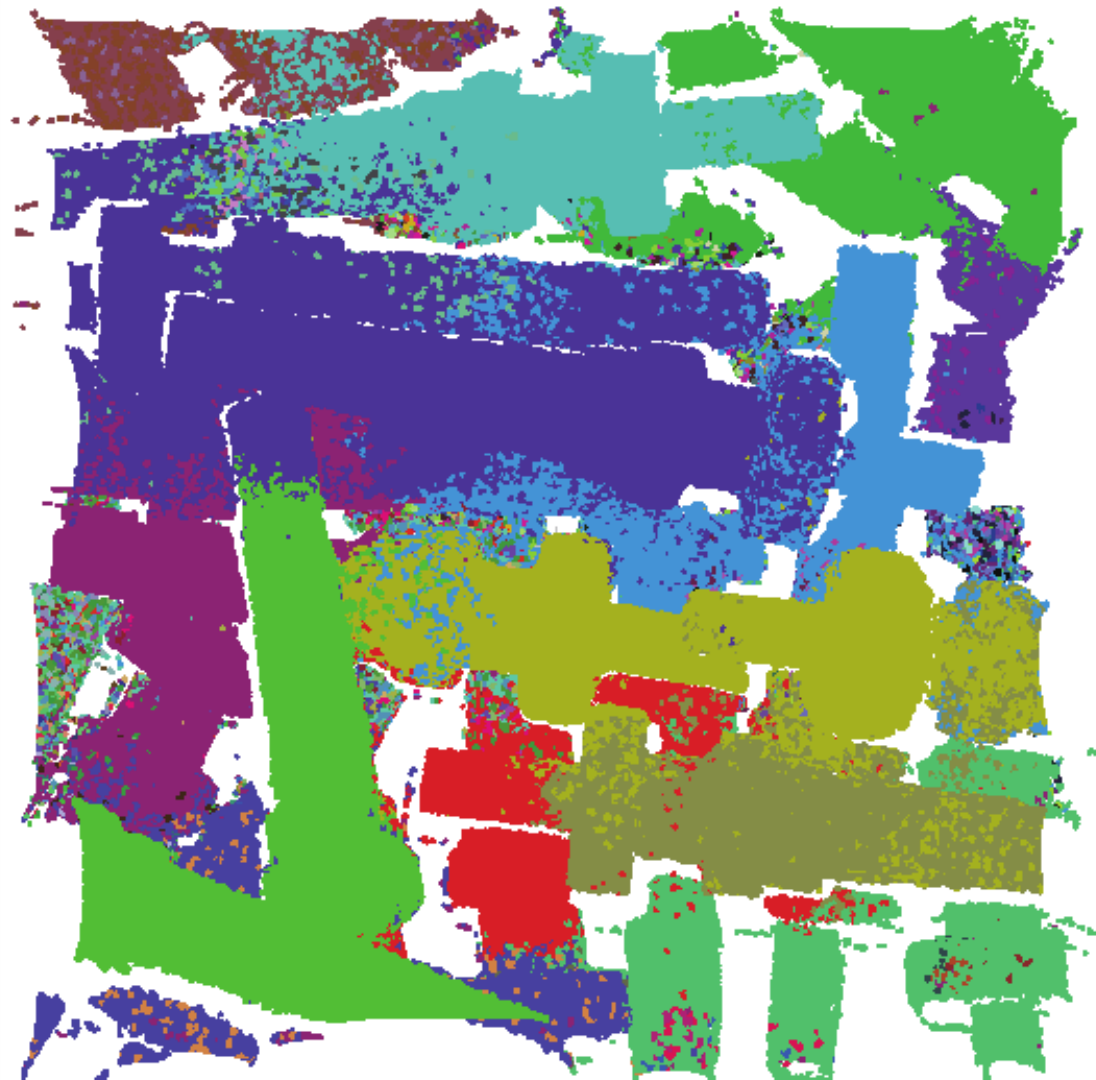}&
\includegraphics[width=0.24\columnwidth]{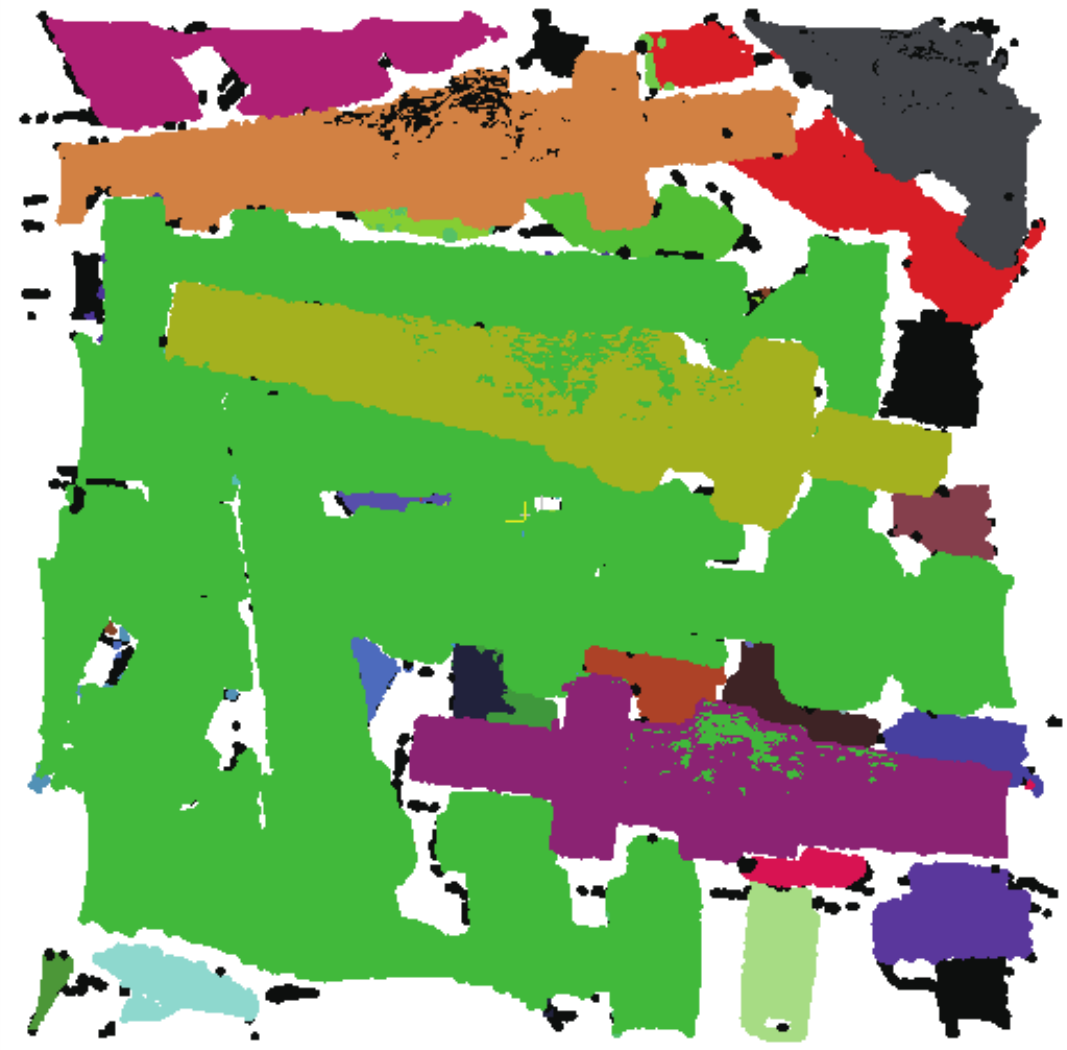}&
\includegraphics[width=0.24\columnwidth]{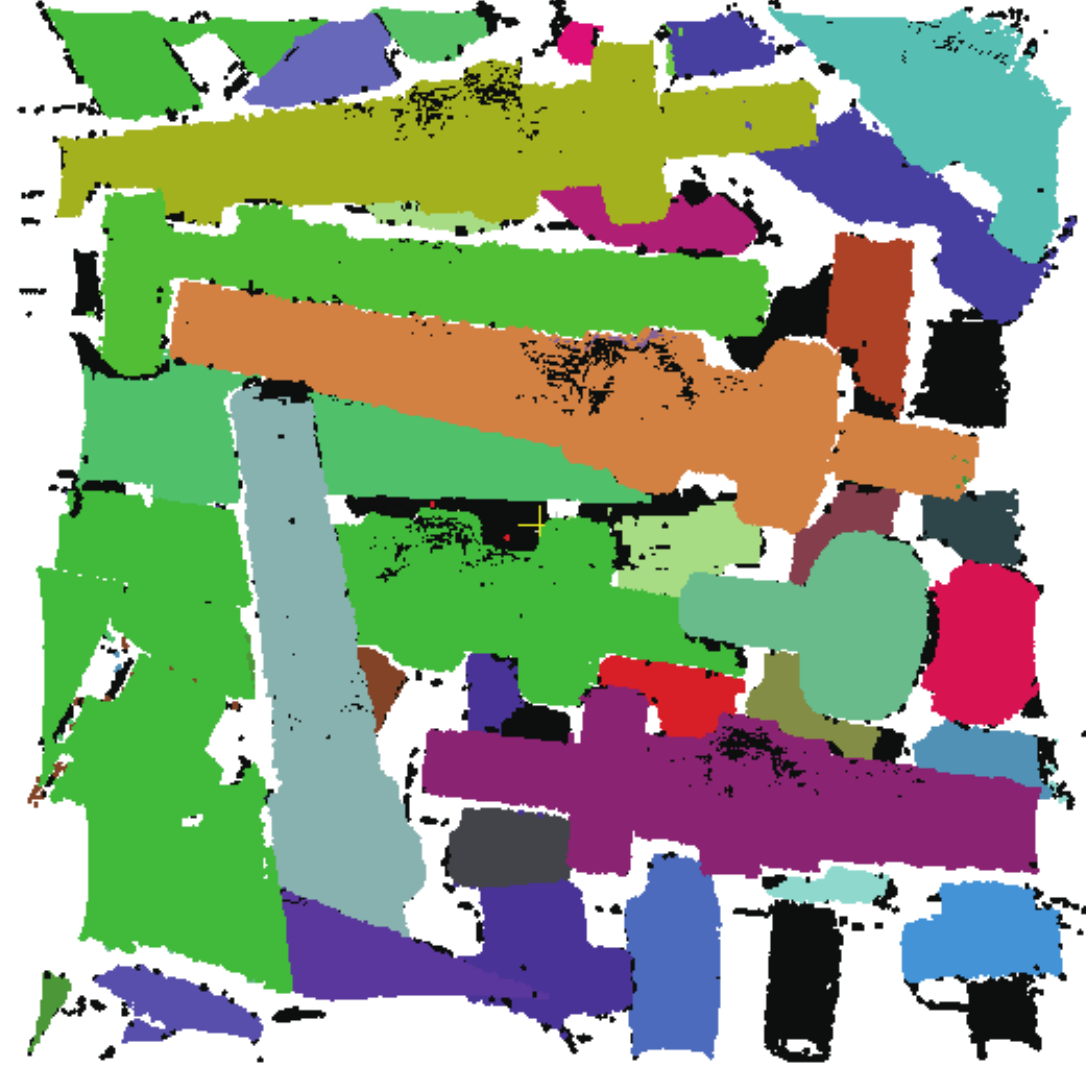}&
\includegraphics[width=0.24\columnwidth]{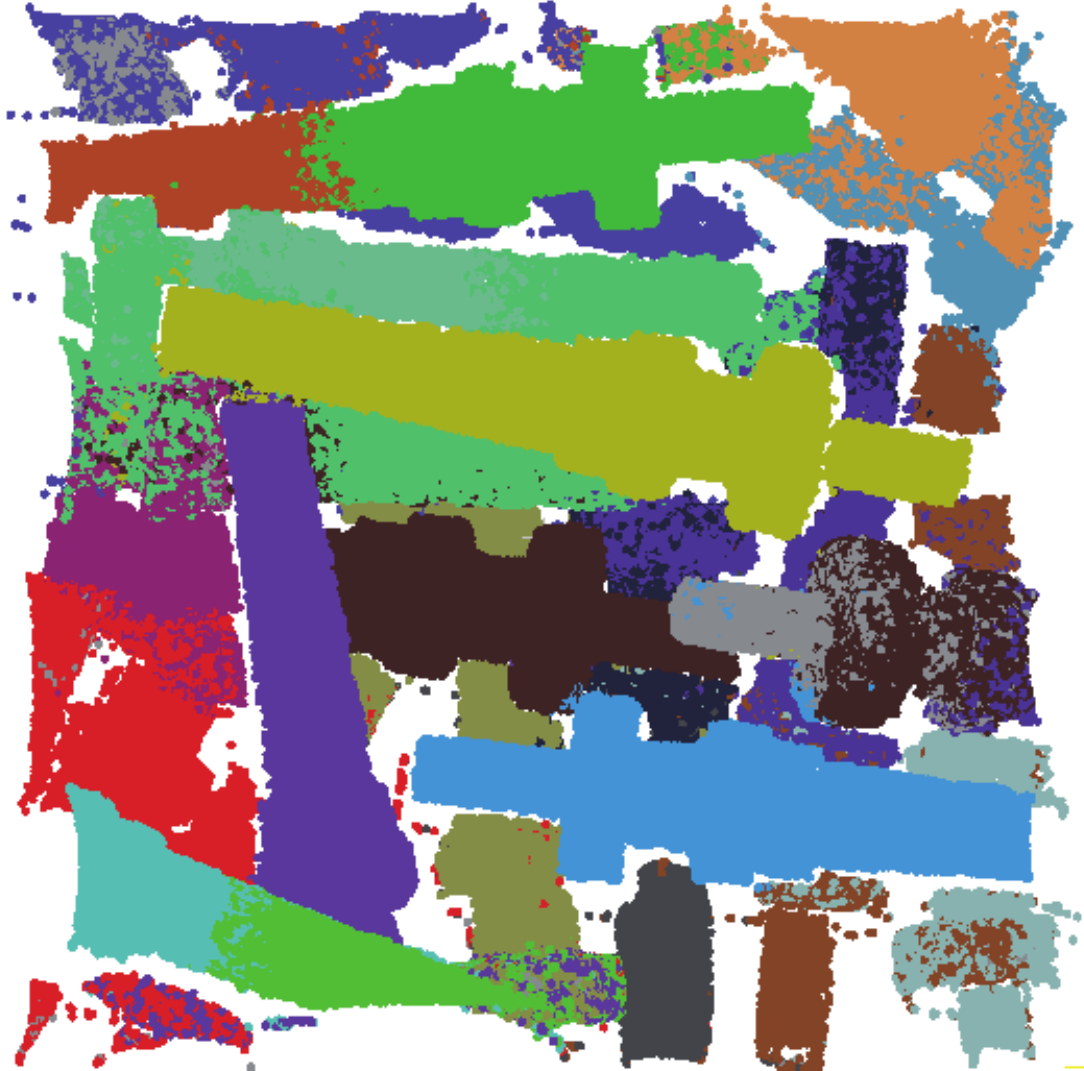}&
\includegraphics[width=0.24\columnwidth]{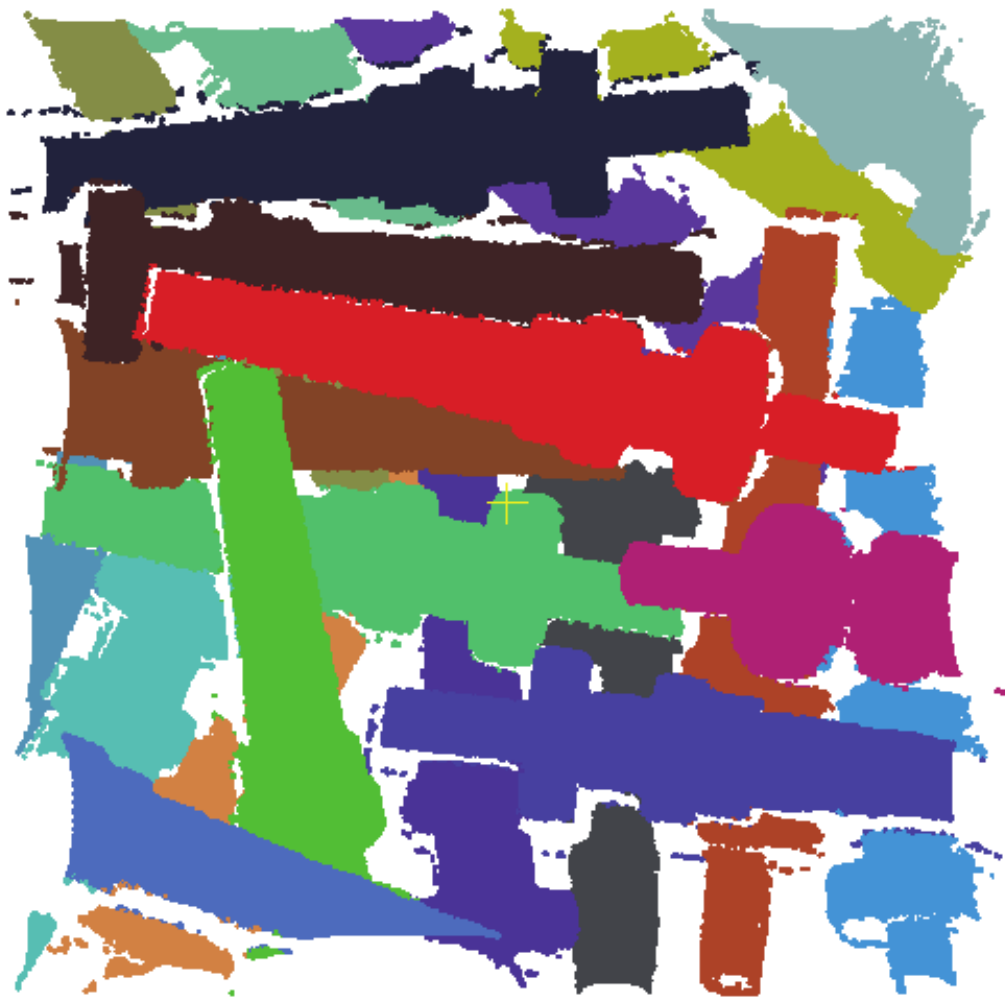}\\
\end{tabular}
}
\subfigure{
\begin{tabular}{ccccccc}
\includegraphics[width=0.24\columnwidth]{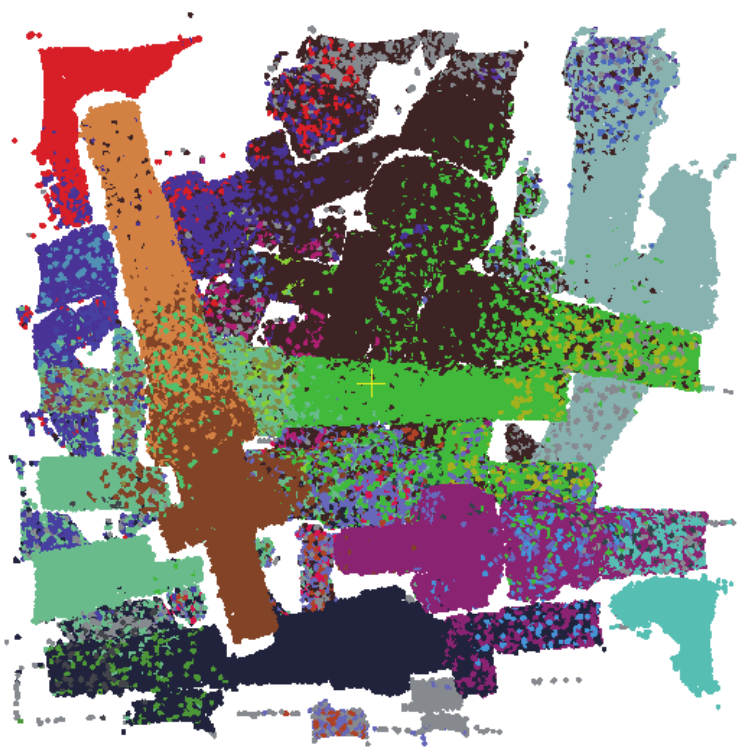}&
\includegraphics[width=0.24\columnwidth]{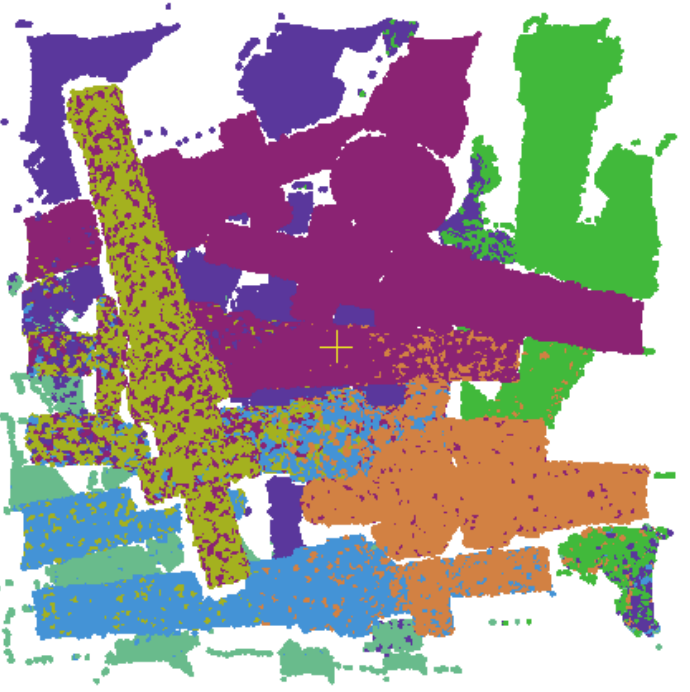}&
\includegraphics[width=0.24\columnwidth]{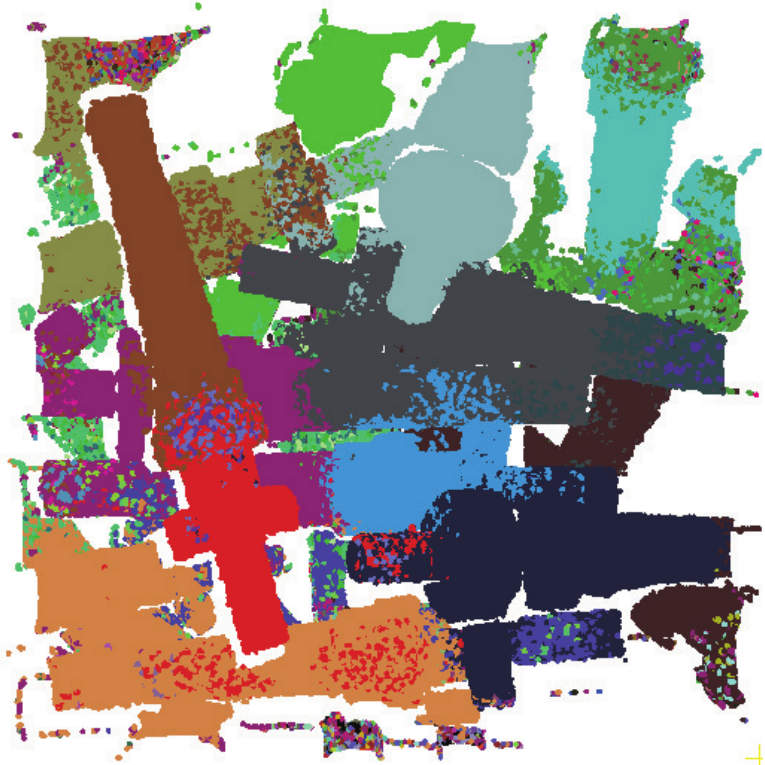}&
\includegraphics[width=0.24\columnwidth]{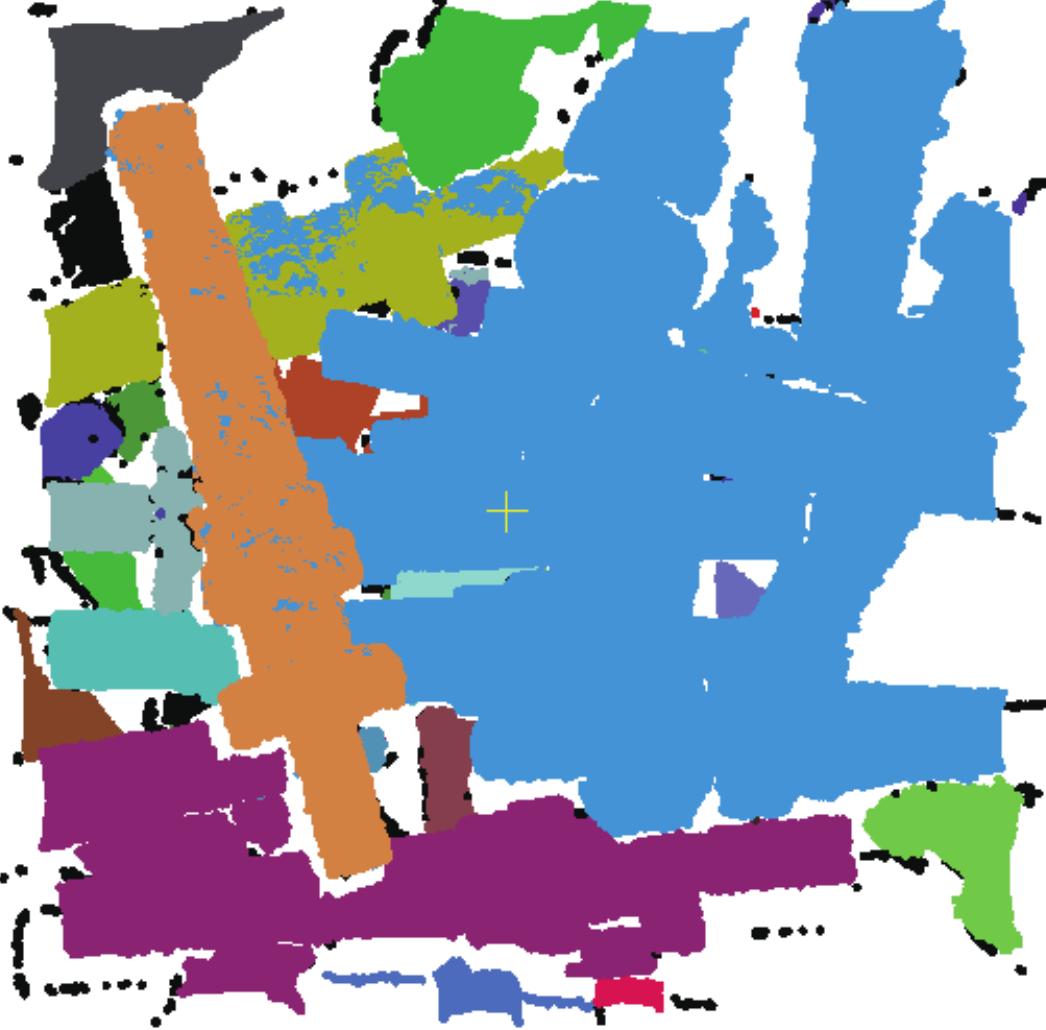}&
\includegraphics[width=0.24\columnwidth]{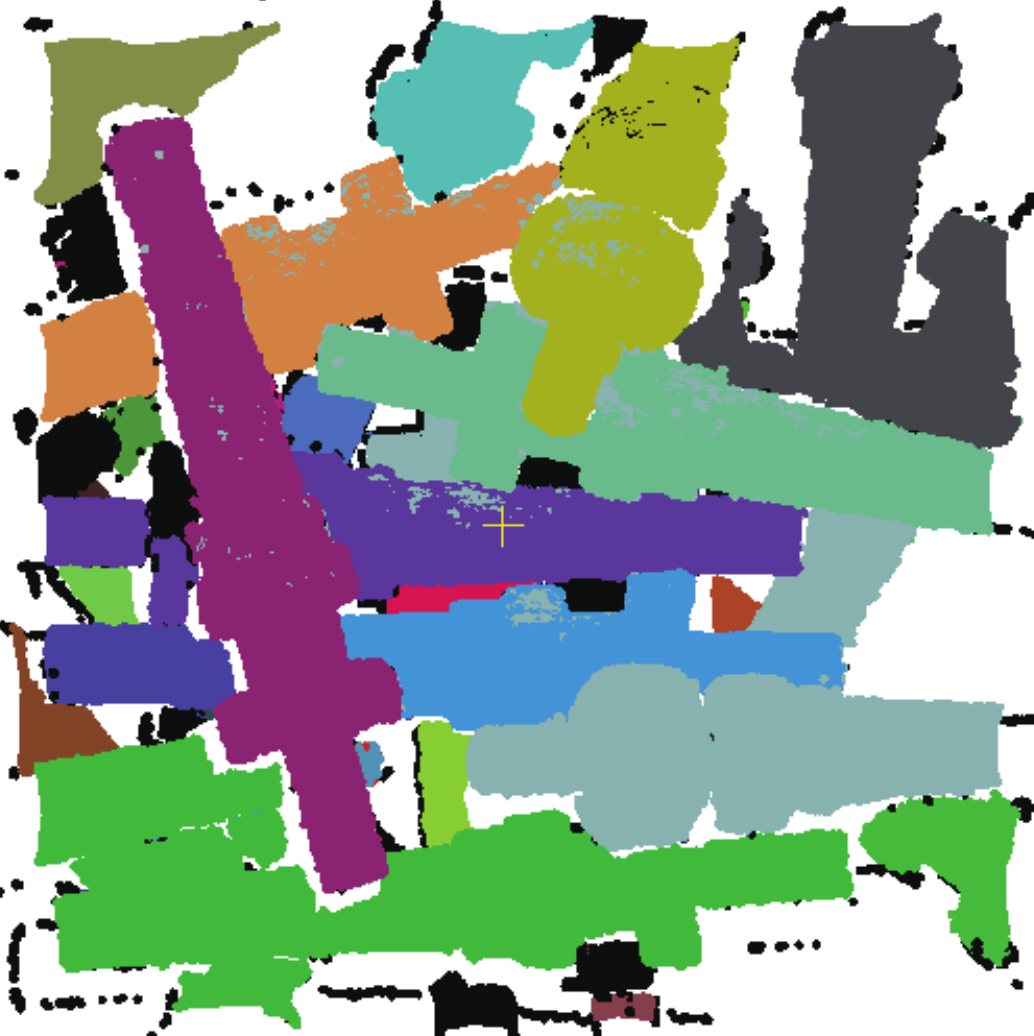}&
\includegraphics[width=0.24\columnwidth]{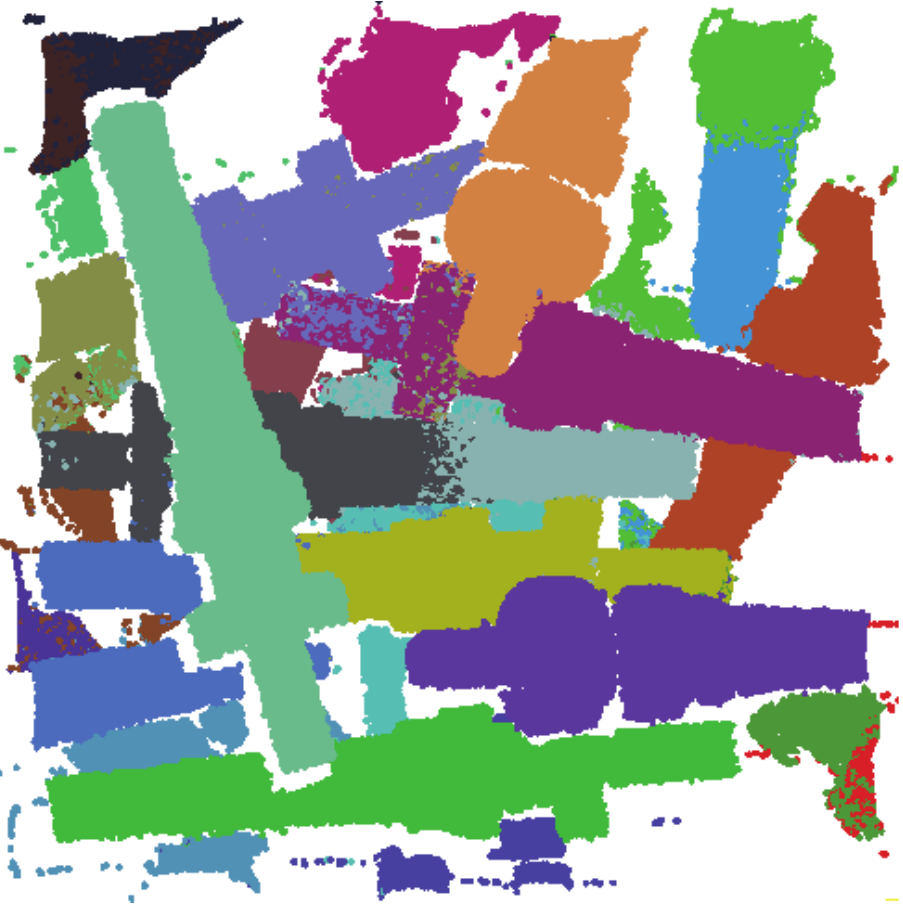}&
\includegraphics[width=0.24\columnwidth]{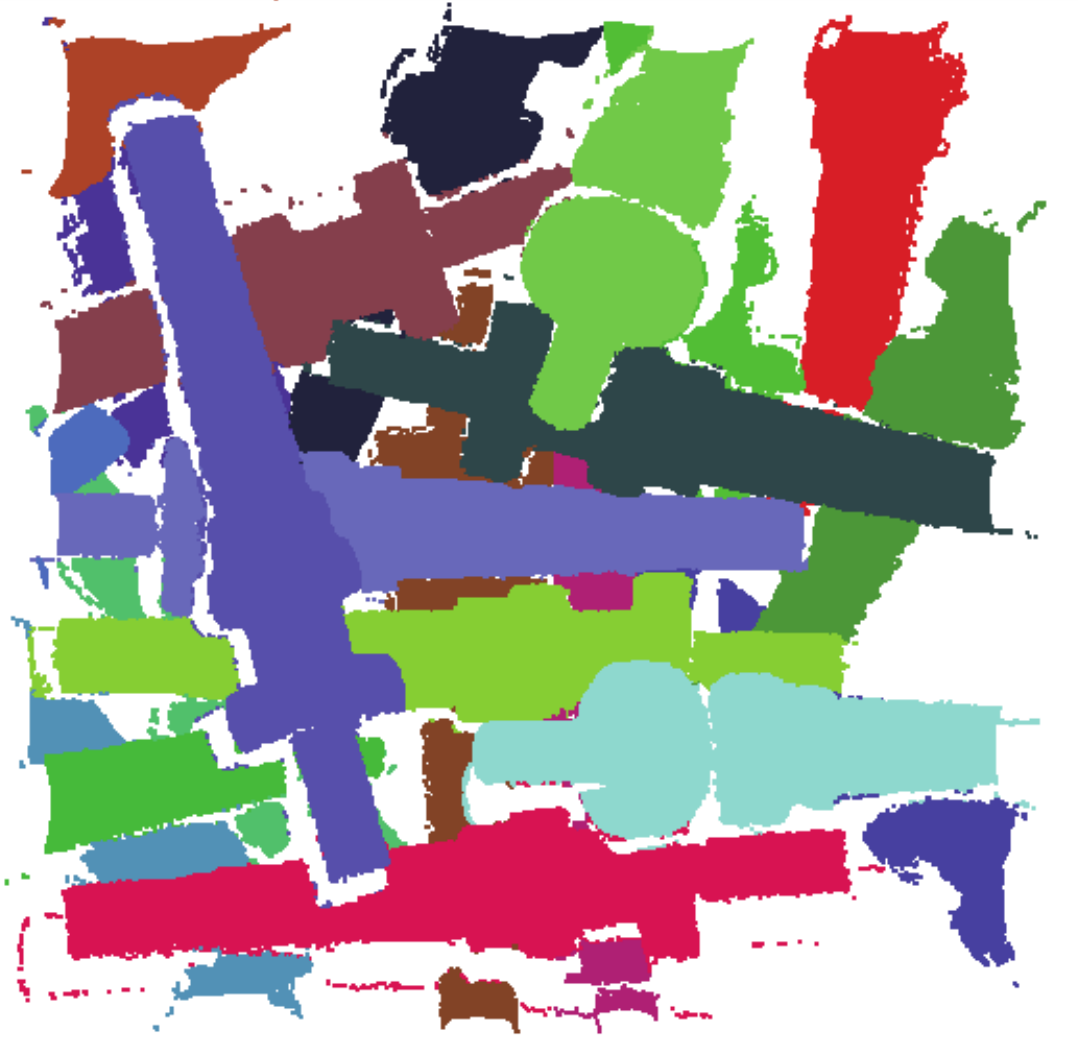}\\

SGPN* \cite{SGPN}&ASIS* \cite{asis}&3D-BoNet* \cite{3dbonet} &PointGroup* &PointGroup \cite{PointGroup}&FPCC&Ground Truth\\
\end{tabular}{}
}
\caption{Comparison results on IPA. The performance of SGPN and ASIS is poor on bin-picking scene. 3D-BoNet has difficulty in distinguishing some overlapping instances. The performance of FPCC is acceptable even in high clutter environment. *: Feature extractor is DGCNN.}
\label{Fig:qualitative}
\end{figure*}

\section{Experiment}
\label{experiment}
\subsection{Dataset}
We test five types of industrial objects, as shown in Fig. \ref{fig: models}. 
Ring screw and gear shaft are from the Fraunhofer IPA Bin-Picking dataset\cite{large-binpicking} and object A, B and C are from XA Bin-Picking dataset\cite{xu}. 
The details of datasets are as follows:
\begin{itemize}
\item Fraunhofer IPA Bin-Picking dataset (IPA)\cite{large-binpicking}: This is the first public dataset for 6D object pose estimation and instance segmentation for bin-picking that contains enough annotated data for learning-based methods. The dataset consists of both synthetic and real-world scenes.   
Depth images, 3D point cloud, 6D pose annotation of each object, visibility score, and a segmentation mask for each object are provided in both synthetic and real-world scenes. 
The dataset contains ten different objects. The training scenes of all objects are synthetic, and only the test scenes of gear shaft and ring screw are real-world data.

\item XA Bin-Picking dataset (XA)\cite{xu}: Y. Xu and S. Arai et al. have developed a dataset of boundary 3D point cloud containing three types of industrial objects as shown in Fig. \ref{fig: models} for instance segmentation on bin-picking scene. 
The training dataset is generated by simulation, while the test dataset is collected from the real-world. 
There are 500 training scenes and 20 test scenes for each object. We supplement the ground truth of object B and C in the test dataset.
Each test scene contains about 60,000 boundary points. 
The examples of synthetic scenes are presented in Fig. \ref{TrainSet}, respectively. 
\end{itemize}

\begin{figure*}
\vspace{0.5cm}
\begin{tabular}{m{0.2\columnwidth}<{\centering}m{1.8\columnwidth}}
\centering
Ring Screw&
\subfigure{
\begin{tabular}{cccc}
\includegraphics[width=.4\columnwidth]{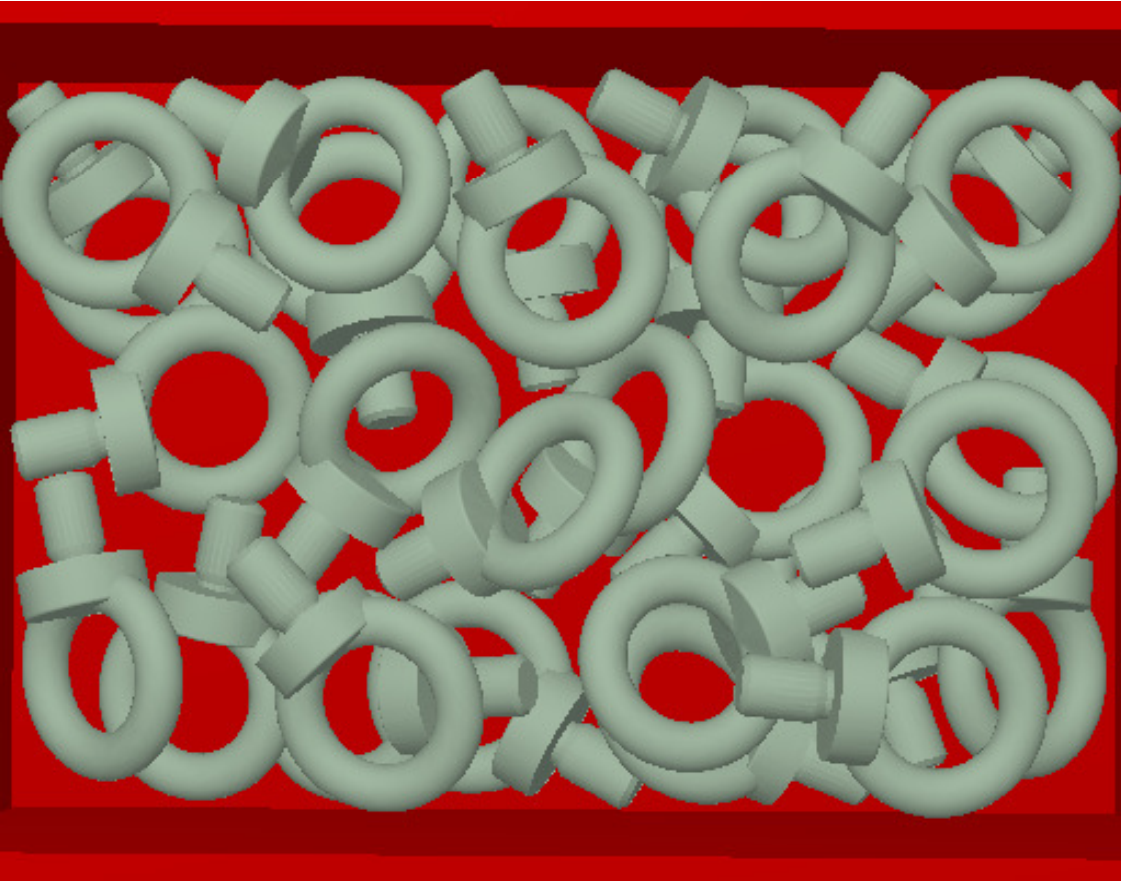}&
\includegraphics[width=.4\columnwidth]{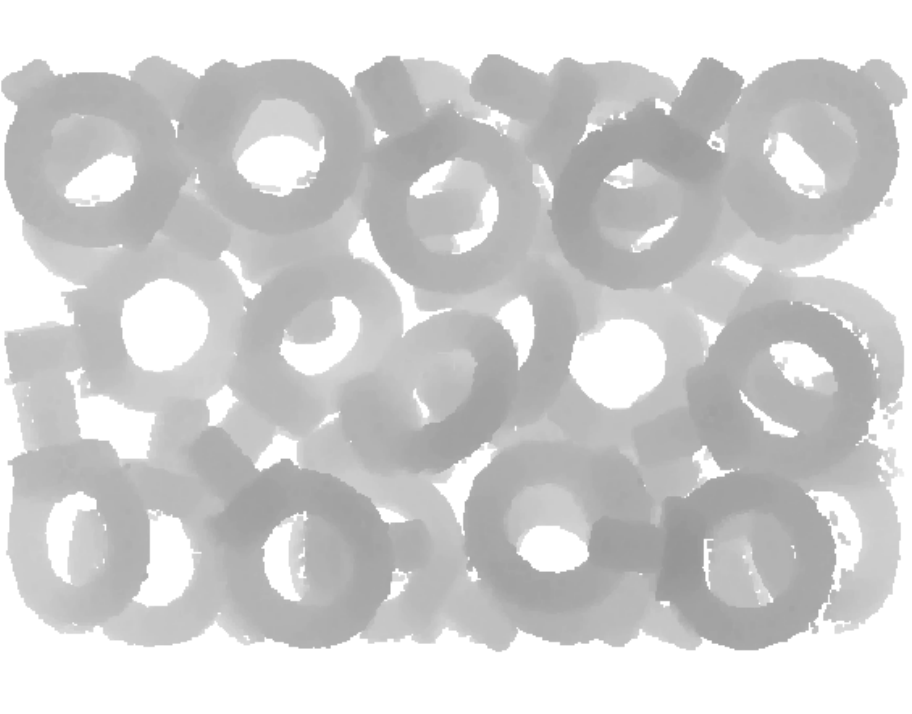}&
\includegraphics[width=.4\columnwidth]{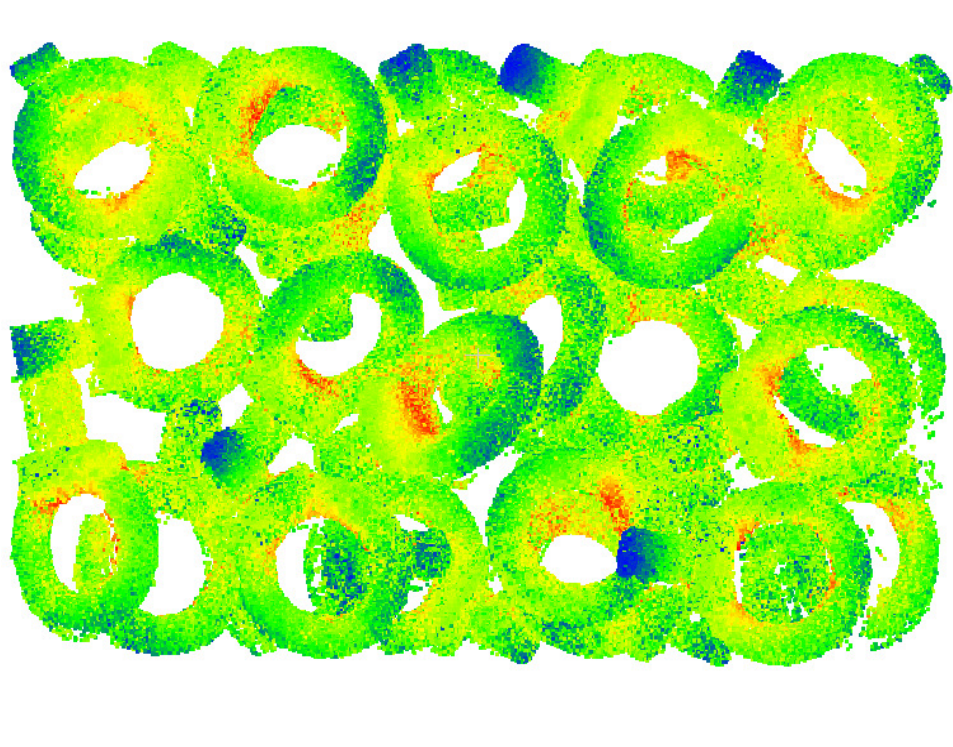}&
\includegraphics[width=.4\columnwidth]{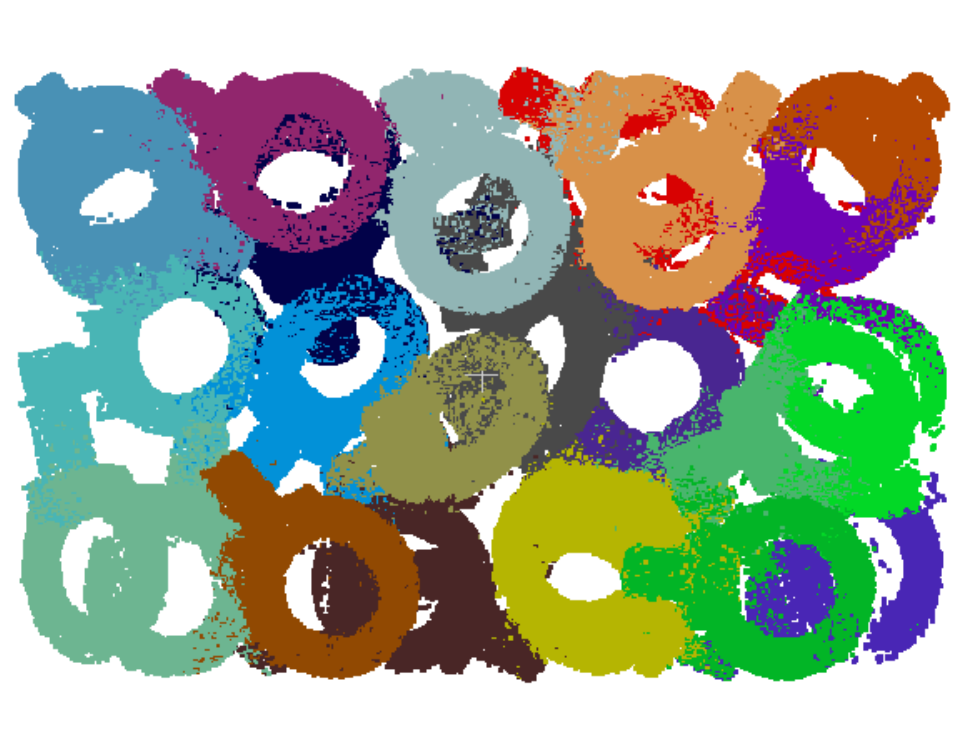}\\
\end{tabular}
}\\
\end{tabular}
\\
\begin{tabular}{m{0.2\columnwidth}<{\centering}m{1.8\columnwidth}}
Gear Shaft&
\subfigure{
\begin{tabular}{cccc}
\includegraphics[width=.4\columnwidth]{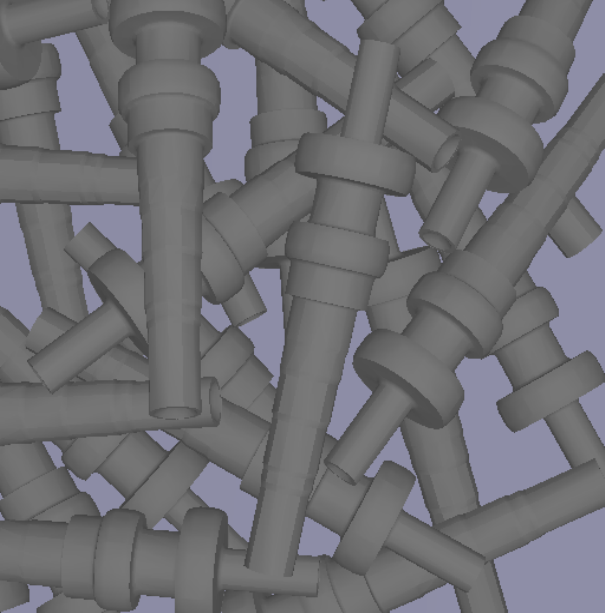}&
\includegraphics[width=.4\columnwidth]{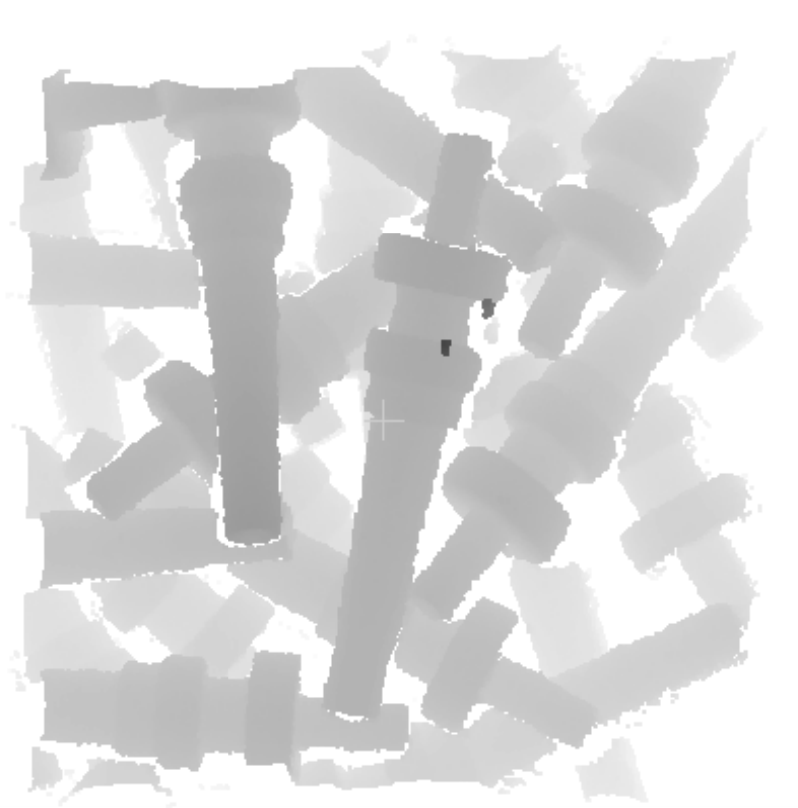}&
\includegraphics[width=.4\columnwidth]{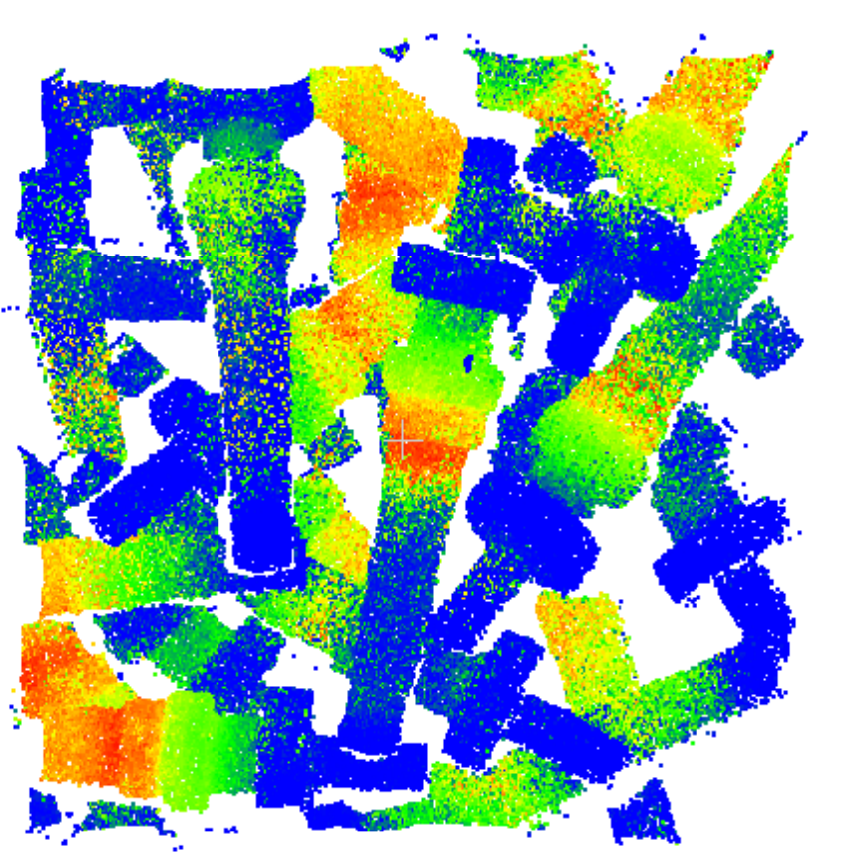}&
\includegraphics[width=.4\columnwidth]{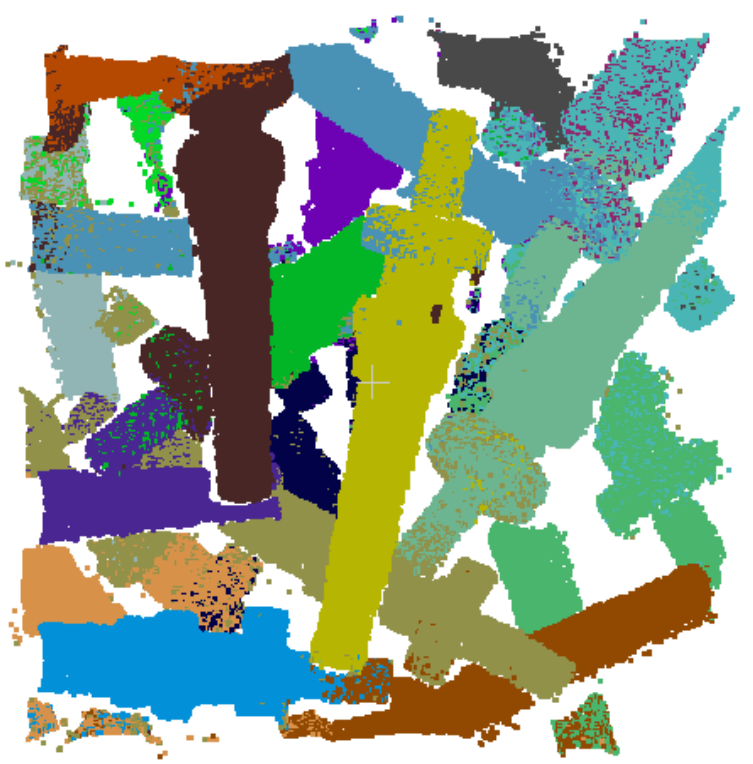}\\
\end{tabular}
}
\end{tabular}
\\
\begin{tabular}{m{0.2\columnwidth}<{\centering}m{1.8\columnwidth}}
Object A&
\subfigure{
\begin{tabular}{cccc}
\includegraphics[width=.4\columnwidth]{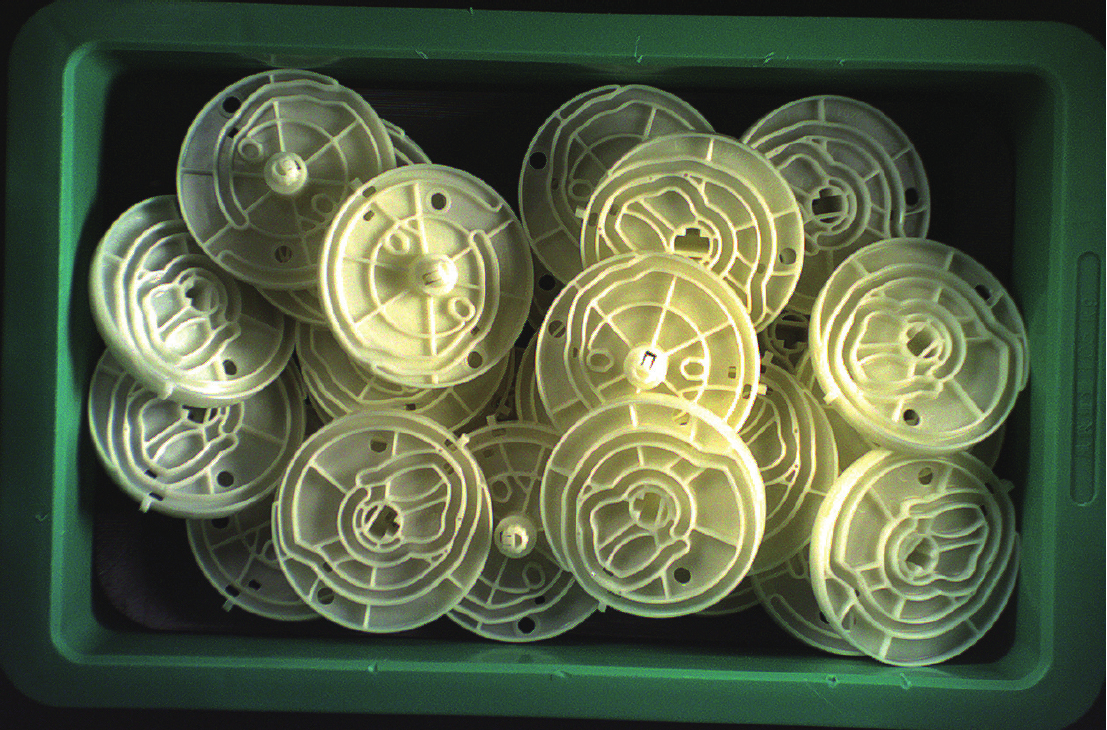}&
\includegraphics[width=.4\columnwidth]{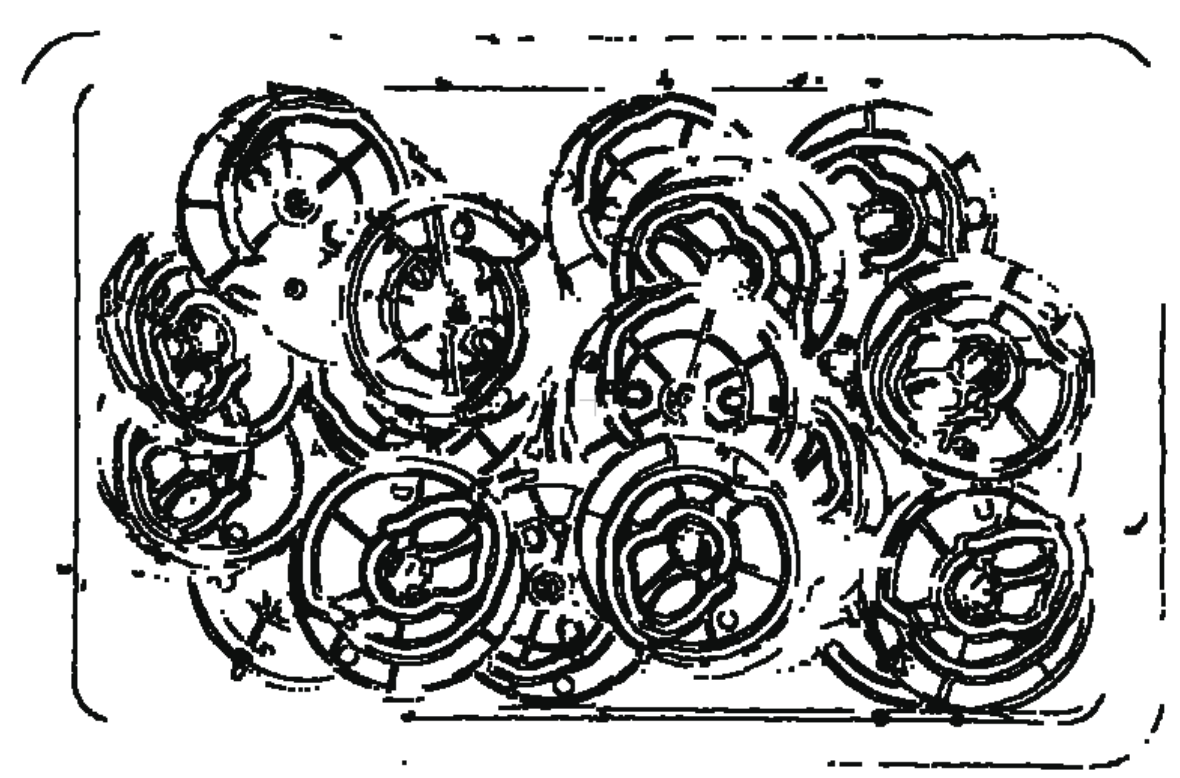}&
\includegraphics[width=.4\columnwidth]{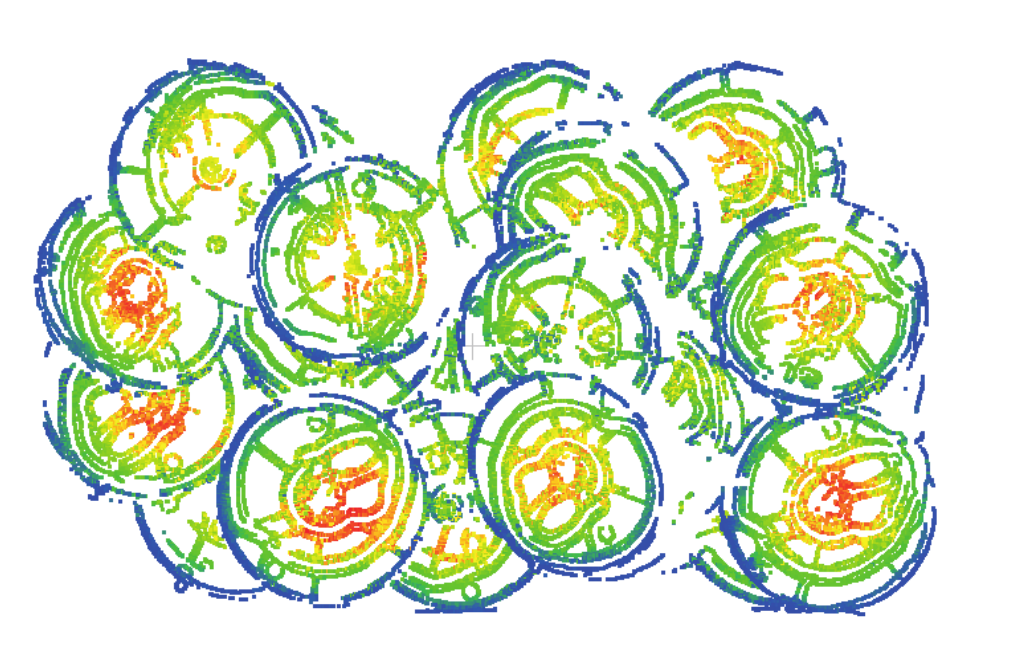}&
\includegraphics[width=.4\columnwidth]{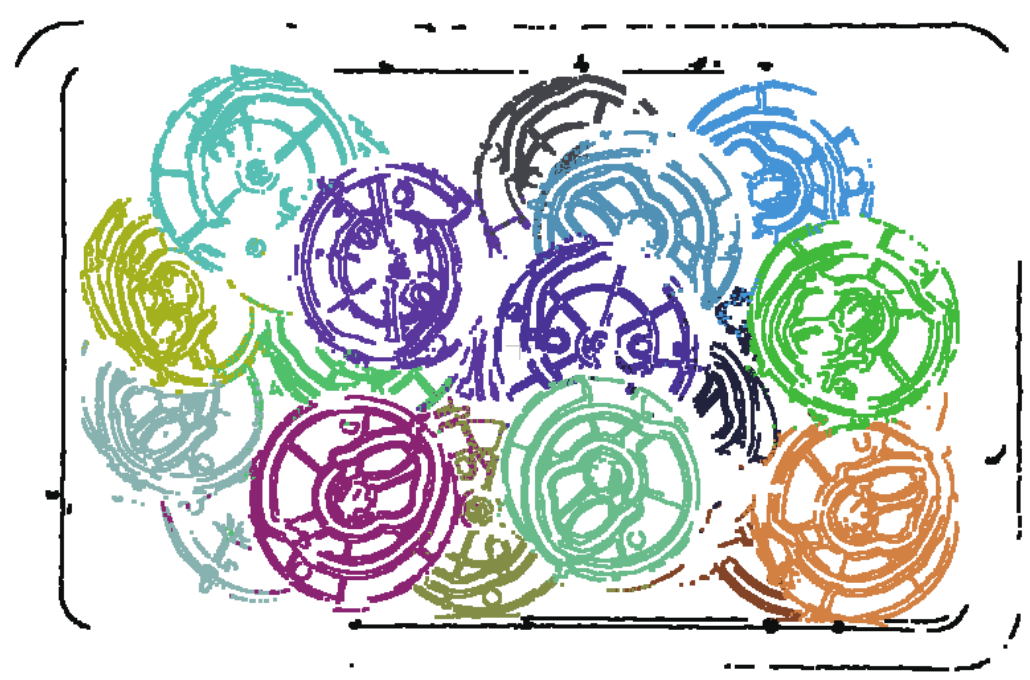}\\
\end{tabular}
}
\end{tabular}
\\
\begin{tabular}{m{0.2\columnwidth}<{\centering}m{1.8\columnwidth}}
Object B&
\subfigure{
\begin{tabular}{cccc}
\includegraphics[width=.4\columnwidth]{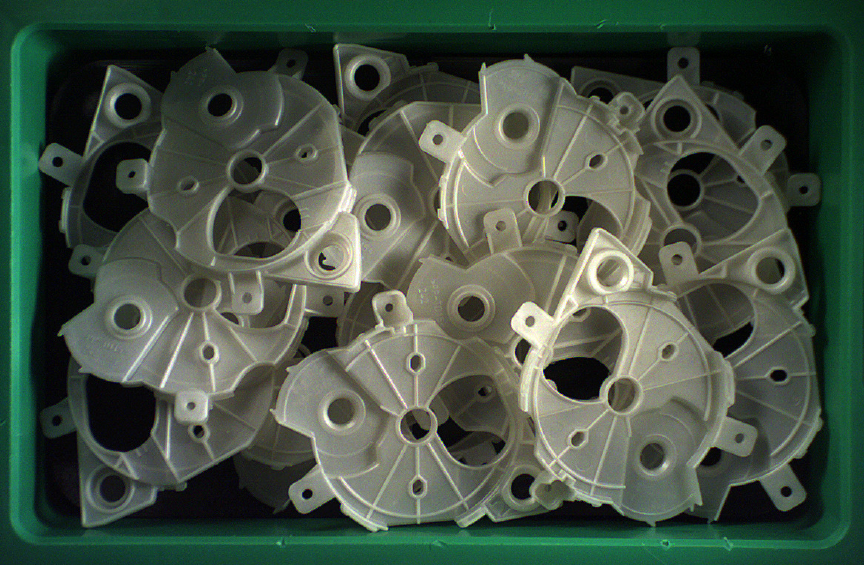}&
\includegraphics[width=.4\columnwidth]{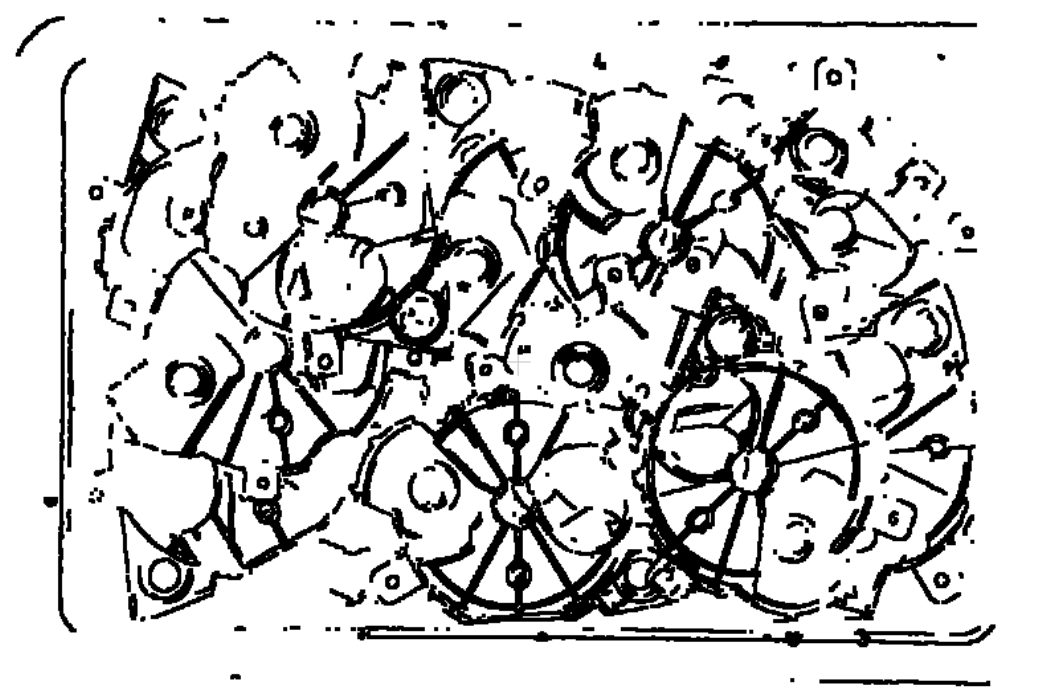}&
\includegraphics[width=.4\columnwidth]{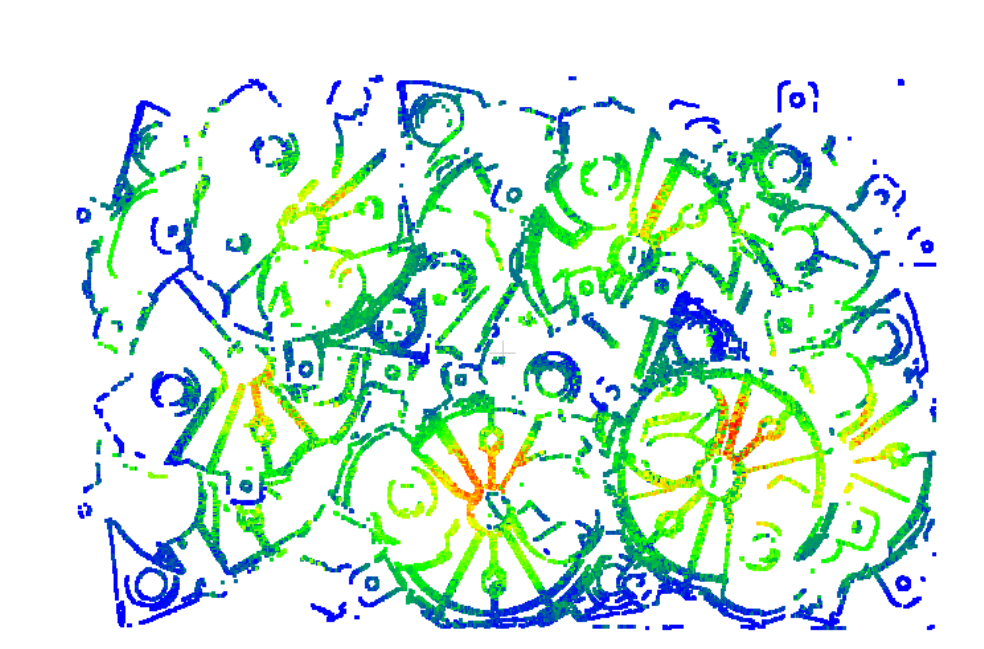}&
\includegraphics[width=.4\columnwidth]{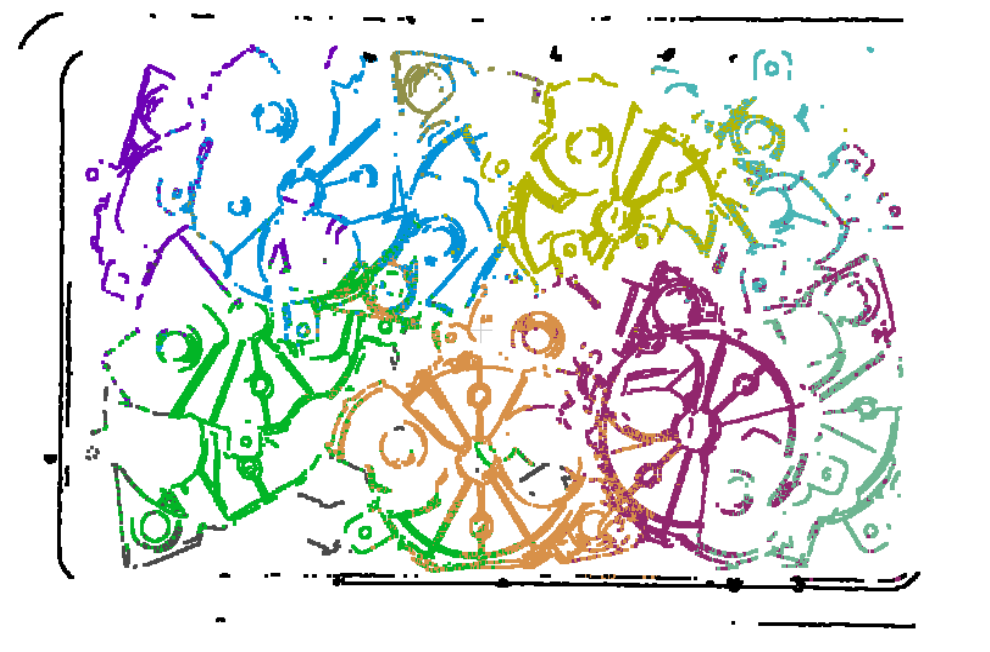}\\
\end{tabular}
}
\end{tabular}
\\
\begin{tabular}{m{0.2\columnwidth}<{\centering}m{1.8\columnwidth}}
Object C&
\subfigure{
\begin{tabular}{cccc}
\includegraphics[width=.4\columnwidth]{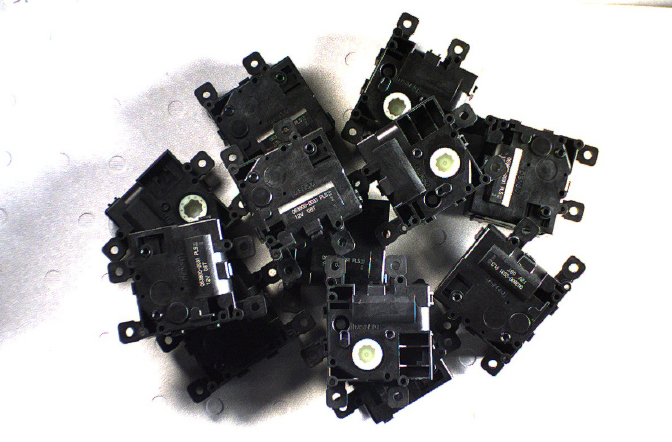}&
\includegraphics[width=.4\columnwidth]{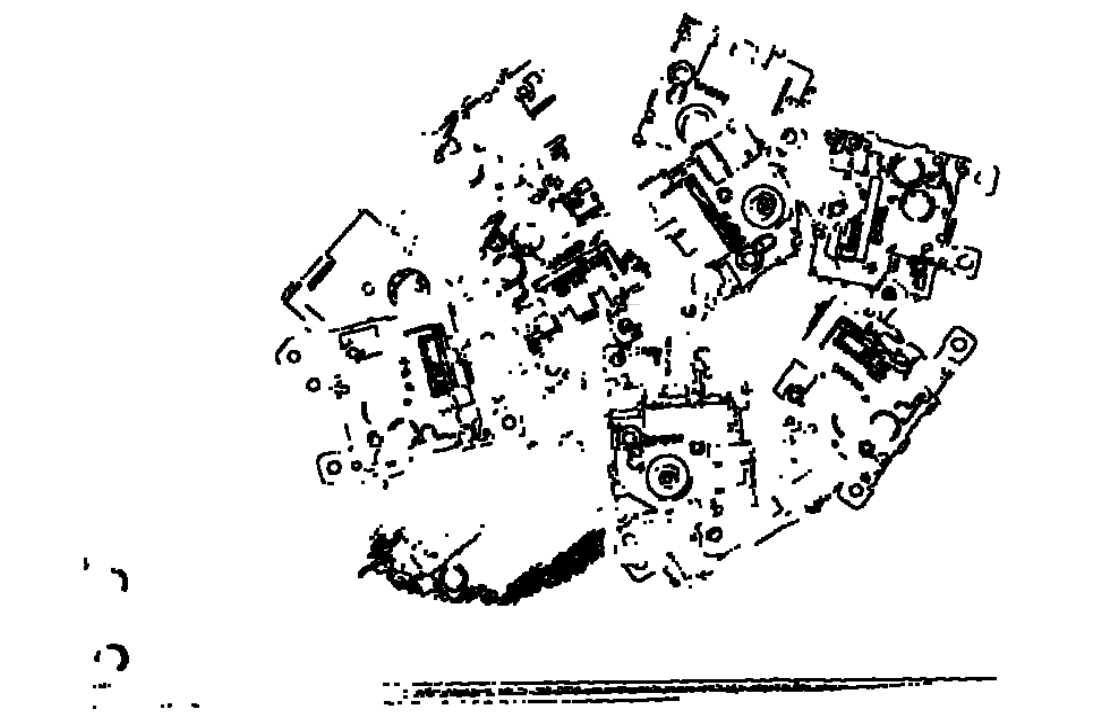}&
\includegraphics[width=.4\columnwidth]{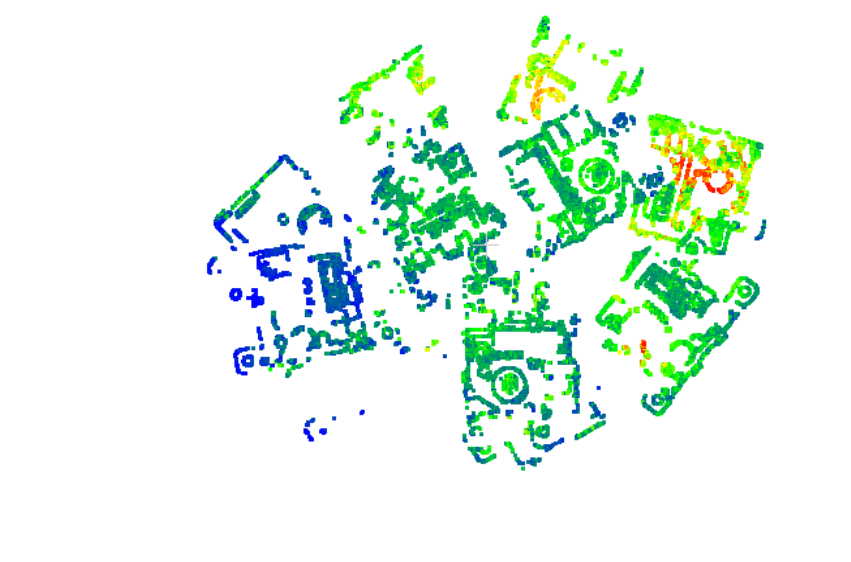}&
\includegraphics[width=.4\columnwidth]{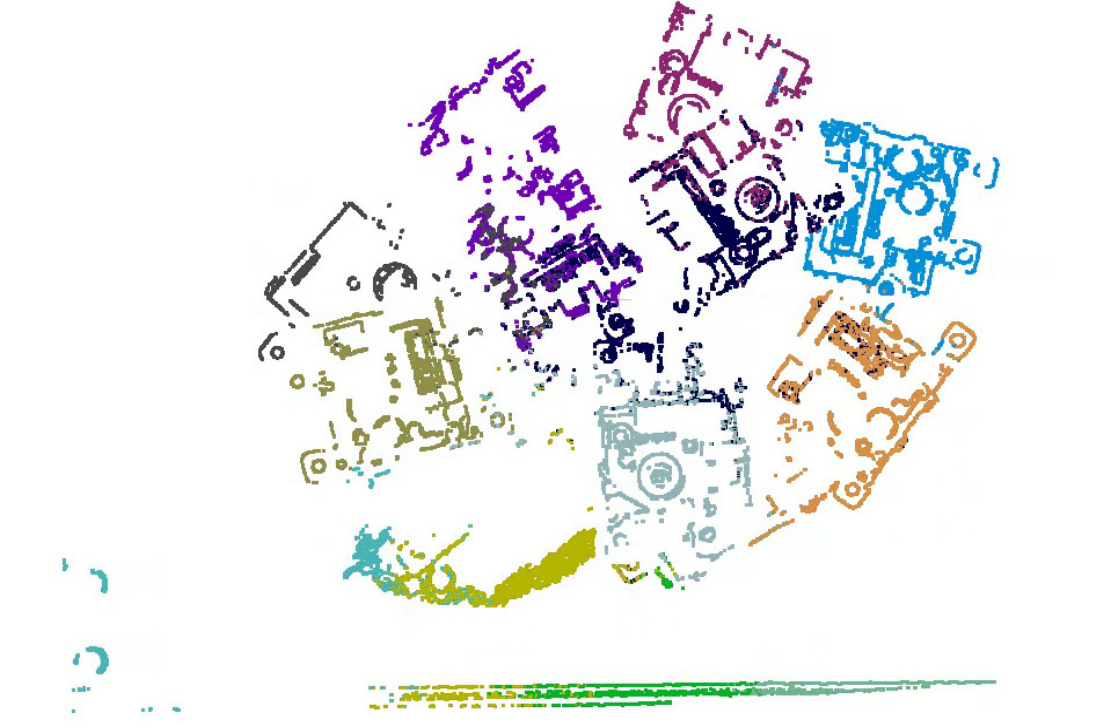}\\
Real Scene&3D Point Cloud& Center Score Prediction & Instance Prediction\\
\end{tabular}
}
\end{tabular}

\caption{Visualization of the results of instance segmentation given by FPCC on IPA\cite{large-binpicking} (Ring Screw, Gear Shaft) and XA\cite{xu} (Object A, B, C). Many objects of the same class are stacked together in the Real Scene. 3D Point Cloud is represented by $(\overline{x},\overline{y},\overline{z}, n_x, n_y, n_z$) and then input into FPCC-Net. 
Center Score Prediction is the predicted center score and the color bar is the same as one in Fig \ref{Fig:center_score}. 
The results for instance segmentation is shown in the last column. 
Different colors represent different instances. 
\label{Fig:ins-results}}
\end{figure*}
\begin{table*}[t]
\caption{Results of instance segmentation on industrial objects. The metric is precision(\%) and recall(\%) with an IoU threshold of 0.5.}
\label{table:ins_part}
\centering
\resizebox{0.98\textwidth}{!}{%
\begin{tabular}{c|c|rr|rr|rr|rr|rr}
\hline
\multirow{2}*{\diagbox{\textbf{Method}}{\textbf{Data}}}&\multirow{2}*{Backbone}&\multicolumn{2}{c|}{\textbf{Ring Screw}}&\multicolumn{2}{c|}{\textbf{Gear Shaft}}&\multicolumn{2}{c|}{\textbf{Object A}}&\multicolumn{2}{c|}{\textbf{Object B}}&\multicolumn{2}{c}{\textbf{Object C}}\\
\cline{3-11}
~&&\textbf{Precision}&\textbf{Recall}&\textbf{Precision}&\textbf{Recall}&\textbf{Precision}&\textbf{Recall}&\textbf{Precision}&\textbf{Recall}&\textbf{Precision}&\textbf{Recall}\\
\hline
SGPN&DGCNN&10.92&14.67&15.25&21.65&41.92&28.98&20.07&25.92&24.62&25.39\\
\hline
ASIS&DGCNN&15.54&11.56&20.14&9.46&64.67&28.72&55.43&23.61&67.08&42.06\\
\hline
3D-BoNet&DGCNN&27.88&19.80&26.57&20.11&66.05&50.23&42.22&26.38&45.88&62.40\\
\hline
PointGroup&DGCNN&3.64&1.91 &3.65&1.73&  6.19&1.54&  7.68&1.89&  5.37&1.02\\
\hline
PointGroup&Sparse Convolution&52.40&41.22&\textbf{58.79}&36.80&\textbf{91.88}&46.22&75.22&39.35&\textbf{79.83}&41.12\\
\hline
FPCC&DGCNN&\textbf{58.43}&\textbf{48.74}&54.29&\textbf{69.53}&89.71&\textbf{67.28}&\textbf{80.88}&\textbf{50.92}&78.64&\textbf{64.28}\\
\hline  
\end{tabular}
}
\end{table*}

\subsection{Experimental setting}
FPCC-Net is implemented in the TensorFlow framework and trained using the Adam\cite{adam} optimizer with initial learning rate of 0.0001, batch size 2 and momentum 0.9. 
All training and validation are conducted on Nvidia GTX1080 GPU and Intel Core i7 8700K CPU with 32 GB RAM. 
During the training phase, $\epsilon_1 = 0.5$, $\epsilon_2 = 1$, $\alpha = 3$ and $\beta = 2$ are set.
$\gamma=1$ and $d_{\max}=0.07,0.08,0.6,1.6,1.2$ correspond to the size of ring screw, gear shaft, object A, object B and object C, respectively.
In each batch in the training process, input points ($N=4,096$) are randomly sampled from each scene and each point can be sampled only once. 
Each point is converted to a 6D vector ($\overline{x}, \overline{y}, \overline{z}, n_x, n_y, n_z$) for inputting FPCC-Net. 
The sampling is repeated until the remained points of the scene is less than $N$. 
The network is trained for 30 epochs. 
It takes around ten hours to train FPCC-Net for each object.
\subsection{Instance segmentation evaluation}
\subsubsection{Precision and Recall}
Fig. \ref{Fig:ins-results} shows results of instances segmentation and center scores predicted by FPCC-Net on the five types of industrial objects. The points near the center have higher score than boundary points. 
FPCC can distinguish the majority of instances clearly even with heavy occlusion. 
For ring-shaped objects, i.e., ring screw, the geometric center point is not on the part, thus the reference point used in the clustering is shifted from the center of object. The shifted reference point reduces the performance of segmentation. We consider that our clustering method is not suitable for objects that are mostly empty in the center, e.g., ring.

Fig. \ref{Fig:qualitative} shows the comparison results of instances segmentation with SGPN, ASIS, 3D-BoNet, PointGroup and FPCC. Since the objects in a bin are identical, we eliminate the branches of semantic segmentation in these compared networks and treat the semantic information of all input points as the same.
It should be noted that the original SGPN, ASIS and 3D-BoNet use PointNet or PointNet++ as their feature extractors. When we re-evaluate their method in the point cloud without color, we found that the training process could not converge at all. We believe that it is because PointNet and PointNet++ cannot effectively learn local geometric feature. Therefore, we replaced their feature extractor with DGCNN. For PointGroup, we provide two results, one is the original PointGroup, and the other is the PointGroup using DGCNN as backbone.
Different predicted instances are shown in different colors. SGPN and ASIS adopt a similar clustering method, which accumulate the error of each potential group in the process of merging and results in predicting multiple instances as one instance in the heavy occlusion. In contrast, FPCC is robust to occlusion because no redundant merging is required. The performance of 3D-BoNet is also poor. This is because the 3D-BoNet trained by synthetic data is difficult to reasonably predict the 3D binding box of each instance on real data. The performance of PointGroup using DGCNN as the backbone is very poor, which shows that the features extracted by DGCNN can not support the clustering algorithm of PointGroup.


Table \ref{table:ins_part} reports the classical metrics precision and recall with an IoU threshold of 0.5 on ring screw, gear shaft, object A, object B and object C. In the case of using DGCNN as the backbone, FPCC achieves the best results and is competitive with the results of the original PointGroup.
Original PointGroup uses sparse convolution as its backbone, extracting more discriminative features of the points to support its clustering algorithm. 
The recall of PointGroup is lower because PointGroup ignores some uncertain segmentation results based on its confidence scores.



\begin{table}[!t]
\centering
\caption{Computation speed comparisons [s/scene]}
\label{Tab:speed}
\resizebox{0.48\textwidth}{!}{%
\begin{tabular}{c|c|rr|rrr}
\hline
\diagbox{\textbf{Method}}{\textbf{Data}}&\textbf{Backbone}&\textbf{Ring Screw}&\textbf{Gear Shaft}&\textbf{Object A} &\textbf{Object B}& \textbf{Object C}\\

\hline
SGPN&DGCNN&298.60&104.90&53.32&72.83&28.21\\
\hline
ASIS&DGCNN&8.37&7.45&3.62&3.09&1.78\\
\hline
3D-BoNet&DGCNN&3.91&3.80&1.29&0.91&0.80\\
\hline
PointGroup&DGCNN&2.16&1.62&1.26&0.78&1.20\\
\hline
PointGroup&Spase Convolution&0.81&0.64&0.39&0.47&0.36\\
\hline
FPCC&DGCNN&1.78&1.43&0.79&0.55&0.65\\
\hline
\end{tabular}
}
\end{table}
\subsubsection{Computation Time}
\label{Time}
\begin{figure}[!t]
\centering
\subfigure{
\includegraphics[width=1.\columnwidth]{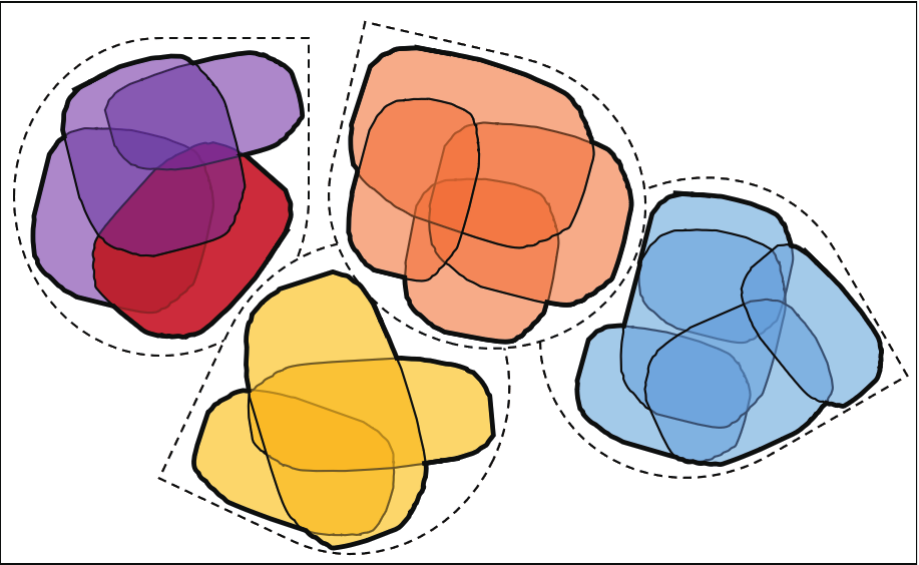}}
\caption{Illustration of merging process in SGPN. 
Dotted lines indicate instance and thin solid lines represent potential groups. 
Groups with an IoU greater than a threshold are merged and consist a new group   represented by the thick solid line. 
Red group will be merged with purple group, since the IoU between red and purple groups is greater than the threshold.}
\label{fig: merge_sgpn}
\end{figure}

\begin{figure}[!t]
\centering
\subfigure[Computation time on ring]{
\includegraphics[width=0.48\columnwidth]{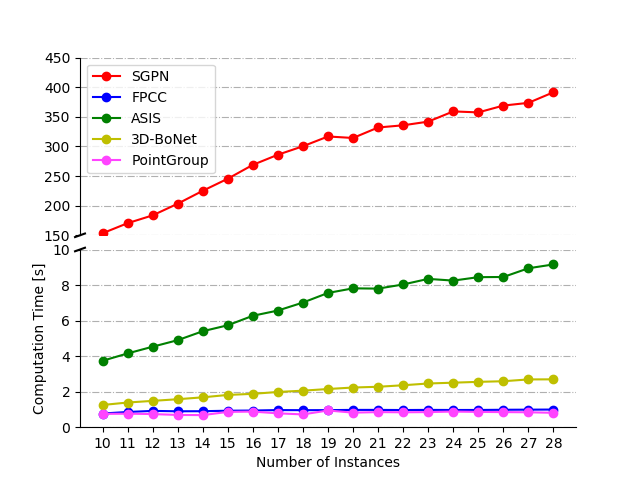}}
\subfigure[Computation time on gear]{
\includegraphics[width=0.48\columnwidth]{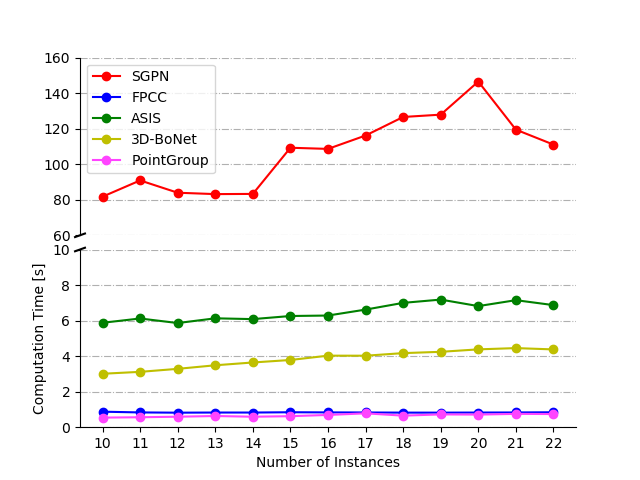}}
\caption{Average computation time for different number of instances. 
As the number of instances increases, the computation time of SGPN and ASIS increase significantly.}
\label{Fig:time_constant}
\end{figure}

Table \ref{Tab:speed} reports the average computation time per scene measured on Intel Core i7 8700K CPU and Nvidia GTX1080 GPU. 
Each bin-picking scene of XA (Object A, B, and C) contains about 20 objects with 60,000 points. 
It takes around $0.6 \sim 0.8$ [s] to process one scene of XA by FPCC. 
A scene of IPA contains about $10 \sim 30$ objects with 15,000 points, which could also be processed in about 1.5 [s] by FPCC. 
PointGroup has faster processing speed with the support of sparse convolution.
In summary, FPCC is about 60 times faster than SGPN and 2 times faster than ASIS and 3D-BoNet, but slower than PointGroup with sparse convolution. 
Most of the existing methods employing a clustering strategies similar to SGPN find some reference points to generate potential groups, and then merge each group according to some metrices, e.g., IoU. 
The computation complexities for clustering processes of SGPN and FPCC are analyzed as follows: Firstly, we assume that there are $m$ instances and $n$ points in the scene ($m \ll n$), and the outputs of clustering algorithms of SGPN and FPCC are correct. 
The clustering of SGPN is divided into two steps: potential groups generating and groups merging. 
Firstly, SGPN takes $N_{\mathrm{SGPN}} \gg m$ points with high confidence as reference points of the clustering. 
Then $N_{\mathrm{SGPN}}$ groups are generated based on the feature distance between the reference points and the other points. 
The computational complexity of groups generating, that is, the number of computation of the feature distances tends towards $\mathcal{O}(nN_{\mathrm{SGPN}})$. 
Next, two groups with an IoU greater than a threshold, such as 0.5, are merged together, as shown in Fig. \ref{fig: merge_sgpn}. 
The computation complexity of groups merging for SGPN, that is, the number of computation of IoUs between two potential groups tends towards $\mathcal{O}(mN_{\mathrm{SGPN}})$. 
In contrast, FPCC does not generate any potential group. 
The reference points of FPCC are found by Algorithm \ref{NMS}. 
Each point in the scene point cloud is directly clustered with the nearest reference point, that is, the center point based on feature distance. Hence the competition complexity of clustering for FPCC tends towards $\mathcal{O}(nm)$. 
In summary, the computational complexities of groups generating of FPCC is much lower than SGPN, that is $\mathcal{O}(nm) \leq \mathcal{O}(nN_{\mathrm{SGPN}})$, since the number of potential groups $N_{\mathrm{SGPN}}$ is much higher than the number of instances $m$. 
In addition, the calculation of IoU between groups which is not required for FPCC needs more computational resources than that of feature distances. 
Thus we can conclude that the computational cost of FPCC is much smaller than that of SGPN theoretically. 
An example of the relationship of computation time and the number of instances is shown in Fig. \ref{Fig:time_constant}. 
ASIS adopted a clustering method similar to SGPN, and their calculation time increases significantly with the increase of the number of instances. 3D-BoNet also need more calculation time because of increase of bounding boxes. However, the computation time of FPCC and PointGroup are not sensitive to the number of instances.

\subsection{Ablation studies}
\subsubsection{Ablation on the $d_{\rm{max}}$}
We use different screening factor $\gamma$ in non-maximum suppression. The performance is shown in Table \ref{table:Ablation_d}. The larger $\gamma$ tends to cluster different objects together, while the smaller $\gamma$ tends to produce over-segmentation.
\begin{table}[t]
\caption{Ablation results for non-maximum suppression with different $\gamma$ on IPA dataset}
\label{table:Ablation_d}
\centering
\resizebox{0.48\textwidth}{!}{%
\begin{tabular}{c|rr|rr}
\hline
\multirow{2}*{\diagbox{\textbf{$\gamma$}}{\textbf{Data}}}&\multicolumn{2}{c|}{\textbf{Ring Screw}}&\multicolumn{2}{c}{\textbf{Gear Shaft}}\\
\cline{2-5}
~&\textbf{Precision}&\textbf{Recall}&\textbf{Precision}&\textbf{Recall}\\
\hline
0.5&27.52&45.21&63.56&38.12\\
\hline
0.8&65.70&54.64&66.35&63.71\\
\hline
1.0&58.43&48.74&54.29&69.53\\
\hline
1.2&10.86&25.66& 28.65&12.19\\
\hline
\end{tabular}
}
\end{table}
\subsubsection{Ablation on VDM and ASM}
Four ablation experiments are conducted on the bin-picking scenes to evaluate the effectiveness of VDM and ASM in FPCC-Net. 
Precision and recall with an IoU threshold of 0.5 is added to interpret the results. 
Four groups for this ablation experiments are explained below:
\begin{enumerate}
  \item VDM and ASM are removed, that is to say, no weights are added to compute the embedded feature loss $L_{\rm{EF}}$.
  \item Only VDM is used in loss $L_{\rm{EF}}$.
  \item Only ASM is used in loss $L_{\rm{EF}}$.
  \item Both VDM and ASM are adopted in loss $L_{\rm{EF}}$.
\end{enumerate}
Table \ref{table:ablation} shows that the performance of the first group is the worst among the four experiments and two weight matrices improve the ability of the network to extract distinctive features of the 3D point cloud. The result indicates that the two weight matrices we designed are reasonable. VDM makes FPCC-Net do not need to care about the feature distance of points with too large Euclidean distance. ASM reduce the network's focus on point pairs whose two points are boundary points.

\begin{table}[t]
\caption{Results of ablation experiments. The metric is precision and recall with an IoU threshold of 0.5.}
\label{table:ablation}
\centering
\begin{tabular}{c|c|c|rr|rr}
\hline
\multirow{2}*{}&\multirow{2}*{VDM}&\multirow{2}*{ASM}&\multicolumn{2}{c}{\textbf{Ring Screw}}&\multicolumn{2}{c}{\textbf{Gear Shaft}}\\
\cline{4-7}
&&&\textbf{Precision}&\textbf{Recall}&\textbf{Precision}&\textbf{Recall}\\
\hline
1&&&52.08&40.54&40.90&23.74\\
\hline
2&\checkmark&&54.55&45.86&41.92&28.12\\
\hline
3&&\checkmark&54.46&45.69&44.85&48.43\\
\hline
4&\checkmark&\checkmark&\textbf{58.43}&\textbf{48.74}&\textbf{54.29}&\textbf{69.53}\\
\hline  
\end{tabular}
\end{table}

\section{Conclusions}
\label{conclusions}
This paper proposes a fast and effective 3D point cloud instance segmentation named FPCC for the bin-picking scene, which has multi instances but a single class. 
FPCC includes FPCC-Net which predicts embedded features and the geometric center score of each point, and a fast clustering algorithm using the outputs of FPCC-Net. 
Two hand-designed weight matrices are introduced for improving the performance of FPCC-Net. 
A novel clustering algorithm is proposed for instance segmentation. 
For multi instances but single class scenes, FPCC achieves better performance than existing methods even without manually labeled data. 
Besides, we theoretically prove that the computational complexity of FPCC is much lower than SGPN. 
  
This study also has a certain limitation that must be addressed in future work, as follows:  
1) FPCC is not particularly suitable for objects whose geometric center is not on the itself. We are considering using other methods to define the reference point of each instance.
2) We use the existing semantic network, i.e., DGCNN, as FPCC's feature extractor. Perhaps the feature extractor does not fit the task of instance segmentation perfectly. The improvement of feature extractor is our future work.

\printcredits

\bibliographystyle{elsarticle-num}
\bibliography{cite}

\bio{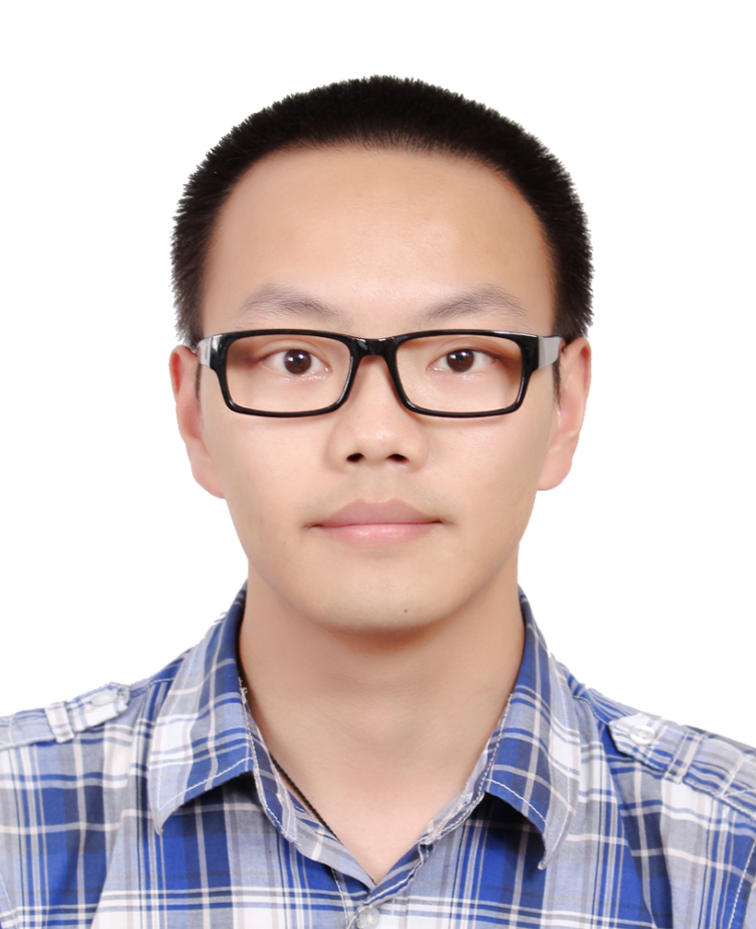}
Yajun Xu received the B.S. in mechanical engineering from Central South University, Changsha, China, in 2015. and the M.E. from graduate school of Engineering, Tohoku University, Japan.

He is currently a student with the Graduate School of Information Science and Technology, Hokkaido University, Japan. His research interests include deep learning, 2D/3D segmentation, and 6D pose estimation.
\endbio

\bio{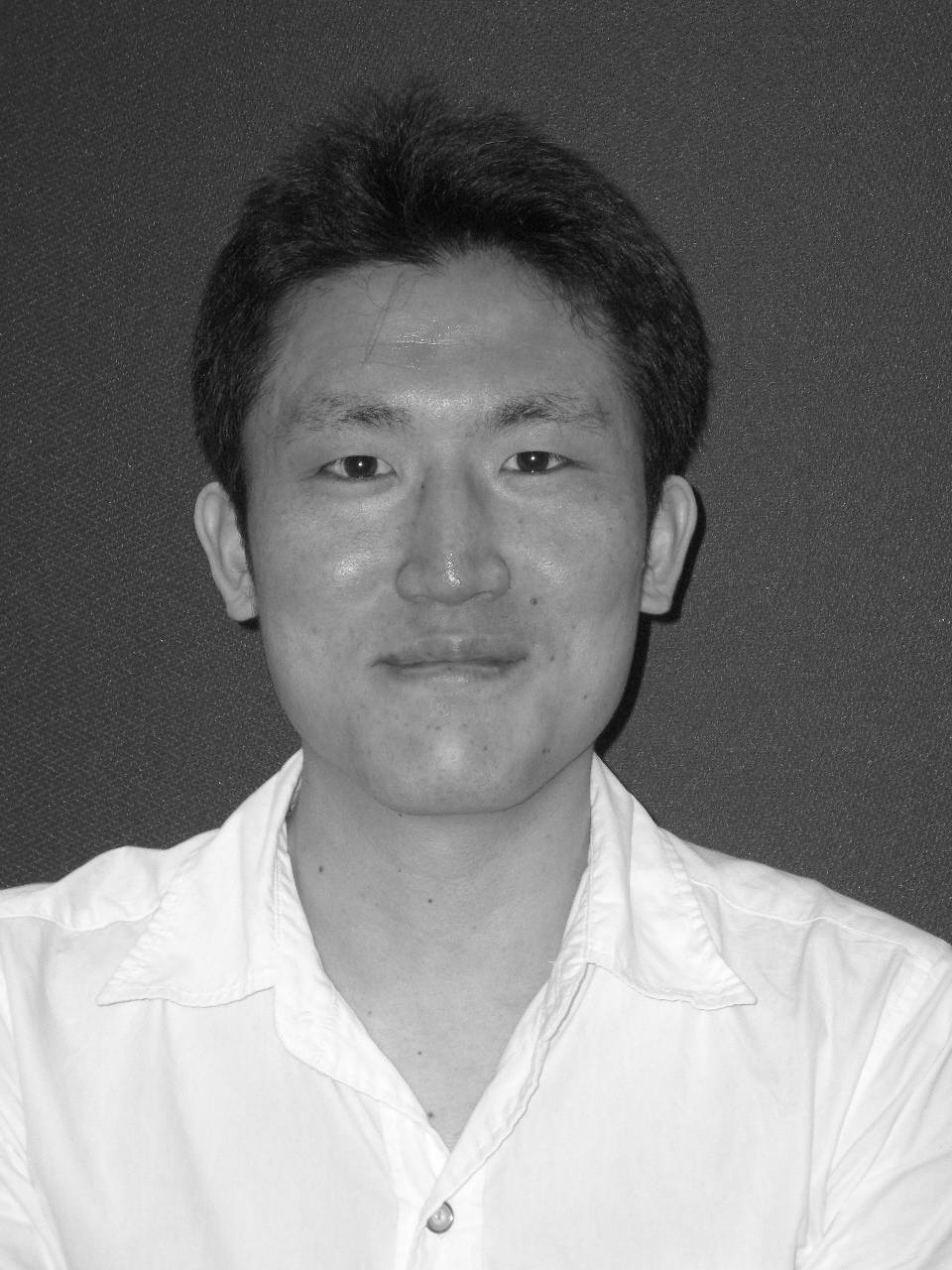}
Shogo Arai received a B.S. degree in aerospace engineering from Tohoku University, Sendai, Japan, in 2005 and M.S. and Ph.D. degrees in information sciences from Tohoku University in 2007 and 2010, respectively.
From 2010 to 2016, he was an assistant professor in the Intelligent Control Systems Laboratory at Tohoku University.
In 2016, he joined the System Robotics Laboratory in the Department of Robotics at Tohoku University as an associate professor, where he is currently an associate professor. He received Best Paper Award from FA Foundation in 2019,
The 32th Best Paper Award from The Robotics Society of Japan in 2019, 
Certificate of Merit for Best Presentation from The Japan Society of Mechanical Engineers in 2019, 
Excellent Paper Award from The Institute of Systems from Control and Information Engineers in 2010, 
Best Paper Award Finalist at IEEE International Conference on Mechatronics and Automation in 2012, 
SI2019 Excellent Presentation Award from The Society of Instrument and Control Engineers in 2019,
SI2018 Excellent Presentation Award from The Society of Instrument and Control Engineers in 2018, 
SI2017 Excellent Presentation Award from The Society of Instrument and Control Engineers in 2017, 
and Graduate School Research Award from Society of Automotive Engineers of Japan, Inc. in 2007.

His research focuses on the fields of robot vision, robotic bin-picking, 3D measurement, production robotics, networked control systems, and multi-agent systems.
\endbio

\bio{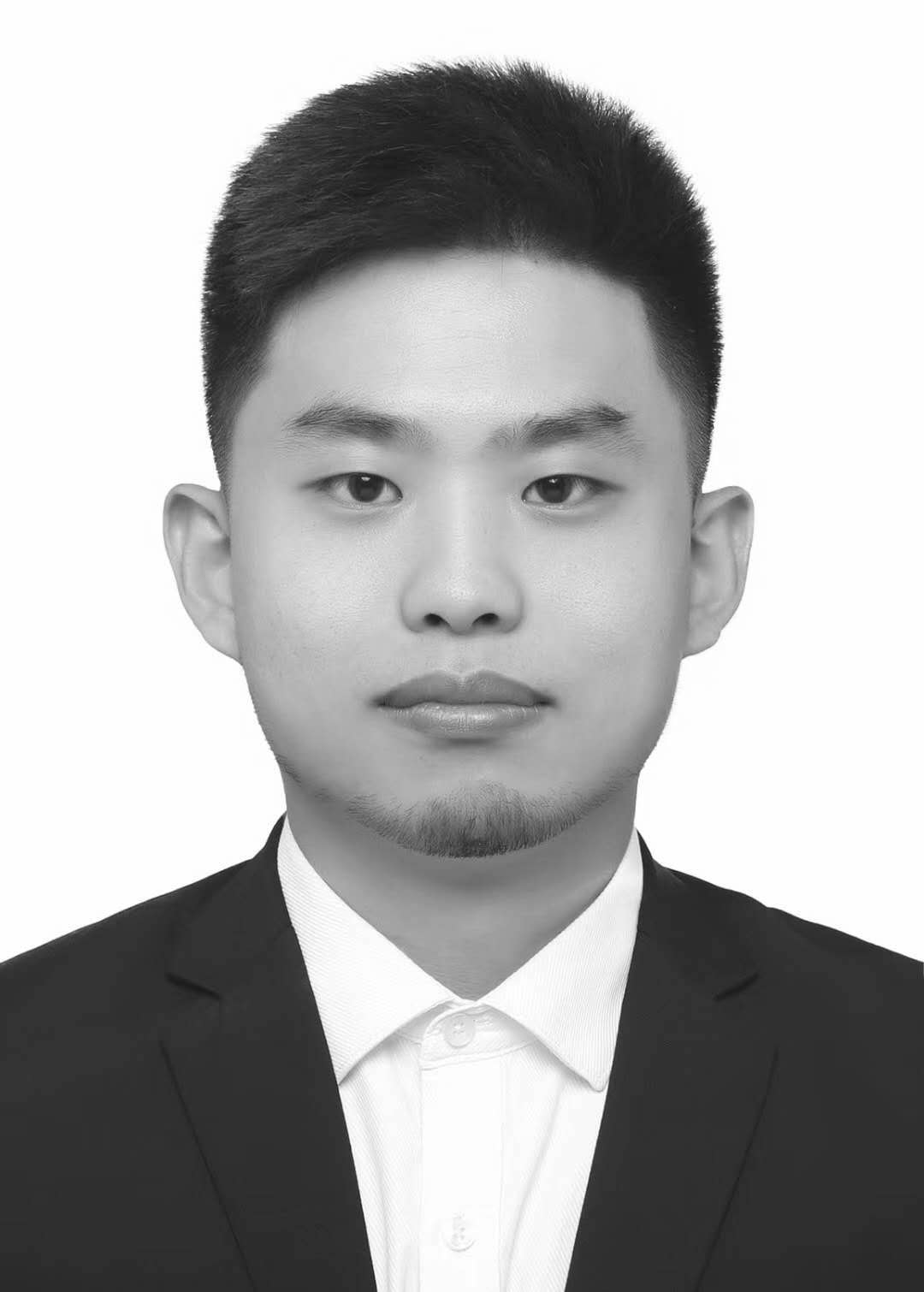}
Diyi Liu is a senior algorithm engineer at Mech-Mind Robotics Technologies, China. He received an MS degree and a Ph.D. in engineering from Tohoku University, Sendai, Japan, in 2016 and 2019, respectively. 

Dr. Liu received the Best Paper Award in Automation at the IEEE International Conference on Mechatronics and Automation in 2016 and a SI2017 Excellent Presentation Award from the Japanese Society of Instrument and Control Engineers in 2017. His research interests include the pose estimation and object detection of industrial parts.
\endbio

\bio{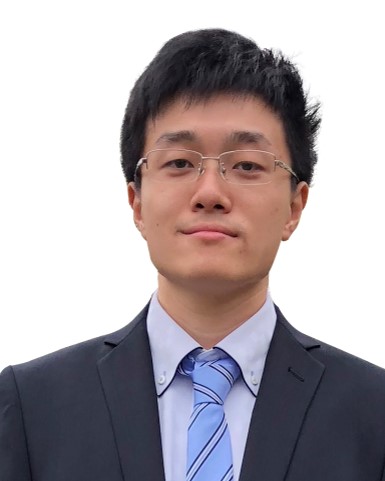}
Fangzhou Lin received an M.S. degree in information sciences from Tohoku University, Japan, Sendai, in 2021.

He is currently pursuing a Ph.D. degree from Tohoku University, Sendai, Japan.  His research interests include deep learning, satellite image processing, visual question answering, recurrent neural network architecture, and point cloud complication.
\endbio

\bio{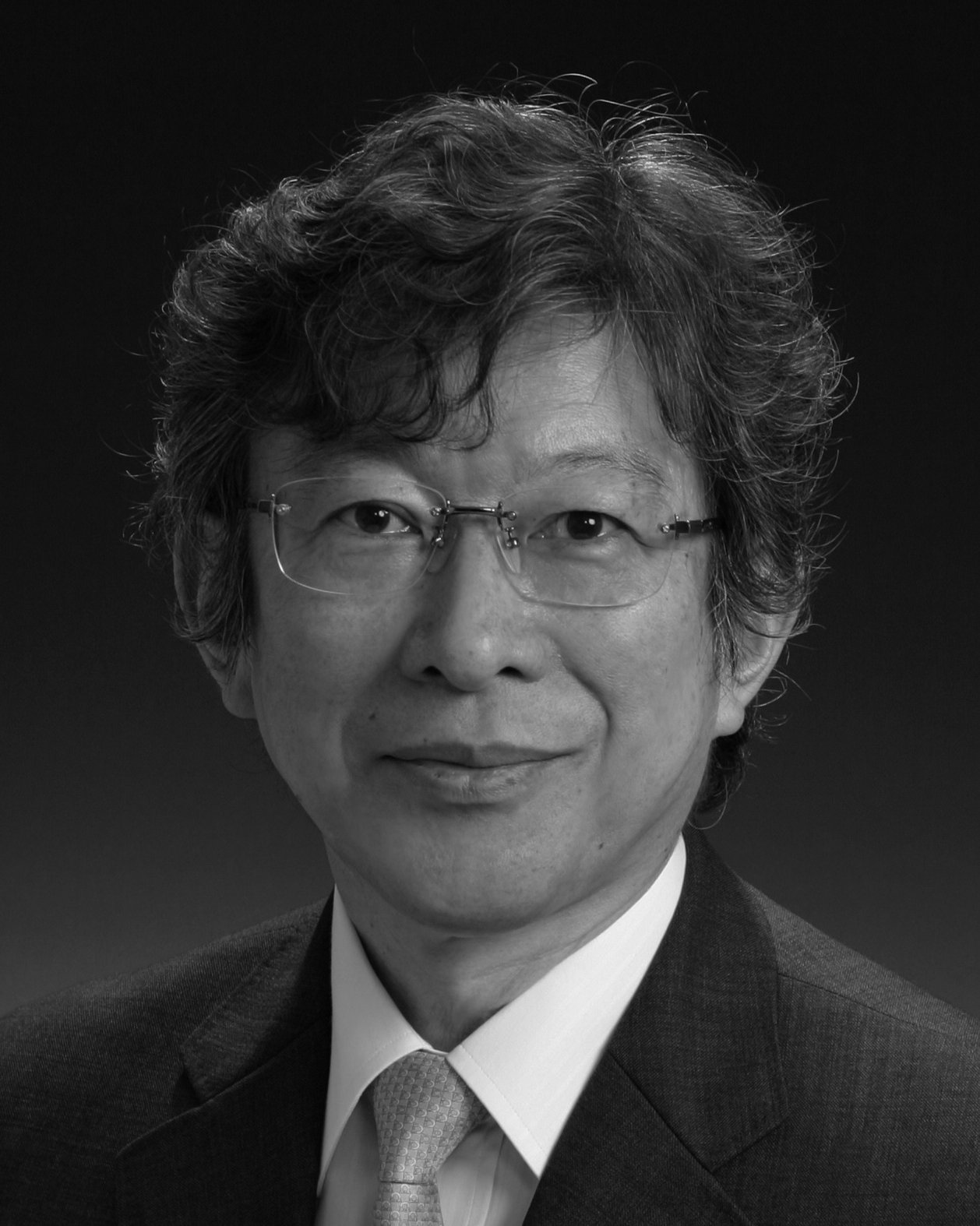}
Kazuhiro Kosuge is a Professor in the Department of Robotics at Tohoku University, Japan. He received the B.S., M.S., and Ph.D. in Control Engineering from the Tokyo Institute of Technology, in 1978, 1980, and 1988 respectively. 

From 1980 through 1982, he was a Research Staff in the Production Engineering Department, Nippon Denso Co., Ltd. (DENSO Co., Ltd.). From 1982 through 1990, he was a Research Associate in the Department of Control Engineering at Tokyo Institute of Technology. From 1990 to 1995, he was an Associate Professor at Nagoya University. From 1995, he has been at Tohoku University. 

Dr. Kosuge received the JSME Awards for the best papers from the Japan Society of Mechanical Engineers in 2002 and 2005, the RSJ Award for the best papers from the Robotics Society of Japan in 2005. Dr. Kosuge is IEEE Fellow, JSME Fellow, SICE Fellow, RSJ Fellow and a JSAE Fellow. Dr. Kosuge was President of IEEE Robotics and Automation Society for 2010-2011 and IEEE Division X Director for 2015-2016. Dr. Kosuge is a member of IEEE-Eta Kappa Nu and IEEE Vice President for Technical Activities for 2019.
\endbio
\end{document}